\documentclass[10pt,twocolumn,letterpaper]{article}

\usepackage{cvpr}              

\usepackage{array}
\usepackage{booktabs}
\usepackage{colortbl}
\usepackage{multirow}
\usepackage{pifont}
\usepackage{xspace}
\usepackage{listings}

\makeatletter
\@ifundefined{nolinenumbers}{\newenvironment{nolinenumbers}{}{}}{}
\makeatother

\newcommand{\method}{Workflow-CAD\xspace}
\newcommand{\bench}{Workflow-CAD-Bench\xspace}
\newcommand{\cmark}{\ding{51}}
\newcommand{\xmark}{\ding{55}}

\usepackage{microtype}

\definecolor{cvprblue}{rgb}{0.21,0.49,0.74}
\usepackage[pagebackref,breaklinks,colorlinks,allcolors=cvprblue]{hyperref}

\def\confName{CVPR}
\def\confYear{2027}

\title{Mining and Simulating Engineer Workflows for Agentic CAD Modeling}

\author{
Sunqi Fan$^{1}$ \quad Runqi Yin$^{1}$ \quad Lingshan Chen$^{1}$ \quad
Zhiguang Liu$^{2}$\\
Yongming Rao$^{2}$ \quad Qun-Ce Xu$^{1}$ \quad Shi-Min Hu$^{1}$\\
$^{1}$Tsinghua University \quad $^{2}$Tencent Hunyuan\\
\texttt{\{fansq20,yrq24,cls24\}@mails.tsinghua.edu.cn}\\
\texttt{leozgliu@tencent.com} \quad
\texttt{raoyongming95@gmail.com}\\
\texttt{quncexu@tsinghua.edu.cn} \quad
\texttt{shimin@tsinghua.edu.cn}
}

\usepackage{pdfrender}
\definecolor{wfnavy}{HTML}{0B3A73}
\definecolor{wfsky}{HTML}{6CC8F0}
\definecolor{wflight}{HTML}{E8F6FE}
\newcommand{\wfoutline}[2]{%
   \textpdfrender{TextRenderingMode=FillStroke,LineWidth=0.9pt,
      StrokeColor=wfnavy,FillColor=#1}{#2}}
\newcommand{\wfwordmark}{{%
   \fontencoding{T1}\fontfamily{Montserrat-LF}\fontseries{black}%
   \fontshape{it}\fontsize{22}{22}\selectfont
   \makebox[0pt][l]{\hspace{0.03in}\raisebox{-0.03in}{%
      \textcolor{wfnavy!30}{Workflow-CAD}}}%
   \wfoutline{wflight}{Workflow}\wfoutline{wfsky}{-CAD}%
}}
\newcommand{\wfprojectpage}{%
   {\large Project Page: \href{https://workflowcad.github.io/}{https://workflowcad.github.io/}}}
\makeatletter
\def\@maketitle{
   \newpage
   \null
   \vspace*{-0.05in}
   \noindent
   \makebox[\textwidth][l]{%
      \raisebox{-0.14in}{\includegraphics[height=0.5in]{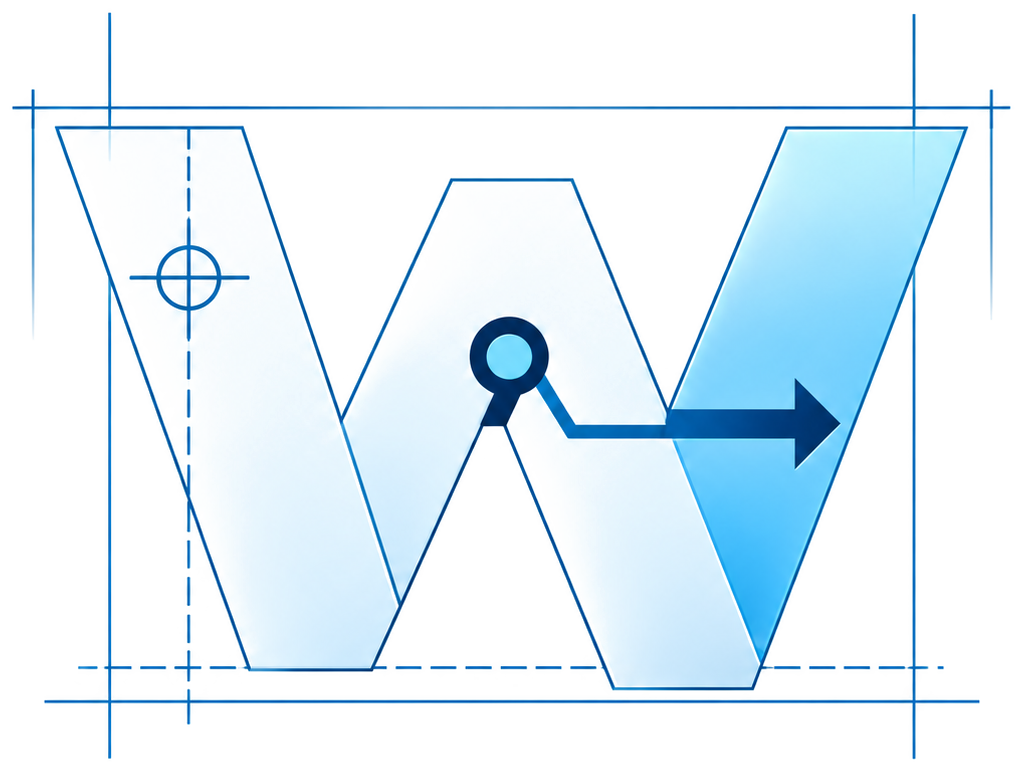}}%
      \hspace{0.06in}%
      \wfwordmark
      \hfill
      \raisebox{-0.13in}{\includegraphics[height=0.44in]{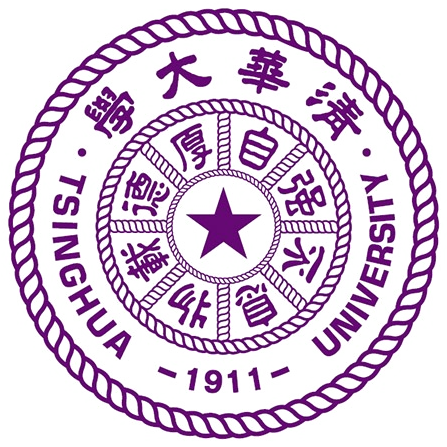}}%
      \hspace{0.14in}%
      \raisebox{-0.13in}{\includegraphics[height=0.44in]{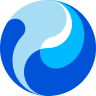}}%
   }\par
   \vspace{0.08in}
   \noindent\color{wfnavy}\rule{\textwidth}{0.6pt}\color{black}
   \vspace{0.22in}
   \begin{center}
      {\Large\bfseries\@title\par}
      \vspace{20pt}
      {\large
        \lineskip .5em
        \begin{tabular}[t]{c}
          \@author
        \end{tabular}
        \par
      }
      \vspace{8pt}
      \wfprojectpage
      \vskip .5em
      \vspace*{6pt}
   \end{center}
}
\makeatother

\begin{document}
\addtocontents{toc}{\protect\setcounter{tocdepth}{-1}}
\twocolumn[{
\renewcommand\twocolumn[1][]{#1}
\maketitle
\centering
\includegraphics[width=\textwidth]{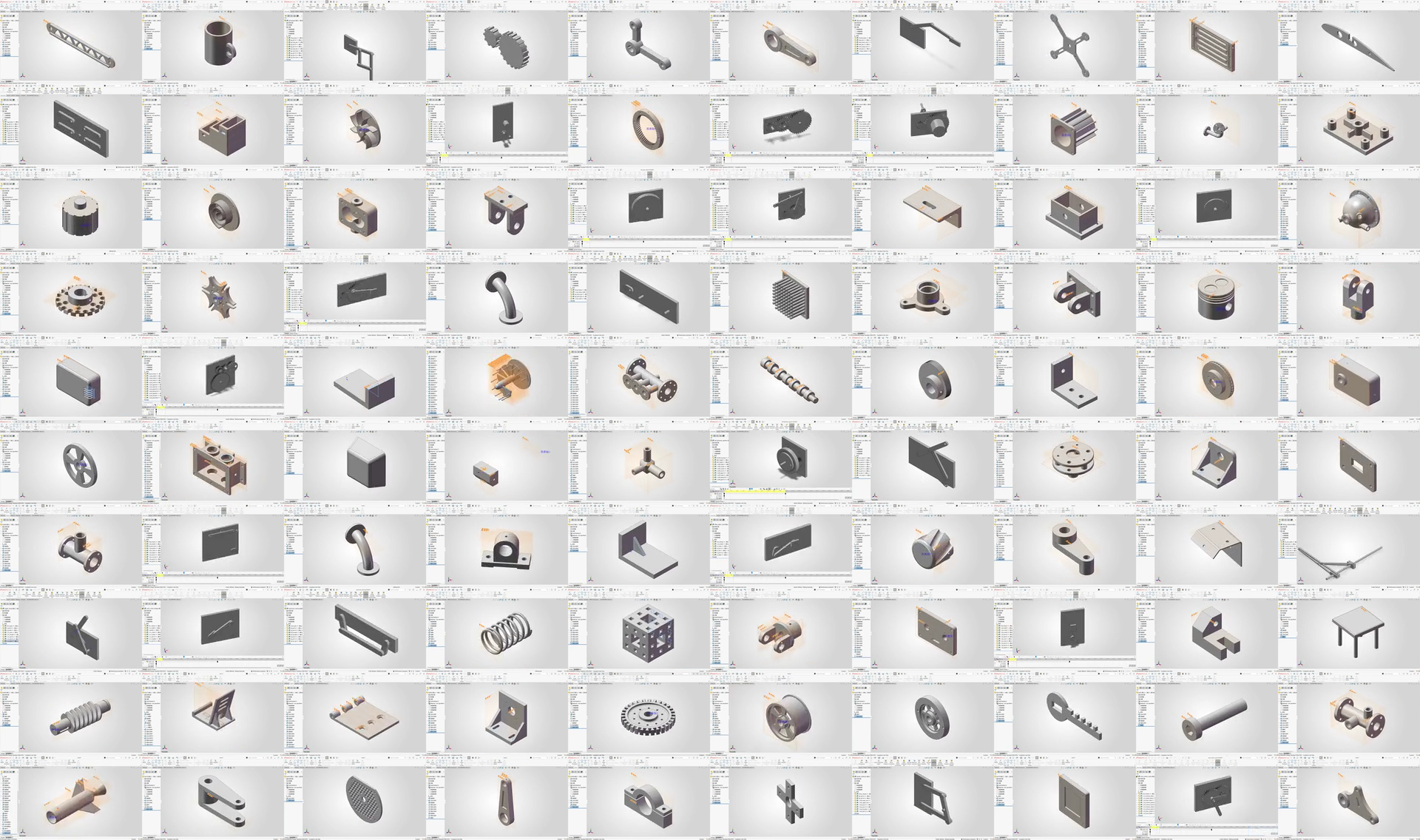}
\captionsetup{hypcap=false}
\captionof{figure}{Examples of native CAD models created fully autonomously by our proposed \method system.}
\label{fig:teaser}
\vspace{1em}
}]
\begin{figure*}[t]
\centering
\includegraphics[width=\textwidth]{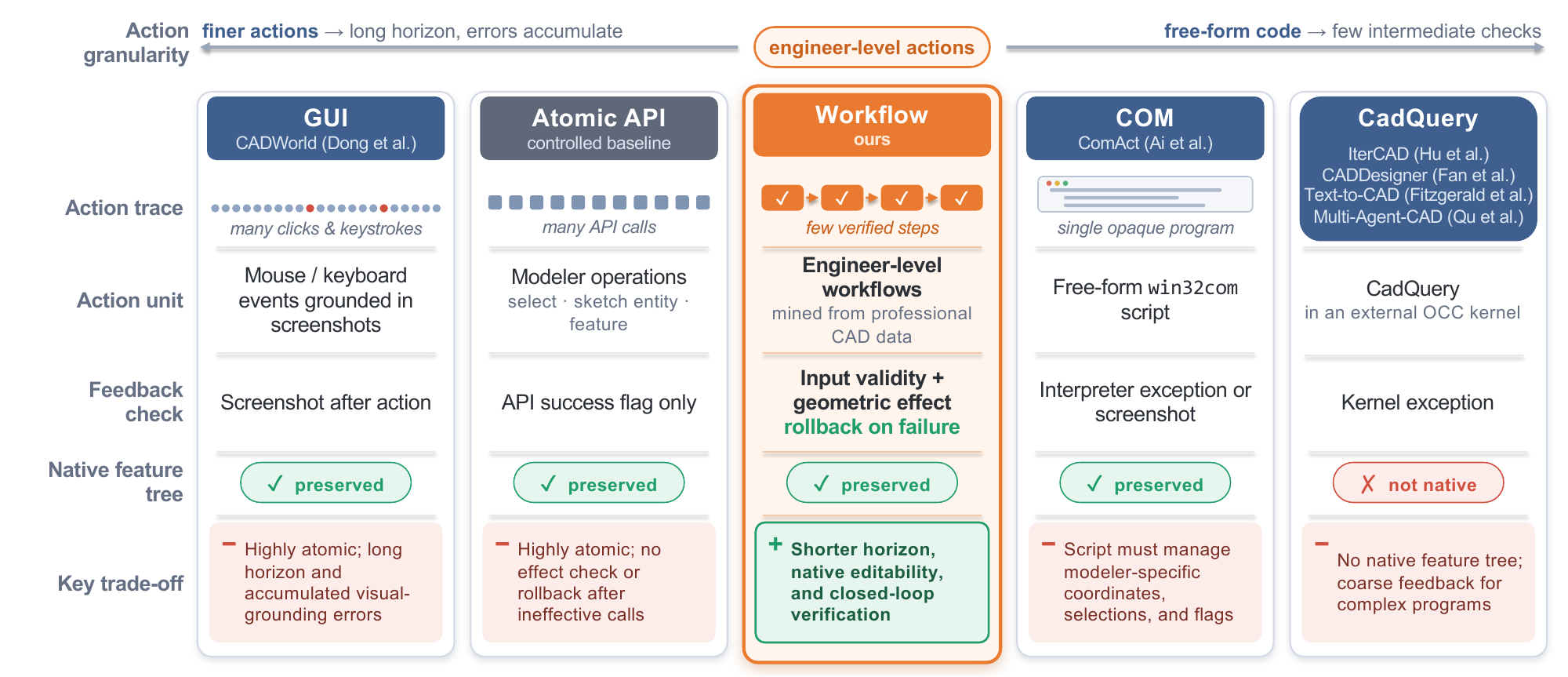}
\caption{Comparison of five action interfaces for LLM-driven CAD.
 GUI,
COM, and CadQuery summarize interfaces used by prior work; Atomic API is our
low-level controlled baseline; and Workflow is our proposed engineer-level interface.
}
\label{fig:levels}
\end{figure*}
\begin{abstract}
Large Language Models (LLMs) are increasingly employed to automate CAD modeling, yet
most Agentic CAD systems either generate intermediate modeling code like CadQuery or emit
low-level API calls one by one. The former struggles to generate complex CAD
models in a single pass, whereas the latter creates excessively long-horizon tasks in
which errors accumulate.
To overcome these limitations, we present \method, a
closed-loop agentic CAD modeling system whose action space consists of engineer-level workflows extracted
from massive feature trees and assembly relations.
We also
introduce \bench, a holistic Agentic CAD Modeling benchmark spanning text-to-part generation,
CAD reconstruction from three-view drawings, and generation of complex, physically plausible and
kinematically valid assemblies.
Experiments show that \method outperforms existing agentic CAD modeling
systems by a large margin. Especially on
motion-aware assembly, \method improves pass@1 over CADDesigner by 22
percentage points.
Beyond comparing Agentic CAD Modeling
systems, \bench also evaluates the underlying LLMs' ability to use tools and
perform autonomous spatial modeling. 
Code and benchmark data are released at
\url{https://github.com/workflowcad/Workflow-CAD}.
\end{abstract}

\section{Introduction}
\label{sec:intro}


\begin{figure*}[t]
\centering
\includegraphics[width=\textwidth]{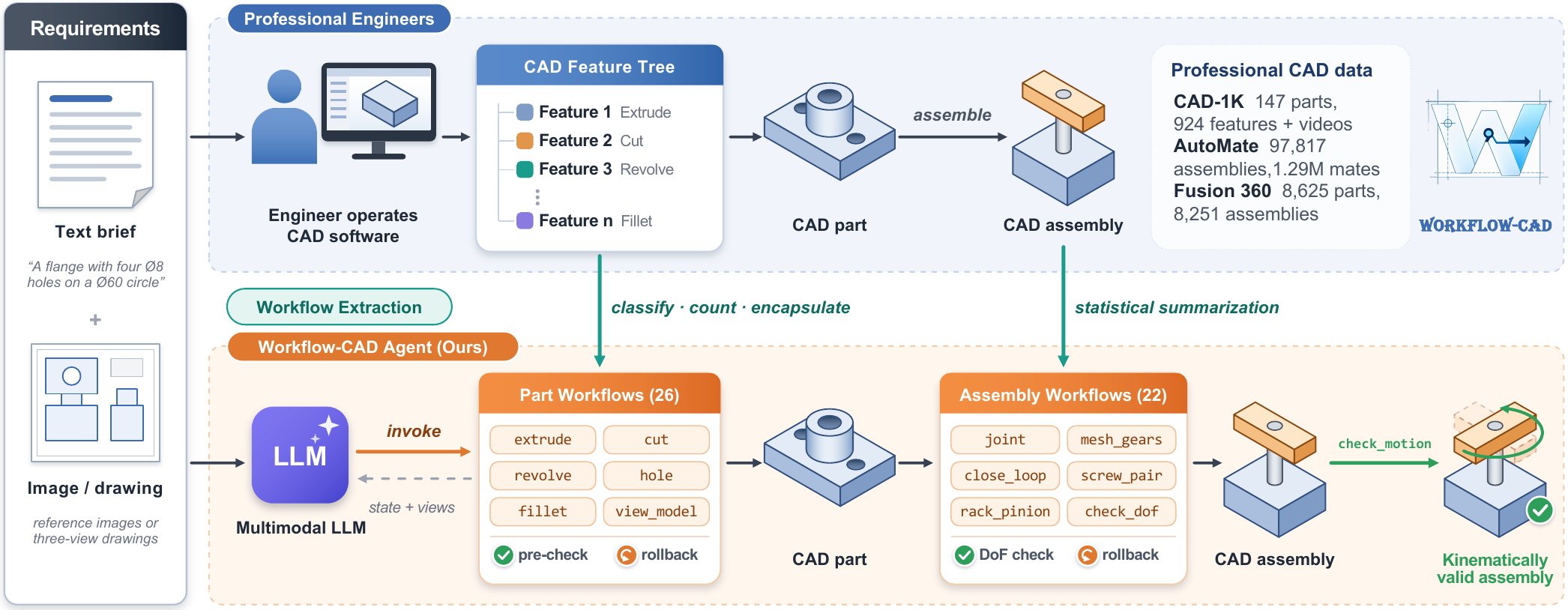}
\caption{Overview of \method. Top: human engineers operate CAD software,
producing large collections of CAD models with complete editing histories.
Bottom: an agent uses workflows mined and extracted from these histories to
construct CAD models automatically in the same way as human engineers.}
\label{fig:overview}
\end{figure*}

Computer-Aided Design (CAD) is an appealing yet demanding frontier for
agents. Given a textual specification or an engineering
drawing, an agent is supposed to translate design intent into precise geometry, maintain
relations across many construction steps, and ultimately produce a model
that remains editable in the engineer's native modeler. For assemblies,
geometric plausibility is not sufficient: components must also be constrained
so that the resulting mechanism exhibits the intended motion. Recently, 
LLMs have made progress by generating CAD programs
directly~\citep{khan2024text2cad,DBLP:conf/iccv/RukhovichDMCKA25,hu2026itercad} or by
operating professional software through graphical or programmatic
interfaces~\citep{DBLP:journals/cad/FanNYWLZTTD26,dong2026cadworld,li2025guieda,DBLP:journals/corr/abs-2606-13239}.
Nevertheless, reliable construction of complex, native CAD models remains
difficult. We argue that this difficulty is not solely a limitation of model
reasoning; it is also a consequence of the \emph{action space} through which
the model is asked to reason.

Existing CAD interfaces expose actions at different levels of abstraction (\cref{fig:levels}).
GUI agents operate the
native modeler, but reduce each engineering operation to many visually
grounded steps, creating excessively long-horizon action sequences in which
errors accumulate. CadQuery compresses a task into a program, but execution
provides only coarse kernel feedback and produces no editable feature tree in
the engineer's native modeler. COM scripts can preserve that native tree, yet
the generated code must directly manage modeler-specific coordinates,
selections, and feature options, while failures are exposed mainly through
raw interpreter errors. The Atomic API baseline replaces GUI events with
typed modeler operations, but remains highly atomic and excessively long-horizon. 
These limitations expose a common trade-off: existing interfaces either
require long sequences of low-level actions or place an entire modeling
procedure inside free-form code with limited intermediate checks. 


We therefore introduce the \emph{workflow} to resolve this trade-off. 
A workflow is a typed, reusable composition of modeler actions that expresses an engineer's
intent while hiding low-level API conventions. 
It is more structured than free-form code,
substantially shorter-horizon than GUI or atomic interaction, and still
executes inside the native CAD system. Crucially, a workflow is not merely a
macro that batches API calls. It resolves semantic geometric references,
normalizes modeler-specific coordinate conventions, validates inputs before
execution, checks the resulting geometric effect, and rolls back ineffective
or invalid changes. The agent can consequently plan in terms of design intent
while retaining closed-loop control over the actual model state. This action
space preserves precision and editability without forcing the LLM to repeatedly
solve implementation details that are incidental to the design task.

The central question is then which workflows should be included in this action space.
A catalog designed only from intuition can omit important operations,
duplicate special cases, or encode the habits of its authors rather than those
of practicing engineers. 
We instead extract workflows from massive human-created CAD parts
and assemblies from different data sources. Their feature trees and assembly relations record how
engineers construct and validate models.
We identify
recurring engineering intent and construction patterns, including reusable
profiles and parameterized features for parts, and recurring structural and kinematic patterns for
assemblies. A pattern is retained as a workflow only when it can be expressed
with reusable parameters, checked before execution, and verified by an
explicit success condition. The resulting workflows therefore distill
engineering practice rather than merely imitating isolated command sequences.
Building on these mining results, we present \method, a closed-loop agent whose
action space is engineer-level workflows for CAD modeling.


Many prior methods are limited to modeling individual, static
parts~\citep{earthtojake_text_to_cad,mac2026,hu2026itercad,DBLP:journals/corr/abs-2606-13239}.
However, in practice, a functional product is usually an assembly of multiple components whose
geometry, constraints, and motion are inseparable: parts that look correct in
isolation may interfere after assembly, remain under-constrained, or fail to
transmit the intended motion. \method therefore treats assembly construction
and motion verification as first-class agent actions. It creates native mate
and joint structures, checks mobility and interference, drives the mechanism,
and verifies the resulting kinematics before completion. These capabilities
translate into large gains in our benchmark (\cref{tab:main}):
\method exceeds CADDesigner by 22 percentage points in pass@1 on the assembly track, 40.5
points in structure score, and 32.0 points in motion score.
Visualizations are provided in \cref{fig:teaser} and
Appendix Section~\ref{sec:suppl:visualization}.


Progress in this direction is also hindered by the lack of a comprehensive 
agentic CAD benchmark.
Existing evaluations largely emphasize a
single input modality or isolated part generation, and commonly compare an
output against one reference shape or judge its rendered
appearance~\citep{wu2021deepcad,khan2024text2cad}. Such
protocols do not reveal whether every stated requirement is satisfied, whether
an alternative valid design is accepted, or whether an assembled mechanism
actually moves as intended. To fill this gap, we introduce \bench, a holistic
benchmark spanning text-to-part, three-view reconstruction,
and motion-aware assembly tasks. It evaluates delivered geometry with
deterministic, fine-grained checks and evaluates assemblies through both
their constraint structure and dynamic motion. 
The benchmark thereby tests
not only plausible geometry, but functionally valid engineering
models.
We systematically compare a broad range of agentic CAD systems and LLM
backbones, showing that GPT-6-Astra achieves outstanding CAD modeling
performance when paired with our agentic system, outperforming the human baseline by 52 percentage points in overall pass@1.

\noindent{\textbf{Contributions.}}
Our contributions are threefold:
\begin{itemize}
    \item We formulate agentic CAD as an action-space design problem and 
    introduce engineer-level workflows as an interface. We build
    \method, an efficient and practical agentic CAD modeling system for
    both part construction and motion-aware assembly.
    \item Our key methodological innovation is a data-driven approach that
    mines reusable workflows from large collections of static feature trees
    and assembly relations, distilling the construction experience encoded in
    human-created CAD models into workflows.
    \item We introduce \bench, a holistic and multi-track benchmark for agentic CAD modeling, together
    with verified evaluation scripts that jointly
    measure static geometry and dynamic mechanism behavior. 
\end{itemize}

\section{Related work}
\label{sec:related}

\noindent{\textbf{Learning-based CAD generation.}}
Learning-based CAD generation has progressed from unstructured 3D shape synthesis toward structured, editable, and parametric design programs.
Some approaches learn structured CAD programs and feature-based representations~\citep{wu2021deepcad,xu2022skexgen,DBLP:journals/tog/WillisPLCDLSM21,DBLP:conf/cvpr/LiGLB024,DBLP:journals/tvcg/ZhangJZFWZ25,DBLP:conf/cvpr/PyatovBGBVFFWB26}.
A complementary line translates point clouds and scans into parametric CAD structures~\citep{DBLP:conf/cvpr/LiuOWS24,DBLP:conf/cvpr/KhanDAC0A24,DBLP:conf/iccv/RukhovichDMCKA25,DBLP:conf/nips/LiuRDZD25,DBLP:conf/nips/KaradenizMRCKA25}.
Beyond reconstruction, recent methods synthesize or edit executable modeling
programs from sketches, images, and natural-language
instructions~\citep{DBLP:journals/tog/LiPBM20,DBLP:conf/siggrapha/0004UHT0D0EYG25,DBLP:conf/cvpr/ChenWC0YZYFLHL25,DBLP:conf/aaai/WangCLXXZY25,DBLP:conf/nips/GuanWXZXY25,khan2024text2cad,DBLP:conf/icml/WangYS025,DBLP:journals/corr/abs-2603-11831,lu2026cadforge,DBLP:conf/iclr/ZhangS00B25,DBLP:conf/icml/YuanS0025,DBLP:conf/cvpr/YuanLXWM26}.
Beyond single parts, JoinABLe~\citep{DBLP:conf/cvpr/WillisJCT0GSTLS22} learns
bottom-up assembly by predicting joints between parametric parts, and
ASSEMCAD~\citep{DBLP:journals/corr/abs-2607-05123} generates assemblies from
natural language through axiom-grounded specifications and a port-and-mate library.
Recently, LLM agents combine planning, program synthesis, execution, and
feedback to operate professional software and construct native CAD models
~\citep{DBLP:conf/nips/YangJWLYNP24,DBLP:journals/tcad/WuHZYZZY24,DBLP:conf/iccv/MallisKCRFCKA25,DBLP:journals/corr/abs-2608-00891,DBLP:journals/corr/abs-2608-26238}.
Text-to-cad directly generates
CadQuery programs from text~\citep{earthtojake_text_to_cad}, while
IterCAD~\citep{hu2026itercad} iteratively revises CadQuery programs using
compiler, execution, and rendered-view feedback.
ComAct operates professional CAD software on Windows through the Component
Object Model (COM)~\citep{DBLP:journals/corr/abs-2606-13239}.
Multi-Agent-CAD uses the \texttt{build123d} Python interface for part 
generation with multiple specialized agents~\citep{mac2026}. 
CADDesigner~\citep{DBLP:journals/cad/FanNYWLZTTD26} is a whole-program CAD 
code-generation agent without native feature trees, making subsequent feature-level editing
difficult because dimensions, constraints, 
and feature dependencies are not preserved.
In contrast, our Workflow-CAD introduces a compact set of workflow interfaces
and invokes workflows inside a native CAD system, 
preserving complete feature trees.
We also summarize the differences between our work and other Agentic CAD Modeling systems in \cref{fig:levels}.

\begin{table}[t]
\newcommand{\tabyes}{\textcolor{ForestGreen}{\cmark}}
\newcommand{\tabno}{\textcolor{BrickRed}{\xmark}}
\newcommand{\tabhalf}{\textcolor{BurntOrange}{$\circ$}}
\caption{Comparison of CAD generation benchmarks.
Multimodal: inputs beyond text (\eg, images, drawings, meshes);
Ref.\ geometry: direct comparison with a reference shape;
Fine-grained: per-requirement and multidimensional checks.
\tabyes: supported; \tabhalf: partially supported; \tabno: not supported.}
\label{tab:bench-compare}
\centering
\footnotesize
\setlength{\tabcolsep}{0.5pt}
\begin{tabular*}{\columnwidth}{@{\extracolsep{\fill}}lccccc@{}}
\toprule
\textbf{Benchmark} &
\textbf{\begin{tabular}[b]{@{}c@{}}Multi-\\modal\end{tabular}} &
\textbf{\begin{tabular}[b]{@{}c@{}}Assem-\\bly\end{tabular}} &
\textbf{\begin{tabular}[b]{@{}c@{}}Ref.\\geometry\end{tabular}} &
\textbf{\begin{tabular}[b]{@{}c@{}}Motion\\checks\end{tabular}} &
\textbf{\begin{tabular}[b]{@{}c@{}}Fine-\\grained\end{tabular}} \\
\midrule
Text2CAD~\citep{khan2024text2cad}                     & \tabno  & \tabno   & \tabyes & \tabno   & \tabno   \\
CADTests~\citep{DBLP:journals/corr/abs-2605-07807}    & \tabno  & \tabno   & \tabno  & \tabno   & \tabyes  \\
CADGenBench~\citep{cadgenbench2026}                   & \tabyes & \tabno   & \tabyes & \tabno   & \tabno   \\
BenchCAD~\citep{DBLP:journals/corr/abs-2605-10865}    & \tabyes & \tabno   & \tabyes & \tabno   & \tabno   \\
CADBench~\citep{DBLP:journals/corr/abs-2605-10873}    & \tabyes & \tabno   & \tabyes & \tabno   & \tabno   \\
Op-CAD~\citep{bai2026op}                              & \tabno  & \tabno   & \tabyes & \tabno   & \tabno   \\
CADWorld~\citep{dong2026cadworld}                     & \tabyes & \tabyes  & \tabhalf & \tabno  & \tabyes  \\
CADEngBench~\citep{DBLP:journals/corr/abs-2608-09296} & \tabyes & \tabhalf & \tabno  & \tabhalf & \tabyes  \\
\midrule
\textbf{\bench (ours)}                                & \tabyes & \tabyes  & \tabyes & \tabyes  & \tabyes  \\
\bottomrule
\end{tabular*}
\end{table}


\paragraph{CAD model evaluation.}
Generic 3D metrics measure geometric fidelity and distribution
quality~\citep{DBLP:conf/icml/AchlioptasDMG18,DBLP:conf/cvpr/FanSG17,DBLP:conf/cvpr/GroueixFKRA18},
and recent 3D evaluators add fine-grained, interpretable quality
assessment~\citep{DBLP:conf/cvpr/DuggalHMKFSK0FM25,DBLP:conf/nips/ZhangZCWLPLL25},
but neither tests engineering requirements.
CAD benchmarks further check executability, construction structure, and explicit
geometric, topological, interface, and engineering
requirements~\citep{DBLP:journals/corr/abs-2605-10873,DBLP:journals/corr/abs-2605-10865,bai2026op,DBLP:journals/corr/abs-2605-07807,cadgenbench2026,DBLP:journals/corr/abs-2608-09296},
whereas assembly work mainly checks static
constraints~\citep{DBLP:journals/corr/abs-2607-05123,zheng2026spada},
mate-level motion labels~\citep{DBLP:journals/corr/abs-2208-01779},
or VLM judgments~\citep{DBLP:journals/corr/abs-2607-05123}.
As compared in \cref{tab:bench-compare} (details in
Appendix Section~\ref{sec:suppl:bench-compare}), most benchmarks evaluate only
single parts; CADWorld~\citep{dong2026cadworld} checks only the joint metadata of assemblies, and
CADEngBench verifies the motion of a single joint between two bodies.
\bench instead drives complete native mechanisms or probes prescribed poses,
and is the only one that combines multimodal inputs, complete assemblies,
mechanism-level motion checks, and fine-grained requirement checks.

\section{Method}
\label{sec:method}

As summarized in \cref{fig:overview}, our method has three
components: design principles that make workflows consistent and
checkable, data-driven extraction of part and assembly workflows, and an
agentic CAD modeling system that executes each workflow and verifies its effect.



\subsection{Workflow design principles}
\label{sec:method:interface}


The six principles in \cref{tab:workflow-principles} make workflows
consistent across modeling contexts while hiding modeler-specific conventions;
full semantics and schemas are given in
Appendix Sections~\ref{sec:suppl:part-agent}
and~\ref{sec:suppl:assembly-agent}.

\begin{table}[ht]
\caption{Workflow design principles.}
\label{tab:workflow-principles}
\centering
\footnotesize
\setlength{\tabcolsep}{2.5pt}
\begin{tabular*}{\columnwidth}{@{\extracolsep{\fill}}
    >{\centering\arraybackslash}p{0.05\linewidth}
    >{\raggedright\arraybackslash}p{0.25\linewidth}
    >{\raggedright\arraybackslash}p{0.59\linewidth}@{}}
\toprule
\textbf{\#} & \textbf{Principle} & \textbf{Design and runtime behavior} \\
\midrule
\textbf{1} & A shared coordinate frame &
Canonical $(u,v)$ coordinates; runtime mapping to the selected plane or face. \\
\midrule
\textbf{2} & Intuitive feature directions &
World direction and termination rule; runtime mapping to native options. \\
\midrule
\textbf{3} & Semantic references &
Meaning- or geometry-based descriptions resolved to native entities. \\
\midrule
\textbf{4} & Typed profiles &
Shared 2D primitives with named dimensions reusable by later features. \\
\midrule
\textbf{5} & Pre-execution checks &
Region validation and repair hints without modifying the current model. \\
\midrule
\textbf{6} & Native visual inspection &
Labeled native views for appearance, occlusion, and component placement. \\
\bottomrule
\end{tabular*}
\end{table}

\subsection{Workflow extraction}
\label{sec:method:tools}

\paragraph{Mining part workflows.}
We extract part workflows from approximately 8,500 CAD parts with complete
feature trees, drawn from two data sources
(CAD-1K~\citep{markovai2026cad1k} and Fusion 360 Gallery~\citep{DBLP:journals/tog/WillisPLCDLSM21}).
Full corpus statistics are given in
Appendix Section~\ref{sec:suppl:part-agent}.
For each part, we gather two types of information: (i) the native feature type
and its predecessor in the feature tree, (ii) for every sketch used by a
downstream modeling feature, a structural summary of its geometric element
types, connectivity, and contour types.
We convert this information into workflows in three steps: group recurring
operations by engineering intent, factor their sketches into shared profile
primitives, and retain only candidates that are common and can be checked
reliably.

\textbf{First}, we group low-level feature names that express the same
engineering intent.
Each resulting operation family becomes a candidate workflow.
We merge variants when they share the same geometry and checks: for example,
blind, mid-plane, and up-to-surface extrusion are options of one
\texttt{extrude} workflow. We keep operations separate when their intent or
validity rules differ.
\textbf{Second}, we factor the collected sketches into reusable profile primitives.
This shared profile language represents \emph{what to draw} independently of
\emph{which modeling operation uses it}, avoiding a separate workflow for
every observed outline. 
\textbf{Third}, we rank candidates by how often the corresponding intent appears in
professional tasks. We retain a geometry-changing candidate only when it has
a reusable schema, a pre-execution check, and a measurable success
condition with rollback. 

The resulting catalog contains 25 extracted modeling workflows
and eight shared profile primitives. We additionally add a
visual inspection workflow, giving the
multimodal part agent 26 callable workflows in total. Full evidence and schemas are
provided in Appendix Section~\ref{sec:suppl:part-agent}.

\paragraph{Mining assembly workflows.}
\label{sec:method:mining-assembly-workflows}
Part feature trees do not reveal how components should be connected or move,
so we separately extract assembly workflows based on approximately 97,800 CAD assemblies
from two different data sources (AutoMate~\citep{jones2021automate} and the Fusion 360
Gallery Assembly dataset~\citep{DBLP:journals/tog/WillisPLCDLSM21}). AutoMate
provides large-scale mate graphs, while Fusion adds joint order and the
geometry used to create each joint; full dataset statistics are given in
Appendix Section~\ref{sec:suppl:assembly-agent}. We treat components as graph nodes and
mates or joints as edges, then record the following statistics: grounded component, joint semantics,
breadth-first depth from the grounded component, independent cycles, coupled relations, and
estimated mobility. 

The assembly graphs from the two datasets expose five recurring regularities.
\textbf{First}, assemblies usually grow outward from one grounded main
component as engineers attach new components to the structure already built.
\textbf{Second}, joint semantics determine which geometry to select and which
native mates to create. \textbf{Third}, engineers close kinematic loops only
after constructing the open connection tree. \textbf{Fourth}, they add motion
couplings only after the supporting joints exist. \textbf{Fifth}, frequent
floating or over-constrained components make mobility checks necessary.

We distill these regularities into five natural-language strategies that guide
the order of construction, and use them to organize 20 mined executable
workflows. A strategy tells the agent how to proceed, whereas a workflow is a
concrete action it can call on the current assembly state; one strategy may
invoke several workflows, and one workflow may support several strategies.
We additionally provide prescribed-pose validation
(\texttt{check\_pose}) and native visual inspection
(\texttt{view\_assembly}), giving 22 callable workflows in total. Detailed
evidence, the strategy-to-workflow mapping, and the complete catalog are
provided in Appendix Section~\ref{sec:suppl:assembly-agent}.
Each callable workflow validates its own operation.
For example, the mined \texttt{revolute\_joint} workflow captures the common
shaft-in-hole pattern. Given the cylindrical and end faces of the two
components, it aligns their axes with a concentric mate and seats the parts
with a coincident mate, leaving only rotation about the shared axis. 


\subsection{Agentic loop}
\label{sec:method:agent}

At each turn, the LLM selects a workflow from the latest structured
observation. The runtime resolves references, validates
inputs, executes the underlying operations, verifies their effect, and either
returns the updated native state or rolls back with a repair hint. 

\noindent\textbf{Part-agent loop.}
Part workflows verify feature effects through volume, extent, and body-count
changes. Part-level \texttt{finish} is refused unless exactly one solid body
exists. Prompt and details are provided in
Appendix Section~\ref{sec:suppl:part-agent}.

\noindent\textbf{Assembly-agent loop.}
The five strategy recipes are included in the task brief, and the 22 assembly
workflows are exposed through function calling.
Assembly workflows operate on components, joints, and motion. A failed
multi-mate operation removes its partial mate set, and a conflicting loop
closure is rolled back before another constraint configuration is tried.
Finish requires a saved
assembly plus either passing sampled motion or a valid prescribed pose with
the expected mobility. Full observations and schemas are given in
Appendix Section~\ref{sec:suppl:assembly-agent}.

\setcounter{dbltopnumber}{1}
\begin{table*}[t]
\caption{Unified results of agentic CAD modeling systems on the three
\bench tracks. 
Except for IterCAD, which uses its own trained model, all systems use
Qwen3.8-Flash-Next~\citep{qiu2026designqwen38nextarchitectureevaluation}
as the LLM backbone.
Tokens denote the mean token count per task within each track and overall.
Overall pass@1, score and token costs aggregate performance across the three tracks.
The best and second-best available values in each column are shown in
\textcolor{red}{red} and \textcolor{blue}{blue}, respectively; lower is
better for Tokens, while higher is better for all other metrics.
``--'' means that a method does not support the corresponding track.}
\label{tab:main}
\centering
\scriptsize
\setlength{\tabcolsep}{1pt}
\resizebox{\textwidth}{!}{%
\begin{tabular}{l cccc cccc cccc ccc}
\toprule
& \multicolumn{4}{c}{\textbf{Text-to-part}} &
\multicolumn{4}{c}{\textbf{Three-view-to-part}} &
\multicolumn{4}{c}{\textbf{Assembly}} &
\multicolumn{3}{c}{\textbf{Overall}} \\
\cmidrule(lr){2-5}\cmidrule(lr){6-9}\cmidrule(lr){10-13}
\cmidrule(lr){14-16}
\textbf{System} &
Pass@1 & Score & Form & Tokens &
Pass@1 & Score & Image & Tokens &
Pass@1 & Score & Motion & Tokens &
Pass@1 & Score & Tokens \\
\midrule
text-to-cad~\citep{earthtojake_text_to_cad}
  & 0.16 & 29.3 & 0.33 & 64.6k
  & - & - & - & -
  & - & - & - & -
  & - & - & - \\
Multi-Agent-CAD~\citep{mac2026}
  & 0.21 & 42.6 & 0.52 & 69.0k
  & - & - & - & -
  & - & - & - & -
  & - & - & - \\
IterCAD~\citep{hu2026itercad}
  & 0.06 & 22.9 & 0.22 & 9.1k
  & 0.04 & 17.9 & 13.6 & 18.7k
  & - & - & - & -
  & - & - & - \\
ComAct~\citep{DBLP:journals/corr/abs-2606-13239}
  & 0.12 & 33.4 & 0.53 & 272.2k
  & 0.14 & 41.6 & 54.1 & 285.0k
  & - & - & - & -
  & - & - & - \\
Atomic API
  & 0.23 & 40.2 & 0.49 & 29.8k
  & 0.36 & 61.4 & 71.0 & 31.3k
  & 0.00 & 22.8 & 4.1 & 139.9k
  & 0.21 & 41.2 & \textcolor{red}{57.7k} \\
\rowcolor{gray!15}
Human Baseline
  & 0.56 & \textcolor{blue}{78.5} & \textcolor{red}{1.00} & -
  & 0.44 & 76.2 & 78.8 & -
  & 0.22 & \textcolor{blue}{70.2} & \textcolor{blue}{43.2} & -
  & 0.45 & 75.9 & - \\
CADDesigner~\citep{DBLP:journals/cad/FanNYWLZTTD26}
  & \textcolor{blue}{0.85} & \textcolor{red}{96.5} & 0.96 & 228.6k
  & \textcolor{blue}{0.76} & \textcolor{blue}{93.6} & \textcolor{blue}{97.0} & 170.6k
  & \textcolor{blue}{0.28} & 50.6 & 36.3 & 4373.2k
  & \textcolor{blue}{0.69} & \textcolor{blue}{84.3} & 1250.3k \\
\method (ours)
  & \textcolor{red}{0.88} & \textcolor{red}{96.5} & \textcolor{blue}{0.98} & 170.1k
  & \textcolor{red}{0.84} & \textcolor{red}{96.4} & \textcolor{red}{97.6} & 136.7k
  & \textcolor{red}{0.50} & \textcolor{red}{91.1} & \textcolor{red}{68.3} & 2691.0k
  & \textcolor{red}{0.78} & \textcolor{red}{95.1} & \textcolor{blue}{792.0k} \\
\bottomrule
\end{tabular}
}
\end{table*}

\begin{table*}[t]
\caption{Results of the \method system across LLM 
backbones, using the same metrics and colors as
\cref{tab:main}. 
}
\label{tab:backbones}
\centering
\scriptsize
\setlength{\tabcolsep}{1pt}
\resizebox{\textwidth}{!}{%
\begin{tabular}{l cccc cccc cccc ccc}
\toprule
& \multicolumn{4}{c}{\textbf{Text-to-part}} &
\multicolumn{4}{c}{\textbf{Three-view-to-part}} &
\multicolumn{4}{c}{\textbf{Assembly}} &
\multicolumn{3}{c}{\textbf{Overall}} \\
\cmidrule(lr){2-5}\cmidrule(lr){6-9}\cmidrule(lr){10-13}
\cmidrule(lr){14-16}
\textbf{Backbone} &
Pass@1 & Score & Form & Tokens &
Pass@1 & Score & Image & Tokens &
Pass@1 & Score & Motion & Tokens &
Pass@1 & Score & Tokens \\
\midrule
GPT-6-Astra~\citep{openai2026gpt6astra}
  & \textcolor{red}{0.97} & \textcolor{red}{99.4} & \textcolor{red}{1.00} & 248.3k
  & \textcolor{red}{1.00} & \textcolor{red}{100.0} & \textcolor{red}{100.0} & \textcolor{blue}{122.5k}
  & \textcolor{red}{0.94} & \textcolor{red}{99.5} & \textcolor{red}{95.7} & \textcolor{blue}{1520.1k}
  & \textcolor{red}{0.97} & \textcolor{red}{99.6} & 534.8k \\
GPT-5.6-Sol~\citep{openai2026gpt56sol}
  & \textcolor{blue}{0.94} & \textcolor{blue}{98.1} & \textcolor{blue}{0.99} & \textcolor{blue}{155.8k}
  & 0.88 & 96.8 & 98.0 & 146.7k
  & 0.74 & 97.2 & \textcolor{blue}{83.7} & 1583.0k
  & \textcolor{blue}{0.88} & \textcolor{blue}{97.6} & \textcolor{blue}{510.3k} \\
Claude-Opus-5~\citep{anthropic2026claudeopus5}
  & 0.85 & 94.8 & 0.97 & 188.2k
  & \textcolor{blue}{0.94} & \textcolor{blue}{98.9} & \textcolor{blue}{99.0} & 172.3k
  & 0.68 & 96.4 & 78.6 & 2621.3k
  & 0.83 & 96.2 & 792.5k \\
Gemini-3.1-Pro~\citep{google2026gemini31pro}
  & 0.84 & 92.9 & 0.95 & \textcolor{red}{88.8k}
  & 0.82 & 93.3 & 97.1 & \textcolor{red}{84.8k}
  & \textcolor{blue}{0.78} & \textcolor{blue}{98.4} & 83.1 & \textcolor{red}{1181.6k}
  & 0.82 & 94.4 & \textcolor{red}{361.0k} \\
DeepSeek-v4-Pro~\citep{deepseekai2026deepseekv4}
  & 0.77 & 93.9 & 0.98 & 221.1k
  & 0.70 & 89.8 & 96.0 & 139.9k
  & 0.44 & 76.9 & 56.5 & 1779.8k
  & 0.67 & 88.6 & 590.5k \\
GLM-5.3~\citep{zai2026glm53}
  & 0.75 & 91.9 & 0.97 & 209.9k
  & 0.66 & 88.7 & 93.6 & 160.4k
  & 0.44 & 88.3 & 61.2 & 2278.7k
  & 0.65 & 90.2 & 714.7k \\
Kimi-K3~\citep{kimiteam2026kimik3openfrontier}
  & 0.50 & 69.2 & 0.79 & 445.4k
  & 0.62 & 82.1 & 85.4 & 289.0k
  & 0.46 & 89.9 & 58.8 & 2523.5k
  & 0.52 & 77.6 & 925.8k \\
\rowcolor{gray!15}
Human Baseline
  & 0.56 & 78.5 & \textcolor{red}{1.00} & -
  & 0.44 & 76.2 & 78.8 & -
  & 0.22 & 70.2 & 43.2 & -
  & 0.45 & 75.9 & - \\
Qwen3.5-397B-A17B~\citep{qwen2026qwen35}
  & 0.23 & 45.2 & 0.83 & 215.7k 
  & 0.20 & 41.0 & 78.0 & 186.2k
  & 0.14 & 38.4 & 19.2 & 1578.0k
  & 0.20 & 42.5 & 548.9k \\
\bottomrule
\end{tabular}
}
\end{table*}

\section{\bench}
\label{sec:bench}

\subsection{Benchmark overview}
\label{sec:bench:scope}

\bench contains 200 tasks across three complementary tracks:
100 text-to-part tasks evaluate geometry
generated from design briefs; 50 three-view-to-part tasks evaluate
reconstruction from orthographic drawings; and 50 motion-aware assembly
tasks evaluate component geometry, mating structure, and functional motion.
Different evaluation methods are used for each track.
Evaluation uses only the delivered CAD model.

\noindent\textbf{CAD model production and annotation.}
\label{sec:bench:anno}
About 50 of the 200 cases are authored by mechanical-domain experts
directly in SolidWorks software and span all functional categories and
difficulty levels.
The remaining approximately 150 cases are produced by imitating these expert
examples: Claude Opus~5.5 calls the native SolidWorks API under continuous
expert guidance and correction.
Every finished case receives at least one expert review.
Once the SolidWorks CAD models are ready, GPT-6.1-Sol generates the
corresponding text prompts, or the three-view drawings are exported
directly from the finished models.

\noindent\textbf{Task taxonomy.}
Across its three tracks, \bench covers a broad range of
mechanisms from diverse functional categories.
The full task lists and taxonomy are provided in Appendix
Section~\ref{sec:suppl:benchmark-tasks}.

\subsection{Text-to-part track}
\label{sec:bench:text}

Each text-to-part task contains a natural language design brief 
that gives the target dimensions and functional details.
The 100 tasks span common sketch-and-feature modeling as well as rare
operations such as routed sweeps and loft transitions.
Difficulty level is orthogonal to these part categories.
The track contains 50 easy, 30 medium, and 20 hard tasks.

\noindent\textbf{Geometric checks.}
\label{sec:bench:scoring}
We use per-task geometric checks that directly test the requirements in each brief.
These checks are executed deterministically using a unified Python script.
Below are some common geometric checks:
\textbf{(1) Topology}: one valid closed solid.
\textbf{(2) Envelope}: sorted bounding box, box in a part frame,
footprint, total height.
\textbf{(3) Features and relations}: whether required features have the
correct counts, dimensions, and spatial relationships.
\textbf{(4) Material distribution}: whether thicknesses, openings,
cross-sections, and symmetries match the specification.
\textbf{(5) Volume.}
The complete check catalog is in
Appendix Section~\ref{sec:suppl:evaluation}.
Every geometric check has a weight.
The \emph{task score} is the weighted percentage of checks that pass.
\emph{Pass@1} is the fraction of tasks solved in a single autonomous run: 
the agent may act over multiple turns without human intervention, and its 
final delivered model must pass all checks.
Two checks are shared in every task: a single valid solid, 
and the primary body that carries the other features. 
Their joint pass rate is reported as the \emph{core-form} rate.




\subsection{Three-view-to-part reconstruction track}
\label{sec:bench:threeview}

This track contains 50 third-angle engineering drawings with 
dimensions and callouts.
The agent must infer all
geometry from the image. The tasks use the same part taxonomy as the
text-to-part track and range from basic prismatic parts to multi-feature models
(Appendix Section~\ref{sec:suppl:benchmark-tasks}).

\noindent\textbf{Drawing-grounded geometric checks.}
In this track, every geometric check is grounded in a
dimension, note, or visible geometric relation in the drawing.
The 50 tasks contain 349 explicit check items (7.0 per task). They verify solid count,
envelope, sections, bores, pockets, wall and floor thicknesses, feature
positions, patterns, symmetry, and volume where the drawing fully determines
it. 

\noindent\textbf{Projection-similarity check.}
We also report an image score 
by comparing the predicted model and the reference model 
through projections along three orthogonal directions. 
Each direction produces silhouette, near-depth, far-depth, and material-thickness maps. 
A pixel matches when
 all map value differences
are within small tolerances.
The directional score is the number of matched pixels divided by the union of the two models' silhouettes.
The final score is 100 times the mean of the three directional scores. 
Details are given in
Appendix Section~\ref{sec:suppl:evaluation}.

\subsection{Motion-aware assembly track}
\label{sec:bench:assembly}

The assembly track contains 50 tasks in which the agent must create the
specified components and assemble them
with the required motion. Briefs define component geometry and functional
relationships but omit initial poses and assembly steps. The suite covers
629 component occurrences across transmissions, closed-loop linkages, and
articulated systems with one to four degrees of freedom; case-level details
are provided in Appendix Section~\ref{sec:suppl:benchmark-tasks}.
Each delivered assembly is evaluated by the following two checks. 

\noindent\textbf{Assembly-structure checks.}
The structure evaluator reads component transforms, mates, and referenced
geometry from the delivered assembly. It scores rigid groups, semantic
joints, mobility, grounding, mate validity, and motion couplings while
accepting different mate decompositions that realize the same joint. The
constraint formulation and criterion weights are detailed in
Appendix Section~\ref{sec:suppl:evaluation}.

\noindent\textbf{Motion checks.}
Motion is evaluated in two ways. For 33 mechanisms with a single rotary
input, the evaluator independently drives that input and measures
trajectories, task-specific kinematics, interference, and unintended drift.
For the other 17 mechanisms, it places the assembly in three prescribed
configurations and checks key-point positions, interference, and fixed
bodies. Both protocols use gated scoring so that a jammed, incomplete, or
drifting mechanism cannot receive high motion credit; exact weights,
sampling, and tolerances are in
Appendix Section~\ref{sec:suppl:evaluation}.

\subsection{Evaluator validation}
\label{sec:bench:validation}

We validate all three track evaluators with independently constructed
references that must score 100 and targeted negative controls that must
trigger the intended failures. For text-to-part, all 100 references pass,
and 206 geometric mutants are detected as expected. For three-view-to-part,
all 50 references score 100, while 133 geometric mutants, rotated references,
and mirrored asymmetric controls verify defect sensitivity and pose
invariance. For assembly, all 50 references score 100 on both structure and
motion with zero sampled interference; injected mating and motion faults are
also correctly detected.
Further validation details are provided in Appendix
Section~\ref{sec:suppl:evaluation}.

\section{Experiments}
\label{sec:exp}

We evaluate three questions: how \method compares with existing agentic CAD
systems (\cref{sec:exp:main,sec:exp:text2cad}), how performance changes with the LLM backbone
(\cref{sec:exp:backbones}), and which components account for its advantage
(\cref{sec:exp:ablation}). 
Experimental settings are detailed in
Appendix Section~\ref{sec:suppl:setup}.

\subsection{Main results on Workflow-CAD-Bench}
\label{sec:exp:main}

\Cref{tab:main} shows that \method is the strongest system across the full
benchmark. It achieves the best or tied-best quality on every track.
Compared with CADDesigner, the
strongest baseline evaluated on all three tracks, \method improves
overall pass@1 by 9 percentage points and score by 10.8 points, while reducing
the average token count by 37\%. 
The advantage is consistent across input modalities and becomes larger as the
task requires more sequential decisions.
The clearest advantage is on motion-aware assembly, the most long-horizon
track. Compared with CADDesigner, \method raises pass@1 from 0.28 to 0.50,
structure score from 50.6 to 91.1,
and motion score from 36.3 to 68.3. It therefore produces substantially more
complete assemblies and more often realizes the required motion, while using
38\% fewer tokens. The text-to-cad,
Multi-Agent-CAD, IterCAD, and ComAct baselines cover fewer tracks and are
substantially weaker on the tracks they support.
The human baseline was produced by graduate students with professional
training in mechanical engineering, with a time limit of two hours per task.
Only CADDesigner and our \method with Qwen3.8-Flash-Next as the LLM backbone
exceed the human baseline in both overall pass@1 and overall score.
Detailed category-level results are reported in
Appendix Section~\ref{sec:suppl:detailed-results}. 

\subsection{LLM backbone comparison}
\label{sec:exp:backbones}

\Cref{tab:backbones} holds the \method system fixed and varies only its LLM
backbone. GPT-6-Astra demonstrates exceptional CAD capability: it achieves
0.97 overall pass@1,
indicating that it can solve
most CAD tasks when paired with the Workflow-CAD interface.
Among the remaining backbones, GPT-5.6-Sol is the strongest overall.
Claude-Opus-5 is particularly competitive
on three-view-to-part,
although its assembly runs consume substantially more tokens.
Gemini-3.1-Pro offers the best performance--token trade-off: it uses the
fewest tokens on every track and only 361.0k tokens overall, while retaining
0.82 overall pass@1 and the second-highest assembly pass@1 of 0.78.
Most tested LLMs, when paired with Workflow-CAD, exceed the human baseline by a
large margin; only Qwen3.5-397B-A17B falls behind the human baseline.
These results provide strong evidence that automated CAD modeling
systems are becoming an essential complement to manual engineering.

\subsection{Results on Text2CAD test set}
\label{sec:exp:text2cad}

We also evaluate 
on the Text2CAD benchmark~\citep{khan2024text2cad}, whose
ground-truth models are drawn from the DeepCAD test
split~\citep{wu2021deepcad}.
Following the evaluation protocol of CADDesigner~\citep{DBLP:journals/cad/FanNYWLZTTD26}
and Pointer-CAD~\citep{DBLP:conf/cvpr/QiWXCZLDMG26}, we use the
abstract-level descriptions as prompts, remove duplicate models, and
sample 200 test cases stratified by command-sequence length.
We report intersection
over union (IoU), Chamfer distance (CD), and Hausdorff distance (HD).
Each shape is
normalized by its own bounding box before scoring.
To calibrate the metrics, we include a prompt-agnostic cube baseline that outputs
the same axis-aligned unit cube for every test case. 
This baseline measures
the score obtainable without using any information from the text; a
meaningful method should clearly outperform it.
\Cref{tab:text2cad-results} shows that \method achieves the best all-case IoU, 
succeeds on all 200 cases, and surpasses the cube baseline on the
largest number of cases. 

\begin{table}[!ht]
\caption{Results on the 200 Text2CAD test cases. Success
requires a generated solid with nonzero volume. CD and HD are averaged
over successful cases only, whereas ``IoU'' assigns zero IoU to failed
cases. Consequently, methods with high failure rates 
can report deceptively low CD and HD.
``Above cube'' counts cases whose IoU exceeds that of the
cube baseline.}
\label{tab:text2cad-results}
\centering
\footnotesize
\setlength{\tabcolsep}{1pt}
\begin{tabular*}{\columnwidth}{@{\extracolsep{\fill}}lccccc@{}}
\toprule
\textbf{System} &
\textbf{Succ. $\uparrow$} &
\textbf{IoU $\uparrow$} &
\textbf{CD $\times 10^3$ $\downarrow$} &
\textbf{HD $\downarrow$} &
\textbf{$>$ Cube $\uparrow$} \\
\midrule
Cube baseline
& \textbf{100.0\%} & 0.127 & 157.6 & 0.512 & -- \\
Multi-Agent-CAD~\citep{mac2026}
& 58.5\% & 0.119 & 44.3 & \textbf{0.308}
& 79/200 \\
text-to-cad~\citep{earthtojake_text_to_cad}
& 90.0\% & 0.172 & 44.1 & 0.316 & 132/200 \\
Atomic API 
& 94.5\% & 0.173 & 47.6 & 0.327 & 142/200 \\
CADDesigner~\citep{DBLP:journals/cad/FanNYWLZTTD26}
& \textbf{100.0\%} & 0.185 & \textbf{43.1} & 0.318
& 154/200 \\
\method
& \textbf{100.0\%} & \textbf{0.186} & 44.2 & 0.320 & \textbf{156/200} \\
\bottomrule
\end{tabular*}
\end{table}


\subsection{Ablation studies}
\label{sec:exp:ablation}

\paragraph{Workflow vs. Atomic API.}
The Atomic API row in \cref{tab:main} replaces our engineer-level workflows
with low-level, atomic SolidWorks API calls; other settings are kept the same.
Across the three tracks, this change
reduces overall pass@1 from 0.78 to 0.21 and score from 95.1 to 41.2. 
Execution traces explain this degradation. Atomic API requires the model to
specify long sequences of coordinates, entity selections, sketch elements,
feature options, and mates without workflow-level semantic references or
local recovery. On the text-to-part track, 21 of 100 runs exhaust the output budget
without producing an executable plan.
The assembly track is more
fragile: all 44 executable plans stop at their first failed operation, in 42
cases because a referenced face cannot be resolved, so no plan reaches the
final save step. Workflows instead package these low-level sequences
behind semantic inputs, effect checks, rollback, and updated observations, allowing
the agent to repair one engineering step before continuing. Consequently,
Atomic API's lower token count in \cref{tab:main} mainly reflects missing or
prematurely terminated trajectories rather than greater efficiency.
The Text2CAD results (\cref{tab:text2cad-results}) also show that
\method significantly outperforms the Atomic API.

\paragraph{Workflow components.}
\Cref{tab:ablation} reports the leave-one-component-out experiments
on all 50 three-view-to-part tasks. 
We identify several key components and features of the workflow, 
each of which corresponds to one of the workflow design principles presented in \cref{tab:workflow-principles}. 
By removing these features while keeping all other parts of the system unchanged, 
we observe a significant performance degradation. 
These results demonstrate the effectiveness of the proposed workflow design principles. 
Detailed descriptions of the ablation studies are in Appendix Section~\ref{sec:suppl:abla}.

\begin{table}[!ht]
\caption{Ablation of workflow components on three-view-to-part tasks
using Qwen3.8-Flash-Next. 
Small red values denote the absolute decrease
in pass@1 relative to the full workflow.}
\label{tab:ablation}
\centering
\small
\setlength{\tabcolsep}{2pt}
\begin{tabular*}{\columnwidth}{@{\extracolsep{\fill}}lcccc@{}}
\toprule
\textbf{Variant} & \textbf{Pass@1} & \textbf{Score} & \textbf{Image} &
\textbf{Tokens} \\
\midrule
Workflow (full) & 0.84 & 96.4 & 97.6 & 136.7k \\
\quad -- canonical $(u,v)$ frames & 0.64\textcolor{red}{\raisebox{-0.4ex}{\scriptsize\,($0.20\downarrow$)}} & 86.4 & 91.6 & 410.6k \\
\quad -- semantic face references & 0.70\textcolor{red}{\raisebox{-0.4ex}{\scriptsize\,($0.14\downarrow$)}} & 91.0 & 95.1 & 138.9k  \\
\quad -- profile pre-checks & 0.76\textcolor{red}{\raisebox{-0.4ex}{\scriptsize\,($0.08\downarrow$)}} & 92.1 & 94.6 & 128.4k \\
\quad -- post-check \& rollback & 0.80\textcolor{red}{\raisebox{-0.4ex}{\scriptsize\,($0.04\downarrow$)}} & 92.1 & 94.4 & 137.4k \\
\quad -- native visual inspection & 0.80\textcolor{red}{\raisebox{-0.4ex}{\scriptsize\,($0.04\downarrow$)}} & 91.7 & 94.7 & 114.5k \\
\bottomrule
\end{tabular*}
\end{table}

\section{Analysis}
\label{sec:analysis}

\paragraph{Failure cases.}
\label{sec:analysis:failures}
We analyze many failure cases of baselines and our system to understand the root cause.
The error type varies with the agent system and the task horizon.
We find that ensuring the plausible motion of assemblies remains a major challenge. 
Appendix Section~\ref{sec:suppl:analysis} provides the details.

\paragraph{Workflow coverage.}
\label{sec:analysis:coverage}
Workflow-CAD obtains broad coverage through 
operation consolidation rather than one
workflow per CAD command. 
This compact interface
therefore covers the main lifecycle of parametric solid and mechanism modeling.
The detailed capability matrix, 
implementation audit and scope boundary are provided in Appendix
Section~\ref{sec:suppl:analysis}.
We further demonstrate the completeness of workflows
through the performance on external benchmarks and Workflow-CAD-Bench in Appendix
Section~\ref{sec:suppl:analysis}.

\section{Conclusion}
\label{sec:conclusion}

To conclude, \method turns human-created feature trees and assembly
relations into engineer-level workflows executed inside a native
modeler. To evaluate the agentic system comprehensively, \bench 
introduces deterministic and fine-grained checks of geometry, structure, and motion. 
Experiments show that workflows substantially improve reliability, especially on complex
assemblies. Overall, our work demonstrates that
LLM-based agents can autonomously translate engineering intent into precise,
editable, and functionally valid CAD models.

%

{
    \small
    \bibliographystyle{ieeenat_fullname}
    \bibliography{main}
}

\clearpage
\onecolumn
\setcounter{page}{1}
\appendix
\addtocontents{toc}{\protect\setcounter{tocdepth}{2}}

\begin{center}
\Large\textbf{\thetitle}\\
\vspace{0.5em}
Appendix
\end{center}
\vspace{1em}

\renewcommand{\contentsname}{Table of Contents in Appendix}
\tableofcontents
\twocolumn
\raggedbottom

\section{Part agent and workflow catalog}
\label{sec:suppl:part-agent}

\subsection{Explanation of workflow design principles}
\label{sec:suppl:impl}
In \cref{sec:method:interface}, we outline six workflow design principles;
below, we describe and explain each principle in detail.

\paragraph{A shared coordinate frame.}
\label{sec:suppl:frames}
For every axis-aligned sketch plane, the workflow interface defines $(u,v)$
from the two world axes lying in that plane: $(X,Y)$ for a plane normal to
$Z$, $(X,Z)$ for one normal to $Y$, and $(Z,Y)$ for one normal to $X$.
The runtime system then applies the sign and axis permutation required by the native
SolidWorks sketch frame. For example, $(u,v)=(20,10)$ on the top plane means
$(X,Z)=(20,10)$\,mm in the workflow frame and is converted to
$(20,-10)$ in SolidWorks' local sketch coordinates. When the agent draws on
an existing flat face, the observation and the drawing workflow use the same
$(u,v)$ frame: each face report includes its $(u,v)$ axes and usable range,
and the runtime converts those coordinates into that face's SolidWorks sketch
frame.

\paragraph{Intuitive feature directions.}
\label{sec:suppl:direction}
SolidWorks asks whether to ``flip'' a feature relative to the selected
plane. Because different planes have different local orientations, the same
Boolean flag does not always mean the same direction in the global $X,Y,Z$
coordinate system. We therefore use intuitive defaults: a boss extends
outward from the selected face, a cut extends into the solid, and
\texttt{direction=both} extends equally on both sides of the sketch plane.
The agent can instead specify a world-axis direction such as \texttt{+X}, as
long as that direction is perpendicular to the sketch plane. It also chooses
how far the feature extends: by a fixed depth, up to the body boundary, or
until a named face. The runtime system converts these choices into the
corresponding SolidWorks API parameters.

\paragraph{Semantic references.}
\label{sec:suppl:references}
The raw SolidWorks API often selects a face by shooting a ray from a 3D
point; a small coordinate error can select the wrong face or none. The
workflow interface instead accepts a semantic or measured description. For
example, \texttt{top} is the outermost planar face on the $+Y$ side, and
\texttt{face(+X, largest)} is the largest planar face whose outward normal
points along $+X$. When several faces remain ambiguous, the agent can use
\texttt{face\#k}, where $k$ is an index from the latest observation. A plane
that is not on the solid can be written by position, such as
\texttt{"Z=40"}; the runtime creates a reference plane there or reuses an
existing one. 

\paragraph{Typed profiles.}
\label{sec:suppl:profiles}
Sketch-driven workflows share one typed 2D profile language rather than a
separate workflow for every observed outline. Seven primitives come from
structural summaries of sketches used by downstream modeling features; we
add an eighth, the outer-tangent \texttt{hull} of two circles, because agents
otherwise had to assemble common link and lug outlines from error-prone
tangent arcs. Each primitive has named dimensions in the canonical $(u,v)$
frame and an explicit open- or closed-region interpretation. Closed
primitives define cross-sections for bosses, cuts, revolves, and lofts,
whereas an open \texttt{path} serves as a rib centerline, sweep trajectory,
or loft guide. Named shapes also accept a uniform or per-corner radius, so
fillets can be drawn directly in the outline instead of added through a
later edge search. \Cref{tab:suppl-profiles} summarizes the complete
arguments, validity rules, and typical uses.

\begin{table*}[t]
\caption{Typed profile primitives shared by every sketch-driven workflow.
Coordinates are in the canonical $(u,v)$ frame of the target plane. Closed
loops may be nested (an outer boundary with holes) but must not partially
overlap. ``Calls'' is the number of sketch-drawing requests that one loop
sends to the atomic-operation service: a non-circular loop is drawn by one
batched polyline or line--arc profile request, however many segments it
has, and all circles of one sketch share a single batched request.}
\label{tab:suppl-profiles}
\centering
\footnotesize
\renewcommand{\arraystretch}{1.12}
\setlength{\tabcolsep}{4pt}
\begin{tabular}{l >{\centering\arraybackslash}p{0.55in}
                >{\raggedright\arraybackslash}p{4.85in}}
\toprule
\textbf{Primitive} & \textbf{Calls} & \textbf{Arguments and role} \\
\midrule
\texttt{circle}
  & 1
  & \texttt{center:[u,v]}, \texttt{diameter} or \texttt{radius}. Disks,
    bores, bosses, and the seed of a hole. \\
\texttt{rect}
  & 1
  & \texttt{center} and \texttt{size:[du,dv]}, or \texttt{corner1} /
    \texttt{corner2}. Optional \texttt{corner\_radius} rounds all four
    corners in the sketch. \\
\texttt{regular\_polygon}
  & 1
  & \texttt{center}, \texttt{sides}, and either \texttt{across\_flats}
    or \texttt{circumradius}, plus \texttt{angle\_deg}. Optional
    \texttt{corner\_radius}. Hex nuts, sockets, and polygonal pads. \\
\texttt{slot}
  & 1
  & \texttt{start}, \texttt{end}, \texttt{width}. A stadium with true
    semicircular ends, used for clearance slots and lightening holes.
    \texttt{start} and \texttt{end} are the centers of the end arcs. \\
\texttt{polygon}
  & 1
  & \texttt{points:[[u,v],...]}. Optional uniform
    \texttt{corner\_radius} or a per-vertex \texttt{corner\_radii}
    list. Tabs, ear plates, and irregular pockets. \\
\texttt{annular\_sector}
  & 1
  & \texttt{center}, \texttt{r\_inner}, \texttt{r\_outer},
    \texttt{start\_deg}, \texttt{end\_deg}, drawn counterclockwise.
    A solid sector when \texttt{r\_inner} is zero. Lightening windows
    and pie cuts. A $360^\circ$ span is normalized to a circle or a
    pair of concentric circles. \\
\texttt{hull}
  & 1
  & \texttt{circles}: exactly two \{\texttt{center}, \texttt{diameter}\}.
    The two circles joined by their outer common tangents: link,
    lever, and lug outlines. Bores are added as separate
    \texttt{circle} loops. Unequal diameters are accepted with a note
    that the straight sides are not parallel. \\
\texttt{path}
  & 1
  & \texttt{start} and \texttt{segments}, each with a required
    \texttt{to:[u,v]}. A segment is a line unless it also has
    \texttt{through:[u,v]} (three-point arc), \texttt{center} with
    optional \texttt{ccw}, \texttt{radius} with \texttt{ccw}, or
    \texttt{tangent}. A center-only $180^\circ$ arc must state
    \texttt{ccw}; a center that misses the equal-radius locus by at
    most 0.2\,mm is snapped onto it. Closed paths are loops; open
    paths are sweep trajectories, loft guides, and rib centerlines. \\
\bottomrule
\end{tabular}
\end{table*}

\paragraph{Pre-execution checks.}
Before a solid-feature profile is sent to SolidWorks, the runtime checks it
as a 2D region. Every boundary must close, enclose a positive area, and not
cross itself. Multiple boundaries may be disjoint (two bosses) or nested (a
hole inside a boss), but they may not partly overlap, because that does not
define an unambiguous material region. A revolve profile must lie entirely
on one side of its rotation axis. For a cut or hole placed on an existing
face, the profile must overlap that face's reported extent. These tests
catch common geometric mistakes before SolidWorks can create a broken sketch
or return an uninformative ``feature failed'' message.

\paragraph{Native visual inspection.}
A visual inspection workflow selects up to four standard views (default: isometric, front,
top, and right). The runtime returns labeled native views with a consistent
display style and world-axis convention; the assembly variant additionally
accepts semantic component roles and highlights the resolved components
(\cref{tab:suppl-tools-ctrl,tab:suppl-assembly-catalog}). 
The composite image
is returned as the next user message to the multimodal LLM, while the
ordinary workflow result retains compact metadata such as the selected views,
bounding box, volume, and unresolved highlight names. A configuration switch
removes the visual schema for a text-only backbone. 


\subsection{Dataset statistics}

\paragraph{CAD-1K.}

CAD-1K contains 114 recorded tasks and 149 delivered parts. Of these, 147
open in SolidWorks 2025 and provide 924 modeling features after excluding
sketches and default planes. We use this collection only to design the
workflow interface, never for evaluation. Three sources of information are
ranked by the number of tasks that use them, not by our tutorial intuition.

\begin{itemize}
\item \textbf{Feature trees.} Among 924 modeling features (excluding sketches
  and default planes), extruded bosses
  (337) and cuts (266) dominate. User-defined reference planes (94) and
  mirrors (40) are the next most frequent constructions that a
  base-plane-only interface cannot express, followed by revolved solids
  (33), ribs (16), chamfers (16), sweeps and lofts (14), Hole Wizard
  holes (12), and shells (2). Linear and circular patterns are rare as
  named features (1) but appear in narrations as hole groups on a pitch
  circle. Extruded bosses and cuts together account for 65.3\% of all
  modeling features.

\item \textbf{Sketches used by downstream modeling features.} The most common
  sketch signature is a mixed line--arc outline
  (\texttt{PLANE}$\rightarrow$\texttt{EXT[mixed]} in 17 tasks). Slots,
  rounded rectangles, and tabs with filleted corners are therefore
  first-class profile primitives rather than a sharp outline plus a
  later fillet. Open paths appear as rib centerlines and sweep
  trajectories.
\item \textbf{Narration and rubric.} These records reveal design choices that
  are not always clear from the final feature tree. Beyond a simple extrusion
  with a fixed depth, engineers use mid-plane (50 tasks), through-all (46),
  up-to-surface (32), two-direction (17), and thin-wall (11) options. They
  also edit sketches or features in 84 tasks and undo or delete operations in
  96 tasks, showing that correction is a routine part of modeling. The task
  rubrics explicitly require through-holes in 82 tasks, counterbores in 9,
  threads in 5, and ribs, gussets, or webs in 45. Sheet-metal base flanges
  and cosmetic threads appear less often (about 5 and 7 tasks), but each
  represents a distinct modeling need; we therefore provide dedicated
  workflows for them instead of leaving them to the raw API.
\end{itemize}

\paragraph{Fusion 360 Gallery.}
The reconstruction subset of Fusion 360
Gallery~\citep{DBLP:journals/tog/WillisPLCDLSM21} contains 8,625 parts
(6,900 train and 1,725 test designs), each with a complete feature tree of
21,557 sketches and 19,333 extrusions in total. Its authors restricted the
subset to sketch and extrude operations, so it cannot rank feature families
as CAD-1K does. Instead, its scale quantifies how the dominant
sketch--extrude pattern is parameterized and which profiles it consumes.

\begin{itemize}
\item \textbf{Feature trees.} Trees are short: a part has 2.24 extrusions
  on average (median 2, maximum 38), 48.7\% of parts use a single extrusion,
  and 83.9\% use at most three. Of all extrusions, 53.3\% create a new body,
  24.4\% join and 22.2\% cut material, and only 19 intersect; cuts appear
  in 27.6\% of parts, and 11.5\% of parts end with several bodies. End
  conditions mirror CAD-1K: besides one-sided extrusions (87.7\%), engineers
  use symmetric (mid-plane, 9.4\%, in 13.9\% of parts) and two-sided (2.9\%)
  extents, and rarely a draft angle (169 extrusions) or an offset start
  (53). These options remain parameters of one \texttt{extrude} workflow.
  Moreover, 23.9\% of extrusions consume several profiles of one sketch at
  once, which motivates batched multi-loop profiles.

\item \textbf{Sketches used by downstream modeling features.} Of 17,370
  sketches consumed by an extrusion (19.4\% of all sketches are never
  extruded), 56.8\% lie on a base plane (XZ 4,682, XY 3,173, YZ 2,006) and
  42.9\% lie on an existing planar face of the solid; user-defined
  construction planes are rare (14). The shared $(u,v)$ frame for both
  base planes and faces therefore covers nearly all sketch placements. The
  curve sets are circles only (31.0\%), lines only (26.2\%), mixed
  line--arc (22.6\%), lines and circles (16.8\%), and freeform splines or
  ellipses (3.4\%). We further classify the 37,751 loops of extruded
  regions: circles (44.4\%), rectangles (17.1\%), general polygons
  (8.8\%), annular sectors (4.2\%), two-circle hulls (1.7\%), regular
  polygons (1.5\%), slots (1.1\%), and rounded rectangles (1.0\%), with
  other line--arc loops (16.7\%) expressible as closed \texttt{path}s. The
  eight typed primitives thus represent 96.4\% of extruded loops; only
  freeform loops (3.6\%) fall outside. In addition, 18.8\% of extruded
  regions contain at least one inner loop, so nested loops must be
  supported. Hole groups again appear in sketches rather than as pattern
  features: 723 sketches contain three or more equal circles, forming 355
  groups on a pitch circle and 87 equally spaced rows.

\item \textbf{Constraints and dimensions.} Fusion stores the sketch
  constraints and dimensions that CAD-1K narrations only describe. Of
  107,450 constraints, coincidences account for 28.1\%, horizontal,
  vertical, perpendicular, and parallel relations for 42.5\%, and
  tangencies for 12.3\%; symmetry (4,835) and concentricity (5,989) are
  also common. Of 39,663 dimensions, 42.8\% are diameters or radii.
  These relations are exactly those built into named primitives such as
  \texttt{rect}, \texttt{slot}, and \texttt{hull}, whereas only 38.1\% of
  extruded sketches are fully constrained in the source models. Typed
  primitives with named dimensions therefore state design intent that
  hand-drawn outlines often leave implicit.
\end{itemize}

\subsection{Part workflow extraction details}

As described in the main text, we first normalize implementation-level feature names into operation
families with shared engineering intent. Variants with the same geometry and
validation rules become options of one candidate workflow. For example, a closed sketch
followed by a boss extrusion becomes \texttt{extrude}, while blind,
mid-plane, and up-to-surface remain end-condition options. In contrast,
operations with different intent or validity rules remain separate even
when SolidWorks stores them in one feature family. Hole Wizard therefore
yields \texttt{hole}, \texttt{counterbore}, \texttt{countersink}, and
\texttt{tapped\_hole} (\cref{tab:suppl-tools-hole}).

The narration and rubrics also reveal how engineers validate and repair
intermediate results: they measure geometry, inspect feature state, and edit
or delete failed features. We retain a geometry-changing candidate only when
its inputs admit a reusable typed schema, its validity can be checked before
execution, and its geometric effect can be measured afterward with rollback
on failure. The observed frequencies determine which operation families are
represented as dedicated workflows.

The \textbf{8 profile primitives} in
\cref{tab:suppl-profiles} are the shared sketch language those workflows
consume: seven come from the sketch statistics above, and the two-circle
\texttt{hull} was added for the link and lug outlines that agents
otherwise had to assemble from tangent arcs. We count a workflow as extracted when it
is a first-class workflow (or a first-class profile primitive) with a
typed schema and an explicit success condition. Geometry-changing workflows
also require a pre-check and rollback.
The interface provides reusable operations rather than a separate workflow
for every recurring shape. Professional models contain tubes, branch bosses,
hole groups, stepped revolves, gussets, mirrored half-bodies, and pipe
elbows. A hole group is expressed by the \texttt{polar} layout of a hole
workflow, while a pipe elbow uses \texttt{sweep} with a rounded 3D path and
a hollow circular profile. The remaining patterns require a short sequence
of existing workflows, such as constructing half a model and then mirroring
it. Thus, these structures are supported by the 25 part workflows but are
not counted as additional extracted workflows.
\label{sec:suppl:not-extracted}

\subsection{Part workflow catalog}
\label{sec:suppl:catalog}
\label{sec:suppl:atomic-calls}

This section lists every part workflow in three groups:
control, reference, editing, and finishing workflows
(\cref{tab:suppl-tools-ctrl}); sketch-driven solid workflows
(\cref{tab:suppl-tools-solid}); and hole, pattern, and symmetry workflows
(\cref{tab:suppl-tools-hole}).
The ``Calls'' column reports how many top-level requests one successful
workflow sends to the C\# atomic-operation service. Selecting a reference,
opening or drawing a sketch, creating or deleting a feature, and reading the
model state each count as one request. A batched operation, such as drawing
all circles in one sketch, also counts as one. Failed-call rollback and
fallback requests are excluded, whereas the state reads before and after a
feature-creating workflow are included because they verify its effect. For a
single-loop profile, the 25 part workflows range from 0 calls for
\texttt{set\_parameters} to 12 for \texttt{counterbore}, with a median of 5;
the 21 workflows that modify the feature tree require 3--12 calls.

Units in workflows are millimeters and degrees. Every numeric
argument may be a literal or an expression over named parameters set by
\texttt{set\_parameters}. For example, after setting \texttt{W=100}, the
expression \texttt{=W/2-8} evaluates to 42\,mm while preserving \texttt{W}
for reuse by later features.

\subsection{Part-agent system prompt}
\label{sec:suppl:part-prompt}

Listing~\ref{lst:part-agent-prompt} gives the principal system prompt of the
part agent. The task statement and
the typed tool schemas are supplied separately at run time. The prompt is
task-agnostic: it defines the coordinate convention, construction and
verification policy, and recovery behavior, but contains no
benchmark-specific geometry or reference solution.

\makeatletter
\setlength{\@fptop}{0pt}
\setlength{\@dblfptop}{0pt}
\makeatother
\clearpage


\begin{table*}[t]
\caption{Control, reference, editing, finishing, and sheet-metal workflows
(12 of 26 callable part workflows; 25 are available when visual inspection
is disabled). $k$ is the number of deleted features, $S$ is defined in
\cref{tab:suppl-tools-solid}, and $E$ is the number of separate non-ray edge
selections.}
\label{tab:suppl-tools-ctrl}
\label{tab:suppl-tools-rest}
\centering
\footnotesize
\renewcommand{\arraystretch}{1.12}
\setlength{\tabcolsep}{4pt}
\begin{tabular}{l >{\centering\arraybackslash}p{0.55in}
                >{\raggedright\arraybackslash}p{4.85in}}
\toprule
\textbf{Workflow} & \textbf{Calls} & \textbf{Arguments, checks, and professional evidence} \\
\midrule
\texttt{get\_model\_state}
  & 1
  & Optional filters \texttt{type} (plane or cylinder), \texttt{axis},
    \texttt{diameter} ($\pm$0.05\,mm), and \texttt{near:[x,y,z]} with a
    search \texttt{radius}. Returns user features, axis-aligned bounding
    box, volume, solid-body count, planar faces with a \texttt{face\#k}
    index, outward normal, offset, $(u,v)$ extent and area, cylindrical
    faces grouped by diameter, axis, and span with the position of every
    axis in the group, reference planes, and the current parameter table.
    One batched observation request reads up to 1{,}000 faces. This is
    the observation that closes the loop. Narrations spend more time
    on measure / mass-properties checks (83 tasks) than on the
    extrusion command itself. \\
\texttt{view\_model}
  & 2
  & Optional \texttt{views} (at most four; default is isometric, front, top,
    and right). Captures native SolidWorks shaded-with-edges views, hides
    reference geometry and shadows, zooms to fit, and restores the prior
    view. The runtime packs the captures into a two-column PNG whose labels
    state the view and world-axis directions. The image is returned to the
    multimodal LLM; metadata retains the views, bounding box, and volume.
    One state read plus one batched capture request. Read-only and disabled
    for text-only backbones. \\
\texttt{set\_parameters}
  & 0
  & \texttt{values}: a map from name to a number or an expression.
    Later workflow arguments may write \texttt{=W/2-8}. Parameters are
    held by the runtime and evaluated before any request is sent, so
    they survive into a later edit. This
    replaces the fully-defined sketching that dominates professional
    time (constraints and dimensions in every one of the 114 tasks). \\
\texttt{delete\_feature}
  & $2{+}4k$
  & \texttt{name}, or \texttt{names} deleted in the given order, from the
    latest state. Deletes each feature and the sketch it consumed;
    dependents go with it and are reported as \texttt{also\_removed}
    together with the remaining feature list. Undo / delete appears
    in 96 professional tasks. The workflow is for retracting a bad step,
    not for changing a dimension (that is \texttt{edit\_feature}). \\
\texttt{finish}
  & 1
  & \texttt{summary}: a short free-text note. Refused while no solid
    body exists or while the part has more than one solid body. The
    agent is prompted to check the envelope, the number and diameters
    of holes and bosses, and the single-body condition before calling
    it. \\
\texttt{ref\_plane}
  & 3
  & \texttt{reference} (base plane, existing user plane, or planar
    face) and signed \texttt{offset} along the reference normal.
    Returns a name that is a legal \texttt{plane} for every
    sketch-driven workflow. A plane at the same position and normal is
    reused (1 call); a plane created on the wrong side is deleted and
    rebuilt flipped (5 calls). Sketch-driven workflows also accept the
    shorthand \texttt{"Z=40"}, which creates or reuses such a plane.
    User planes are the most frequent professional feature that default
    planes cannot replace (94 instances, 48 tasks), and the transition
    \texttt{PLANE}$\rightarrow$\texttt{EXT} occurs in 35 tasks.
    Angled and through-point planes are not yet exposed. \\
\texttt{edit\_feature}
  & 3
  & \texttt{name}, \texttt{value} (mm), optional \texttt{dimension}
    (defaults to the first driving dimension of that feature).
    Changes depth, radius, or thickness and rebuilds. Edit-sketch /
    edit-feature narrations occur in 84 tasks. Profile features
    (sketches used by downstream modeling features) are skipped so that
    the solid feature is edited, not its sketch wrapper. Reference
    planes are not edited; the agent writes a new \texttt{"Z=c"} plane
    instead. \\
\texttt{fillet}
  & 3 ($E{+}2$)
  & \texttt{radius}; \texttt{edges} is \texttt{all}, a named box edge
    such as \texttt{top\_front}, a probe ray
    \{\texttt{origin}, \texttt{direction}\} that selects the nearest
    edge on the hit face, or an array of rays placed in one feature;
    optional \texttt{propagate} (default true) along tangent edges.
    Rays that graze a face or hit a face corner are flagged as
    ambiguous, and the result estimates the filleted edge length from
    the volume change. Ninety-three fillet features in 41 tasks;
    a finishing sequence of fillets and chamfers closes 12 tasks. \\
\texttt{chamfer}
  & 3 ($E{+}2$)
  & \texttt{distance}; \texttt{edges} as for \texttt{fillet}. Sixteen
    chamfer features in 11 tasks. \\
\texttt{draft}
  & 3
  & \texttt{neutral} plane, \texttt{faces} to taper,
    \texttt{angle\_deg}. Narrations mention draft in 9 tasks. Used
    after a prismatic boss when a molded or cast face must release. \\
\texttt{shell}
  & 3
  & \texttt{thickness}, \texttt{open\_faces} (semantic face list);
    optional \texttt{outward}. Removes the listed faces and leaves a
    uniform wall. Two shell features in 2 tasks; shell narrations in
    16 tasks. Typical of open boxes and housings. \\
\texttt{sheet\_metal}
  & $S{+}4$
  & \texttt{plane}, closed \texttt{loops}, \texttt{thickness}. Legal
    only as the first feature of an empty part. Converts the outline
    to a sheet-metal base flange. About 5 professional tasks are
    sheet-metal families. Edge flanges, bends, and corner breaks are
    not yet workflows. \\
\bottomrule
\end{tabular}
\end{table*}

\begin{table*}[t]
\caption{Sketch-driven solid workflows (7 of 26). $S$ is the number of
sketch-drawing requests of a profile (\cref{tab:suppl-profiles}); $S{=}1$
for a single loop or for circles only. A sketch-driven call reads the state,
opens the sketch, draws, issues one feature request, and reads the state
again for the effect check. For \texttt{loft}, $n$ is the number of
sections. A plane written as \texttt{"Z=c"} adds three calls when it
creates a new reference plane and none when an existing one is reused.}
\label{tab:suppl-tools-solid}
\centering
\footnotesize
\renewcommand{\arraystretch}{1.12}
\setlength{\tabcolsep}{4pt}
\begin{tabular}{l >{\centering\arraybackslash}p{0.55in}
                >{\raggedright\arraybackslash}p{4.85in}}
\toprule
\textbf{Workflow} & \textbf{Calls} & \textbf{Arguments, checks, and professional evidence} \\
\midrule
\texttt{extrude}
  & $S{+}4$
  & \texttt{plane}, \texttt{loops}; then exactly one of \texttt{depth}
    or \texttt{up\_to} (a face reference or \texttt{body}). Optional
    \texttt{direction} among world axes or \texttt{both} (mid-plane,
    \texttt{depth} is the total thickness), and \texttt{draft\_deg} /
    \texttt{draft\_outward}. Nested loops
    become holes in the boss; partial overlaps are rejected. A
    successful call must increase volume and keep a single solid body;
    a boss that only touches the part along a line or point is rolled
    back. The result reports the feature's world extent, the fill ratio
    (volume added over profile area $\times$ depth), and new cylindrical
    faces. Extruded bosses are the most
    common professional feature (337 instances, 101 tasks). Up-to and
    mid-plane match the end-condition narrations (32 and 50 tasks). \\
\texttt{cut}
  & $S{+}4$
  & Same plane and loop language as \texttt{extrude}. End condition is
    \texttt{depth}, \texttt{through\_all}, or \texttt{up\_to}. Default
    direction is into the solid on a face; on a base plane it is toward
    the body, or both ways when the plane passes through the body. A
    successful call must decrease volume without splitting the part.
    The result reports world extent, removal ratio, and new or removed
    cylinders, and warns when a one-sided through-all cut starts inside
    the part. Cuts are the
    second most common feature (266 instances, 86 tasks). Through-all
    appears in 46 narrated tasks. \\
\texttt{revolve}
  & $S{+}5$
  & \texttt{plane}, \texttt{loops}, \texttt{axis\_start},
    \texttt{axis\_end} in $(u,v)$; optional \texttt{angle\_deg}
    (default 360) and \texttt{cut}. The profile must lie entirely on
    one side of the axis, and may touch it; the axis is one extra
    centerline request. Used for shafts, bosses,
    and revolved pockets (33 features, 18 tasks). \\
\texttt{rib}
  & 5
  & \texttt{plane}, open segment \texttt{start}/\texttt{end},
    \texttt{thickness}, optional \texttt{both\_sides} (default true).
    The segment is expected to bridge two existing faces, typically
    on an offset reference plane. Sixteen rib features occur in 13
    tasks, 12 of which create a plane first; the rubric mentions
    rib / gusset / web in 45 tasks. \\
\texttt{sweep}
  & 4--$(S{+}8)$
  & Optional \texttt{cut}. The path is one of a planar open path
    (\texttt{path\_plane}, \texttt{path\_start}, \texttt{path\_segments}),
    a 3D polyline \texttt{path\_3d} with optional
    \texttt{corner\_radius} at every bend, or a \texttt{helix} (axis,
    start center, diameter, pitch, turns or height, handedness). The
    section is either a circle \texttt{profile\_diameter} with optional
    \texttt{inner\_diameter} for a hollow tube, valid along any path, or
    closed \texttt{loops}. When \texttt{profile\_plane} is omitted the
    section plane is placed at the path start, normal to the start
    tangent, and loops are given relative to the path start. A circular
    section on a 3D path costs 4 calls; a planar path with an explicit
    section plane costs $S{+}7$. Circular sections also return path
    length and nominal volume. Covers pipe elbows, springs, threads,
    and swept cuts (sweeps and lofts together: 14 features,
    12 tasks; sweep narrations in 12 tasks). \\
\texttt{loft}
  & $\sum_i S_i$\newline${+}\,2n{+}2$
  & \texttt{sections}: an ordered list of
    \{\texttt{plane}, \texttt{loops}\}; the first or last section may be
    \{\texttt{plane}, \texttt{point}\} for a pointed end. Optional
    \texttt{cut}, \texttt{guides} (planar open paths or 3D polylines and
    splines that touch every section), and \texttt{start\_tangent} /
    \texttt{end\_tangent} \texttt{normal} with \texttt{tangent\_length}
    for a smooth blend into an adjacent extrusion. Each section is an
    independent typed profile. Guides, point sections, or end tangency
    add one call plus three per planar guide (one per 3D guide); a
    tangent loft that overshoots its end sections is retried with the
    other tangent orientation. Used for
    square-to-round adapters and other blends (loft narrations in
    7 tasks). \\
\texttt{axis\_cylinder}
  & $\geq5$
  & \texttt{origin}, arbitrary 3D \texttt{direction}, axial
    \texttt{from}/\texttt{to}, and \texttt{diameter}; optional
    \texttt{inner\_diameter}, \texttt{cut}, and drill
    \texttt{tip\_angle}. It reduces an oblique or radial cylinder to a
    circular sweep over a 3D line, then applies the same checks and rollback
    as \texttt{sweep}. This general extension covers radial bosses, hollow
    nozzles, and oblique holes without one workflow per orientation. \\
\bottomrule
\end{tabular}
\end{table*}

\begin{table*}[t]
\caption{Hole, pattern, and symmetry workflows (7 of 26). All centers of one
hole call are drawn by one batched circle request, so the
count does not grow with the number of holes. The two-stage hole workflows
are two complete cut transactions plus a state read before and after.}
\label{tab:suppl-tools-hole}
\centering
\footnotesize
\renewcommand{\arraystretch}{1.12}
\setlength{\tabcolsep}{4pt}
\begin{tabular}{l >{\centering\arraybackslash}p{0.55in}
                >{\raggedright\arraybackslash}p{4.78in}}
\toprule
\textbf{Workflow} & \textbf{Calls} & \textbf{Arguments, checks, and professional evidence} \\
\midrule
\texttt{hole}
  & 5
  & \texttt{plane}, \texttt{diameter}, and \texttt{centers:[[u,v],...]}
    and/or a \texttt{polar} layout \{\texttt{center}, \texttt{radius} or
    \texttt{pcd}, \texttt{count}, \texttt{start\_deg}, \texttt{span\_deg}\}
    whose generated centers are echoed in the result;
    \texttt{depth} or \texttt{through\_all}; optional
    \texttt{direction}. Equal-diameter circular cuts. Through-holes
    appear in 82 rubric tasks. A pitch-circle group is a \texttt{polar}
    argument, not a separate \texttt{hole\_group} workflow. \\
\texttt{counterbore}
  & 12
  & \texttt{plane}, \texttt{centers} or \texttt{polar},
    \texttt{through\_diameter}, \texttt{head\_diameter},
    \texttt{head\_depth}; \texttt{depth} or
    \texttt{through\_all}. Socket-head cap-screw seats. If the through
    stage fails, the head stage is rolled back. The rubric
    marks counterbores in 9 tasks; stacking two cuts is possible but
    loses the Hole Wizard intent. \\
\texttt{countersink}
  & 11
  & \texttt{plane}, \texttt{centers} or \texttt{polar},
    \texttt{through\_diameter}, \texttt{head\_diameter}, and the
    included cone \texttt{angle\_deg} (default $90^\circ$; derived from
    \texttt{head\_depth} when only that is given). A drafted blind cut
    forms a true conical seat whose depth follows from the diameters
    and angle; its removed volume is checked against the frustum
    volume before the through hole is cut. \\
\texttt{tapped\_hole}
  & 5 / 7
  & \texttt{plane}, \texttt{centers} or \texttt{polar}, tap-drill
    \texttt{diameter}; \texttt{depth} or \texttt{through\_all}; optional
    \texttt{thread\_diameter} adds a cosmetic thread (7 calls). Threads
    appear in 5 rubric tasks and as \texttt{SweepThread} features in 7
    tasks. \\
\texttt{circular\_pattern}
  & 3--11
  & \texttt{count}, \texttt{axis} $\in \{X,Y,Z\}$; optional
    \texttt{angle\_deg} and \texttt{center:[x,y,z]}. With
    \texttt{center}, the runtime builds (or reuses) a reference axis
    as the intersection of two axis-aligned planes through that point
    (5 calls for an axis through the origin); otherwise the unique
    cylindrical face supplies the axis (3 calls). Instances the latest
    feature, including the seed, and compares the volume change with
    the seed's change $\times(\texttt{count}{-}1)$. \\
\texttt{linear\_pattern}
  & 3
  & \texttt{count1}, \texttt{spacing1}, \texttt{direction1}
    (e.g.\ \texttt{+X}); optional second direction. Instances the
    latest feature on a one- or two-direction grid and applies the
    same seed-volume check. \\
\texttt{mirror}
  & 3
  & \texttt{plane}; either \texttt{features:[name,...]} or
    \texttt{body=true}. Mirrors about a base plane, a user plane, or
    a planar face and merges the result. Forty mirror features occur
    in 22 tasks; narrations mention mirror in 47 tasks. The common
    ``model half, then mirror'' motif (10 tasks after an extrusion)
    is this workflow plus a preceding half-body. \\
\bottomrule
\end{tabular}
\end{table*}

\onecolumn

\begin{nolinenumbers}
\begin{lstlisting}[
  basicstyle=\ttfamily\footnotesize,
  caption={Principal system prompt of the part agent.},
  label={lst:part-agent-prompt}
]
You are a senior mechanical design engineer. Build the part described by the user step by step in SolidWorks using tools.

COORDINATES AND UNITS
- All lengths are in mm and angles are in degrees. SolidWorks base planes are front_plane = XY (normal +Z), top_plane = XZ (normal +Y), and right_plane = YZ (normal +X).
- If the request specifies a coordinate system or coordinates, follow them exactly; do not swap axes or translate the part. Otherwise, use +Y as up, place the bottom at Y=0, center the part in X and Z, and align solids of revolution with Y.
- Every sketch plane has fixed (u,v) coordinates in model coordinates: normal Z -> (X,Y), normal Y -> (X,Z), and normal X -> (Z,Y).
- Prefer plane references such as "Z=10", "Y=-12.5", or "X=0". A reference such as face(+Z, outermost) denotes a semantic planar face. The aliases top/bottom/front/back/left/right always denote +/-Y, +/-Z, and +/-X. face#k is valid only for the latest get_model_state result.

SKETCH AND FEATURE RULES
- One extrude or cut describes one region: an outer loop and any inner loops. Partially overlapping loops are invalid; split them into features or use one polygon/path.
- The result must be one solid body. Every new boss must share a face with, or overlap by at least 0.5 mm with, the body. Line, point, or tangent contact is disconnected and rejected.
- Base planes and "axis=value" planes extrude along their positive normal by default; solid faces extrude outward. With direction=both, depth is the total two-sided depth.
- Cuts default toward the body. A one-sided through_all cuts only one side of the sketch plane. To cross material on both sides, sketch on an end face or use direction=both.
- Prefer one polygon/path extrusion for constant-section L, U, T, and stepped profiles over assembling multiple blocks.
- Put outline rounds directly in rect/polygon corner_radius, slot, hull, annular_sector, or path arcs. Reserve fillet and chamfer for edges that cannot be represented in the profile, and apply them near the end.
- Define a semicircular path arc with a point on the arc. A 180-degree center arc must state ccw. Use hull for the outer tangents of two circles; two intersecting circular cavities require two cuts or a path made from major arcs.
- A hull with unequal circle diameters has nonparallel sides. Use rect or equal-circle hull for a constant-width handle.
- slot start/end are the centers of the end arcs. Account for width/2 when checking the clear opening. Model spokes as separate rectangular or circular members between hub and rim; do not use a wide slot through the axis.
- Use hole.polar for equally spaced holes. Angles run from +u toward +v. Use circular_pattern for other circumferential features and always provide its center; use linear_pattern for linear repetition.
- When extending a cutting profile past an outer boundary, extend along the original boundary direction without moving a dimensioned location.
- A revolve profile is one half-section and must lie entirely on one side of its axis. Add concave junction fillets after the sharp revolve rather than duplicating them in the profile.
- Join side bosses and branch pipes with extrude(..., up_to=face or body), rather than guessing depth.
- Use counterbore/countersink for headed holes, tapped_hole for threaded pilot holes, rib for gussets, sweep for bent tubes, loft for changing sections, shell for hollow enclosures, draft for taper, and sheet_metal for sheet parts.
- For an annular handle on a cup or tube, place the ring center outside the wall and keep its inner diameter out of the cavity.
- For a radial hole that ends at another bore, set depth to the distance from the outer surface to the near bore wall (at most 1 mm extra). Do not use through_all or cut the far wall.
- For a blind drilled hole with a specified tip angle, pass tip_angle to hole/tapped_hole; depth is the full-diameter depth. Do not construct the tip manually.
- Pass inch drill and thread designations directly as diameter values so that the tool can use its lookup table.
- Use axis_cylinder for cylinders and holes whose axes are not coordinate-aligned. Give an axis point, a direction vector (or latitude/longitude or a target point), and axial bounds. Reuse the exact origin/direction for a nozzle boss and bore.
- For two plates joined by posts, first span the plate gap with the posts (or use one sectional feature), then add the second plate at its specified thickness. Do not thicken a plate only to force connectivity.
- Adjacent intervals in the request meet end-to-end. Start each segment at the preceding segment's end; do not embed the full length of one segment in another.
- Create all bosses that occupy a location before cutting its holes, cavities, and slots; a later boss can refill a prior cut.
- For a socket-head bolt, create shank then head, cut the socket from the head end, and add the shoulder fillet last.
- For sweep, normally omit profile_plane. A circular tube, wire, or spring uses profile_diameter (and inner_diameter for a tube). Use path_3d with corner radii for spatial pipes and a helix, including piecewise pitch, for springs or helical cuts.
- For formula-defined curves (involutes, scrolls, cam profiles, airfoils, cycloids, and variable-pitch/noncircular helices), pass a parametric curve expression instead of manually listing many points. Trigonometric functions use radians; sind/cosd use degrees. Use offset for a planar equidistant curve.
- Loft endpoints may collapse to a point. Use guides to control intermediate shape and normal start/end tangency for smooth joins.
- Change an existing dimension with edit_feature. Remember that delete_feature also removes dependent later features.

CHECK AFTER EVERY STEP
- Read volume_delta_mm3, bbox_mm, extent_mm, fill/removal ratio, and new_cylinders from every result and compare them with the expected effect. Resolve warnings before continuing.
- For fillets/chamfers, compare estimated_edge_length_mm with the expected edge length. A concave fillet should add volume and an outer-edge fillet should remove volume; the wrong sign indicates selection of the adjacent edge.
- For a pattern, compare actual volume change with the reported seed-based expected_delta_mm3.

WORKING PROCEDURE
1. Parse the request into overall dimensions, feature positions, and counts. Optionally define driving values with set_parameters and reuse them as "=expression". A stated spacing, eccentricity, or center distance is center-to-center; do not divide it by two. Resolve angular relations explicitly.
2. Build the base body first, then add features. Independent tool calls may be issued in parallel in one response.
3. A failed tool call is rolled back. Read its error code and correction hint, then change the parameters or strategy. The same failed call will not execute a third time.
4. Track llm_calls_left and reserve calls for verification and finish.
5. After modeling, call get_model_state. Verify overall size, cylinder groups (hole/boss counts and diameters), and the single-solid condition. Continue editing if anything differs.
6. Call finish only after verification, summarizing the part and its key dimensions.
\end{lstlisting}
\end{nolinenumbers}

\twocolumn

\section{Assembly agent and workflow catalog}
\label{sec:suppl:assembly-agent}

\subsection{Dataset statistics}

\paragraph{AutoMate.}
AutoMate~\citep{jones2021automate} collects public Onshape documents with
1,292,016 mates over 451,967 parts. After removing 25,140 top-level
assemblies with duplicate mate signatures, 97,817 assemblies remain; 66.0\%
contain at least one movable joint connected to the ground, and the other
34.0\% are static. In Onshape a mate is a joint with a fixed set of
retained degrees of freedom, whereas SolidWorks usually needs one to three
standard mates for the same joint. AutoMate stores mates as an unordered
graph, so it supplies mate types, graph topology, and couplings, but not
construction order.

\begin{itemize}
\item \textbf{Mate types and component pairs.} Fastened mates dominate
  (54.2\%), followed by planar (22.1\%), revolute (9.8\%), cylindrical
  (7.2\%), slider (4.0\%), parallel (2.0\%), ball (0.5\%), and pin--slot
  (0.3\%) mates. Part shape alone does not predict the mate: after
  classifying parts by bounding box as rods, discs, toothed discs, plates,
  blocks, or compact parts, every pair without a toothed disc has 48.9--58.6\%
  fastened and 5.7--10.7\% revolute mates, so the joint must be chosen from
  the concrete interface instead. One mate connects 91.3\% of component
  pairs, two mates 6.1\%, and three 2.5\%. The multi-mate combinations are
  mostly native expansions of one joint, such as cylindrical plus planar
  (21.4\%, a revolute joint) and three planar mates (21.0\%, a fixed
  joint), or redundant ones, such as planar plus slider (8.7\%). This
  motivates choosing one semantic joint per pair and expanding it into a
  non-redundant SolidWorks mate set.

\item \textbf{Ground and topology.} Only 44.5\% of assemblies fix any
  component explicitly. When exactly one is fixed, it is the largest part
  by volume in 51.1\% of cases and the part with the most mates in 42.2\%,
  and blocks and plates account for 56.9\% of fixed parts. Rooted at the
  ground, 43.5\%, 23.2\%, and 13.5\% of movable bodies lie one, two, and
  three joints away, so three levels cover 80.2\%. The joint that attaches
  a moving body directly to the ground is revolute in 29.4\% of cases,
  planar in 17.6\%, slider in 14.6\%, and cylindrical in 8.4\%. Most
  assemblies (78.3\%) have no independent loop. Loops of length three and
  four account for 44.7\% and 25.3\% of all loops, but most of them are
  planar-mate layout constraints; classic mechanisms are a small minority,
  with four-bar (four revolute) loops at 3.1\% and crank--slider loops at
  0.7\%. Mate order in the stored graph carries no construction signal:
  the fraction of mates that attach to the already connected set is 0.828
  in file order and 0.828 after random shuffling.

\item \textbf{Couplings and mobility.} Only 4.6\% of assemblies contain a
  mate relation, but the types are concentrated: gear 64.1\%, screw
  19.4\%, rack--pinion 10.8\%, and linear 5.8\%. Their supporting joints
  are equally regular: 93.4\% of relations are a gear on two revolute
  joints, a screw on one cylindrical joint, a rack--pinion on a revolute
  joint and a slider, or a linear coupling on two sliders. A Gr\"ubler
  estimate of the ground-connected mobility $M$ gives $M=1$ in 21.2\% of
  assemblies, $2\leq M\leq5$ in 31.0\%, $M\geq6$ (floating components) in
  16.7\%, and $M\leq-3$ (strong redundancy) in 20.9\%. Only 41.3\% of mated
  toothed discs can actually move relative to the ground; the rest are
  fastened or floating. Existing assemblies therefore often do not encode
  the intended motion, so mobility and motion must be checked explicitly
  rather than learned from the data.
\end{itemize}

\paragraph{Fusion 360 Gallery.}
The assembly subset of Fusion 360
Gallery~\citep{DBLP:journals/tog/WillisPLCDLSM21} contains 8,251
assemblies with a median of 8 bodies. Of these, 1,684 contain joints:
12,970 joints and 2,840 as-built joints in total, with a median of 4
joints and 11 component occurrences per jointed assembly. In 1,049 of them
at least one movable joint is connected to the ground. We map Fusion joint
types to the Onshape types above, treat a joint to the root component as
grounding, and treat a nested component that takes part in no joint as
rigidly attached to its parent. Unlike AutoMate, every joint and component
carries a timeline index, and each joint records the geometry on both
sides. Fusion is smaller, so we use it to cross-validate AutoMate and to
add the order and interface signals that AutoMate lacks.

\begin{itemize}
\item \textbf{Joints and topology.} Fastened joints again dominate
  (59.7\%), but motion joints are more common than in AutoMate: revolute
  28.1\%, slider 6.4\%, and cylindrical 4.5\%. Planar joints are rare
  (0.75\%), which confirms that AutoMate's planar mates are mostly an
  Onshape layout habit rather than a general joint type. Grounding shows
  the same pattern as AutoMate: 42.0\% of jointed assemblies ground a
  component explicitly and 12.7\% attach a joint to the root, while 45.3\%
  give no ground. A single grounded component is the largest part in
  56.6\% of cases and the first inserted component in 61.6\%. The joint
  graphs are even shallower: 90.9\% of movable bodies lie within three
  joints of the ground, the joint to the ground is revolute in 62.1\% of
  cases and slider in 25.0\%, and 91.1\% of assemblies have no
  independent loop. Loops are mostly real mechanisms: four revolute
  joints (21.0\%), three revolute joints and a slider (13.5\%), and its
  variant with one cylindrical joint (7.0\%). The mobility estimate gives
  $M=1$ in 33.0\% of assemblies, $M\geq6$ in 20.8\%, and $M\leq-3$ in only
  3.7\%, because Fusion has few stacked planar mates.

\item \textbf{Construction order.} In the 973 assemblies with at least
  three timestamped joints, 85.2\% of joints attach a new component to the
  already placed subassembly, compared with 76.2\% after randomly
  shuffling the joint order within each assembly. The 271 cycle-closing
  joints sit at a mean normalized timeline position of 0.687, compared
  with 0.488 for spanning-tree joints. The mean position of every joint
  type lies between 0.46 and 0.54, except planar joints (0.64). Designers
  therefore do not create all fixed joints before the motion joints;
  they place and connect one component at a time, and close loops last.

\item \textbf{Joint interfaces and limits.} The referenced geometry
  depends strongly on joint type. Revolute joints most often align the
  centers of two circular edges (34.8\%), which in SolidWorks corresponds
  to a concentric mate between the cylindrical faces plus a coincident
  mate between the end faces. Sliders use either two circular edges
  (24.3\%, a shaft in a sleeve) or two planar faces (24.3\%, a guide), ball
  joints align two sphere centers (59.5\%), and fastened joints use circle
  centers (29.4\%) or planar faces (17.2\%). Motion limits are enabled on
  31.1\% of sliders but only 8.8\% of revolute joints, so the slider
  workflow accepts an optional stroke limit. Among contacts between
  bodies, planar--planar pairs account for 57.1\% and
  cylindrical--cylindrical pairs for 20.6\%, consistent with the face and
  axis references used by the joint workflows.
\end{itemize}

\subsection{Assembly strategies details}
\label{sec:suppl:assembly-workflows}

As described in the main text, we mine assembly behavior from 97,817 deduplicated top-level AutoMate
assemblies (1.29 million mates) and 8,251 Fusion 360 Gallery assemblies,
including 1,684 with joints. A mate constrains the relative position or
motion of two components. AutoMate supplies large-scale mate graphs, while
Fusion additionally supplies the construction timeline and the face, edge,
axis, or center referenced by each joint.
For each connected assembly graph, we infer or read the ground, traverse
outward by breadth-first search, identify cycle-closing edges and coupled
relations, and estimate mobility from the constraints. Fusion timelines
add operation order, and its joint references expose the geometric
interfaces used by each joint. 
The assembly graphs quantify the five regularities or construction strategies 
described in the main paper~\cref{sec:method:mining-assembly-workflows}.

\begin{table*}[t]
\caption{Assembly strategies mined from AutoMate and Fusion 360 Gallery.}
\label{tab:suppl-assembly-strategies}
\centering
\footnotesize
\renewcommand{\arraystretch}{1.08}
\setlength{\tabcolsep}{2pt}
\begin{tabular}{>{\raggedright\arraybackslash}p{1.65in}
                >{\raggedright\arraybackslash}p{4.90in}}
\toprule
\textbf{Strategy} & \textbf{Rule and executable workflows} \\
\midrule
\texttt{assembly\_topology}
& Choose the main housing, base, or plate as ground; insert it first;
  grow outward one component at a time; check mobility after each step.
  Workflows: \texttt{ground\_component}, \texttt{check\_dof}. \\
\texttt{assembly\_joint\_mapping}
& Choose one semantic joint per component pair, then expand it to a
  non-redundant SolidWorks mate set. Workflow: \texttt{joint} with fixed,
  revolute, slider, cylindrical, planar, ball, and pin--slot types. \\
\texttt{assembly\_linkage}
& Build the spanning tree first and add the cycle-closing joint last;
  reject and roll back an over-defined closure. Workflow:
  \texttt{close\_loop}. \\
\texttt{assembly\_coupling}
& Add a relation only after its supporting joints exist. Workflows:
  \texttt{mesh\_gears}, \texttt{rack\_pinion},
  \texttt{screw\_pair}, and \texttt{linear\_coupler}. \\
\texttt{assembly\_motion\_check}
& Drive the specified input, densely sample component transforms, and
  check mobility, interference, ratio, stroke, and fixed bodies; visually
  inspect the completed mechanism and highlight ambiguous roles when useful.
  Workflows: \texttt{check\_dof}, \texttt{check\_motion},
  \texttt{check\_pose},
  \texttt{view\_assembly}. \\
\bottomrule
\end{tabular}
\end{table*}

\textbf{First, assemblies usually grow outward from one grounded
main component as engineers attach new components to the structure already
built. We denote this strategy by
\texttt{assembly\_topology}.}
A single grounded component is the largest component in 51.1\% of AutoMate
and 56.6\% of Fusion assemblies, while 80.2\% and 90.9\% of movable bodies,
respectively, lie within three joints of that ground. In the Fusion
timelines, 85.2\% of joints attach a new component to the already placed
subassembly. Shuffling the joint-creation order within each assembly lowers
this rate to 76.2\%, showing that the observed outward growth is not
explained by assembly topology alone. These dominant grounds, shallow graphs,
and outward construction sequences motivate inserting and grounding the main
component first, then attaching components one at a time.

\textbf{Second, joint semantics determine which geometry to select
and which native mates to create. We denote this strategy by
\texttt{assembly\_joint\_mapping}.}
Joint type predicts both the geometric references and the native mate set.
A revolute joint typically connects cylindrical parts such as a shaft and a
hole, and is implemented by a concentric mate between the cylindrical faces
plus a coincident mate between the end faces. This correspondence motivates
mapping each semantic joint to its characteristic geometric interfaces and
a non-redundant native mate set.

\textbf{Third, engineers close kinematic loops only after
constructing the open connection tree. We denote this strategy by
\texttt{assembly\_linkage}.}
Cycle-closing joints occur later than spanning-tree joints in Fusion, at mean
normalized positions 0.687 and 0.488, respectively. This ordering motivates
building the open spanning tree first and adding each cycle-closing joint
only after its supporting links are in place.

\textbf{Fourth, engineers add motion couplings only after the
supporting joints exist. We denote this strategy by
\texttt{assembly\_coupling}.}
Couplings have typed prerequisites: a gear relation couples two revolute
joints, a screw relation couples rotation and translation within one
cylindrical joint, a rack--pinion relation couples a revolute joint to a
slider, and a linear relation couples two sliders. The strategy therefore
adds each coupling only after the required supporting joints exist.

\textbf{Fifth, frequent floating or over-constrained components
make mobility checks necessary. We denote this strategy by
\texttt{assembly\_motion\_check}.}
Mobility defects are common. Writing $M$ for the estimated number of
unconstrained degrees of freedom, $M\geq6$ in 16.7\% of AutoMate and 20.8\%
of Fusion assemblies, usually indicating floating components, while
$M\leq-3$ flags strongly redundant constraints in 20.9\% and 3.7\% of the
two datasets. These defects motivate explicit degree-of-freedom checks during
construction and motion or pose validation before completion.

\begin{table*}[!t]
\caption{Complete catalog of the 22 executable workflows exposed during
assembly: 20 mined assembly workflows, prescribed-pose validation, and
native visual inspection. References such as a cylindrical face, plane, edge, sphere, or
component are resolved from semantic geometric descriptions rather than
screen coordinates. The last column makes
the abstraction boundary explicit by listing the SolidWorks atomic
operations composed by each workflow.}
\label{tab:suppl-assembly-catalog}
\centering
\footnotesize
\renewcommand{\arraystretch}{1.05}
\setlength{\tabcolsep}{2pt}
\begin{tabular}{>{\raggedright\arraybackslash}p{1.30in}
                >{\raggedright\arraybackslash}p{1.60in}
                >{\centering\arraybackslash}p{0.58in}
                >{\raggedright\arraybackslash}p{2.95in}}
\toprule
\textbf{Workflow} & \textbf{Engineer-level effect} &
\textbf{Calls} &
\textbf{Composed SolidWorks atomic operations} \\
\midrule
\texttt{get\_assembly\_state}
& Return components, transforms, constraint states, faces, and mates.
& $N+2$
& \texttt{IAssemblyDoc.GetComponents};
  \texttt{IComponent2.IsFixed}, \texttt{GetConstrainedStatus}, and
  \texttt{Transform2}; \texttt{IBody2.GetFaces}; surface parameters;
  traverse \texttt{MateGroup}/\texttt{IMate2}. \\
\texttt{view\_assembly}
& Capture up to four labeled native views; optionally highlight components
  resolved from semantic role names.
& 1
& \texttt{IModelDoc2.ShowNamedView2}, \texttt{ViewZoomtofit2},
  \texttt{IModelDoc2.SaveBMP}; temporary shaded-with-edges display,
  reference hiding, and \texttt{IComponent2.Select4} highlighting; restore
  selection, display toggles, orientation, translation, and scale. \\
\texttt{insert\_component}
& Insert a saved part at an initial pose and assign an alias.
& 1--2
& \texttt{ISldWorks.OpenDoc6}; \texttt{IAssemblyDoc.AddComponent5};
  \texttt{IMathUtility.CreateTransform}; set
  \texttt{IComponent2.Transform2}; rebuild mates. \\
\texttt{ground\_component}
& Fix the base component at its current pose.
& 1
& \texttt{IModelDocExtension.SelectByID2};
  \texttt{IAssemblyDoc.FixComponent}. \\
\texttt{mate}
& Add one standard or mechanical native mate.
& 3
& Select component entities; standard mates use
  \texttt{IAssemblyDoc.AddMate5}; slot, rack--pinion, screw, and linear
  coupling use \texttt{CreateMateData}/\texttt{CreateMate};
  \texttt{IFeature.ModifyDefinition}; rebuild and inspect errors. \\
\texttt{delete\_mate}
& Remove a named mate and update the recorded joint graph.
& 1
& \texttt{IFeature.Select2};
  \texttt{IModelDocExtension.DeleteSelection2}; \texttt{EditRebuild3}. \\
\texttt{mount\_on\_shaft}
& Concentric shaft mounting, axial location, and optional keyed lock.
& 6--7
& Two \texttt{AddMate5} calls: concentric (optionally
  \texttt{lockRotation}) plus coincident or distance; inspect signed offset
  when needed; delete created mates on failure. \\
\texttt{revolute\_joint}
& Leave only rotation about a shared axis.
& 6
& \texttt{AddMate5} concentric $+$ coincident end-face mates;
  \texttt{DeleteSelection2} rollback on failure. \\
\bottomrule
\end{tabular}
\end{table*}

\subsection{Assembly workflow catalog}

\Cref{tab:suppl-assembly-catalog} summarizes the 22 executable workflows
available to the assembly agent. They cover the complete construction loop:
observing and visualizing the current assembly, inserting and grounding
components, creating or deleting native mates, building semantic joints and
motion couplings, checking mobility, interference, motion, and prescribed
poses, and finally saving and completing the assembly. For each workflow, the
table separates its engineer-level effect from the SolidWorks atomic
operations used to implement it and reports the number of service calls,
making explicit how the interface turns several low-level operations into one
reusable assembly action.
As in the part-workflow catalog, ``Calls'' counts top-level requests from the
workflow runtime to the atomic assembly service along a successful execution
path; a batched geometry query or motion sample counts as one request. In the
table, $N$ denotes the number of components. Call-count ranges reflect
optional arguments and joint types, and rollback adds one delete request for
each created mate. ``Input-dependent'' means that the count varies with the
number of prescribed poses, temporary mates, and angle-mate variant retries.

\begin{table*}[t]
\ContinuedFloat
\caption{Complete catalog of the 22 executable assembly workflows
(continued).}
\centering
\footnotesize
\renewcommand{\arraystretch}{1.05}
\setlength{\tabcolsep}{2pt}
\begin{tabular}{>{\raggedright\arraybackslash}p{1.30in}
                >{\raggedright\arraybackslash}p{1.60in}
                >{\centering\arraybackslash}p{0.58in}
                >{\raggedright\arraybackslash}p{2.95in}}
\toprule
\textbf{Workflow} & \textbf{Engineer-level effect} &
\textbf{Calls} &
\textbf{Composed SolidWorks atomic operations} \\
\midrule
\texttt{slider\_joint}
& Leave one translation along a shared axis.
& 6
& \texttt{AddMate5} concentric $+$ coincident guide-plane mates;
  \texttt{DeleteSelection2} rollback on failure. \\
\texttt{mesh\_gears}
& Check gear geometry and phase, then couple two rotations.
& 6
& Read \texttt{IFace2}/cylinder parameters and component
  \texttt{Transform2}; set follower \texttt{Transform2};
  \texttt{AddMate5} gear mate; \texttt{ModifyDefinition} for ratio/direction. \\
\texttt{crank\_slider}
& Build three pin joints and one slider guide in one transaction.
& 19
& Six \texttt{AddMate5} calls (concentric, coincident, and distance);
  reverse-order \texttt{DeleteSelection2} rollback. \\
\texttt{check\_interference}
& Detect colliding component pairs at the current pose.
& 1
& Configure \texttt{InterferenceDetectionManager};
  \texttt{GetInterferences}; read \texttt{IInterference.Components} and
  \texttt{Volume}; \texttt{Done}. \\
\texttt{check\_motion}
& Drive and sample one cycle; measure motion, fixed-body drift, and collision.
& 4
& Motion-study manager; \texttt{CreateDefinition}/\texttt{CreateFeature}
  for a rotary motor; \texttt{SetDuration}, \texttt{SetTime};
  read component \texttt{Transform2}; run interference at each frame. \\
\texttt{check\_pose}
& Temporarily impose one or more prescribed configurations; verify component
  key points, interference, and fixed-body drift; then remove the temporary mates.
& \shortstack[c]{Input-\\dependent}
& Create angle, offset, or distance mates with \texttt{AddMate5}; rebuild;
  read \texttt{IComponent2.Transform2}; run native interference detection;
  remove temporary mates with \texttt{DeleteSelection2}. \\
\texttt{save\_assembly}
& Save the native assembly.
& 1
& \texttt{IModelDoc2.SaveAs2}. \\
\texttt{finish}
& End only after saving and a passing motion check, or a passing pose check
  paired with the expected mobility.
& 0
& No mutating SolidWorks call; Python-side gate over saved, motion/pose, and
  mobility state. \\
\texttt{joint}
& Expand a fixed, revolute, cylindrical, slider, planar, ball, or pin--slot joint.
& 3--9
& Compose \texttt{AddMate5} lock/concentric/coincident/distance mates;
  pin--slot uses \texttt{CreateMateData}/\texttt{CreateMate};
  delete the partial mate set on failure. \\
\texttt{close\_loop}
& Add a minimally constrained closing joint and retry safely if over-defined.
& 4--10/try
& The same mate calls as \texttt{joint}; read
  \texttt{GetConstrainedStatus}; \texttt{DeleteSelection2} rollback before
  an alternative expansion. \\
\texttt{rack\_pinion}
& Couple an existing slider and revolute joint.
& 4
& Read edge/cylinder geometry; \texttt{CreateMateData} with
  \texttt{IRackPinionMateFeatureData}; \texttt{CreateMate}; rebuild and
  inspect feature errors. \\
\texttt{screw\_pair}
& Couple screw rotation to nut translation by lead.
& 3--5
& Read matching cylinder faces; \texttt{CreateMateData} with
  \texttt{IScrewMateFeatureData}; \texttt{CreateMate};
  \texttt{ModifyDefinition}/\texttt{IDimension.SetSystemValue3}; rebuild. \\
\texttt{linear\_coupler}
& Couple two slider displacements by a signed ratio.
& 4
& Read line edges; \texttt{CreateMateData} with
  \texttt{ILinearCouplerMateFeatureData}; set reference components and ratio;
  \texttt{CreateMate}; rebuild. \\
\texttt{check\_dof}
& Estimate mobility and report floating, ungrounded, or over-defined components.
& 1
& \texttt{GetComponents}, \texttt{IsFixed}, and
  \texttt{GetConstrainedStatus}; Python-side joint-graph traversal and
  Gr\"ubler count; no model mutation. \\
\bottomrule
\end{tabular}
\end{table*}

\clearpage
\clearpage

The five strategies summarized in
\cref{tab:suppl-assembly-strategies} are stored as Markdown recipe files and
concatenated, in full and in a fixed order, before each assembly task brief.
Separately, the agent receives JSON schemas for all 22 assembly workflows
through native function calling. The model maps the recipe instructions to
workflow calls and revises that choice after each returned assembly state.
Consequently, strategy compliance is not enforced by a hard-coded global
state machine. Enforcement is local to the workflows: preconditions produce
structured errors, failed loop closure is rolled back, mobility and motion
or pose checks return measured diagnostics, and \texttt{finish} requires a
saved assembly plus a passing motion check or a passing pose check paired
with the expected mobility.

This final validation is intentionally stricter than merely reproducing the
mate patterns found in existing assemblies. During construction, the agent
does not treat a gear assembly as complete simply because mates have been
added. It must use \texttt{check\_motion} to verify that the gears actually
move and transmit rotation with the required ratio and direction.

\section{Benchmark task specifications}
\label{sec:suppl:benchmark-tasks}

\subsection{Benchmark taxonomy}
\label{sec:suppl:benchmark-taxonomy}

\begin{figure}[t]
\centering
\includegraphics[width=\columnwidth]{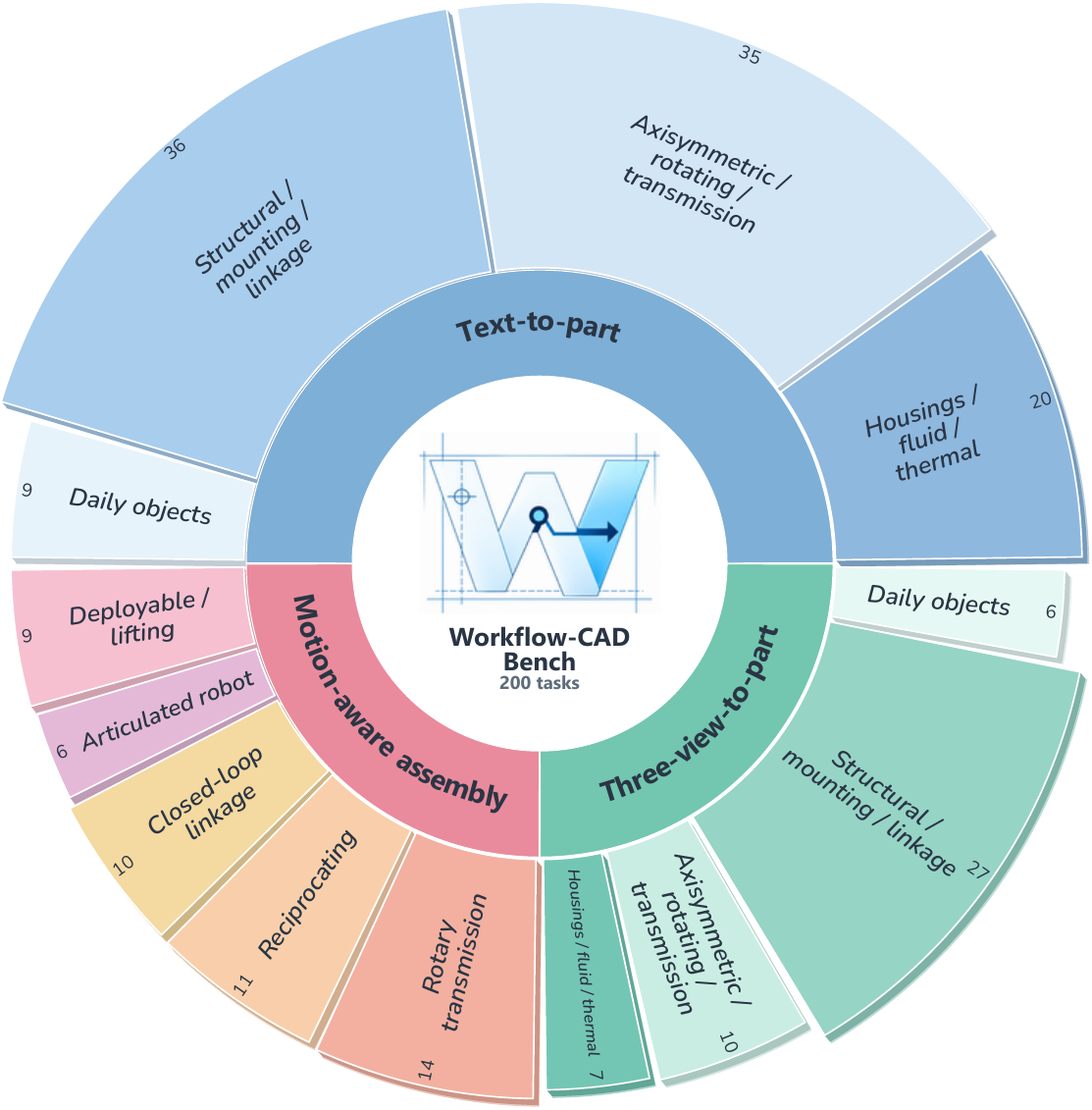}
\caption{Taxonomy of \bench. Numbers denote the cases in each mutually
exclusive functional category.}
\label{fig:suppl-bench-taxonomy}
\end{figure}

The three benchmark tracks use mutually exclusive functional categories. The
two part tracks share four part categories, while the assembly track uses five
kinematic categories. The categories are illustrated in \cref{fig:suppl-bench-taxonomy}.

\subsection{Text-to-part tasks}
\label{sec:suppl:text-tasks}

Across the three splits, the tasks cover sketch-based extrusion, revolution,
cutting, hole making, patterning, filleting, chamfering, ribbing, shelling,
drafting, sheet-metal bends, sweeps, and lofts. Less common constructions
include offset-plane extrusions terminated by a face, flanged elbow sweeps,
and polygon-to-circle loft transitions. The 50 easy tasks contain 3--10
explicit requirements (5.9 on average), the 30 medium tasks contain 7--14
(10.1 on average), and the 20 hard tasks contain 7--12 (9.2 on average).
Easy tasks usually require a base body and a few secondary features; medium
tasks couple requirements across more construction stages; hard tasks derive
their difficulty from analytic geometry, engineering calculations, or
delicate CAD operations rather than requirement count alone.

\Cref{tab:suppl-tasks} lists the 50-task easy split of the text-to-part track.
\Cref{tab:suppl-medium-tasks} lists the 30-task medium
split.  
Each brief in the medium split
specifies a single solid and combines many dependent features.
\Cref{tab:suppl-hard-tasks} lists the 20-task hard split.
These hard tasks concentrate on exact analytic curves, derived engineering
quantities, and CAD operations that remain delicate in manual workflows.

\begin{table*}[t]
\caption{Text-to-part easy-split cases 1--33. ``Items'' is the number of
explicit rubric items. Volumes are analytic; an interval means that the
brief admits several valid constructions (\eg a rim fillet on either edge
loop), and ``--'' means that volume is not graded because the brief leaves
the shape underdetermined.}
\label{tab:suppl-tasks}
\centering
\footnotesize
\renewcommand{\arraystretch}{1.08}
\setlength{\tabcolsep}{3.5pt}
\begin{tabular}{r >{\raggedright\arraybackslash}p{1.1in}
                  >{\raggedright\arraybackslash}p{4.13in} c r}
\toprule
\textbf{\#} & \textbf{Task} & \textbf{Brief (paraphrased)} & \textbf{Items} &
\textbf{Volume (mm$^3$)} \\
\midrule
1 & Phone stand & One-piece stand: $110\times78\times8$ base, 6-mm backplate
at $74^\circ$ reaching 118 mm, front lip, two side ribs, U-shaped cable
slot, three cooling slots, two pad recesses on the bottom, R4 fillets,
flat bottom, bilateral symmetry. & 10 & -- \\
2 & Stool & $200\times200\times20$ seat on four $\varnothing$20$\times$150 legs,
30 mm from the adjacent edges. & 6 & 988{,}496 \\
3 & Rounded stool & The stool with R5 fillets on the four top edges of the
seat. & 7 & 984{,}252 \\
4 & Storage box & Open-top box, $120\times80\times50$ outside, 8-mm walls and
floor, R3 fillet on the rim. & 7 & 198{,}990--200{,}448 \\
5 & Die & 20-mm cube, R1.5 on all edges, pips 1--6 with opposite faces
summing to seven, pips $\varnothing$3 and 1.5 deep. & 5 & 7{,}666--7{,}740 \\
6 & Mug & Cylinder 90 high, OD 70, 5-mm wall, 10-mm floor, ring handle of OD
40, 8-mm section. & 5 & 120{,}166--126{,}600 \\
7 & Vase & Hollow axisymmetric vase, 150 high, flat bottom, maximum
$\varnothing$90 in the middle, neck $\varnothing$40, mouth $\varnothing$55,
10-mm bottom. & 6 & -- \\
8 & Key & 2-mm key: $\varnothing$20 ring head with $10\times5$ hole,
$42\times8$ shaft, five evenly spaced tooth grooves on one edge. & 5 & -- \\
9 & Soap dish & Open-top tray, $110\times75\times25$ outside, 3-mm walls and
floor, R8 on the four outer vertical edges, 15 $\varnothing$5 drain holes
through the floor in a centered $3\times5$ grid at 18-mm pitch. & 8 &
46{,}121--46{,}593 \\
10 & Hex nut & 30 across flats, 8 thick, $\varnothing$10 through-hole,
centered on the origin in thickness. & 5 & 5{,}607 \\
11 & Wrench & 8-mm single-ended wrench, $\varnothing$40 head with a 13-mm
hexagonal jaw, $120\times15$ handle merged with the head. & 5 &
20{,}940--23{,}282 \\
12 & Gear disc & $\varnothing$60$\times$8 disc, $\varnothing$12 bore, 12 equal
through-slots near the rim. & 4 & -- \\
13 & Flange & $\varnothing$100$\times$12, $\varnothing$30 bore, six $\varnothing$8
holes on an 80-mm pitch circle. & 6 & 82{,}146 \\
14 & Pulley & $\varnothing$100$\times$20, $\varnothing$25 bore, annular top
groove (OD 90, ID 70, 10 deep), C1 on the rim. & 6 & 121{,}817--121{,}973 \\
15 & Stepped shaft & One revolve, 90 long, segments
$\varnothing$35$\times$40, $\varnothing$30$\times$30, $\varnothing$25$\times$20,
centered on the origin. & 5 & 107{,}207 \\
16 & Socket-head bolt & Shaft $\varnothing$10$\times$50, head
$\varnothing$18$\times$6, blind hexagonal socket 8 across flats and 4 deep,
R0.5 at the head--shaft junction. & 6 & 5{,}234 \\
17 & L bracket & 5-mm L bracket, legs 30 wide, outer lengths 50 and 40; two
$\varnothing$6 holes in the horizontal leg, one in the vertical leg. & 6 &
12{,}326 \\
18 & Spacer ring & OD 40, ID 20, 10 thick, C1 on both outer edges. & 4 &
9{,}301 \\
19 & Mounting plate & $120\times80\times6$ plate, R8 vertical corners, four
$\varnothing$7 holes 12 mm from the edges, central $40\times20$ window. &
6 & 51{,}547 \\
20 & Stepped block & $80\times50\times15$ lower block with a
$40\times50\times15$ step flush with one end; $\varnothing$10, 8-deep blind
hole in the exposed step face. & 5 & 89{,}372 \\
21 & Knob & $\varnothing$30$\times$20 body, $\varnothing$10$\times$5 top boss,
12 full-height axial grooves $2\times1$, $\varnothing$6, 12-deep bottom bore.
& 6 & 13{,}716 \\
22 & Hex standoff & M3 standoff, 8 across flats, 20 long,
$\varnothing$3.3, 6-deep holes in both ends. & 4 & 1{,}006 \\
23 & U channel & 100 long, $40\times30$ outside, 4-mm walls and floor,
opening up, three $\varnothing$8 floor holes at 30-mm pitch, centered. & 5
& 36{,}197 \\
24 & Flanged bushing & Sleeve OD 20, ID 12, 30 long including a
$\varnothing$36$\times$5 flange; through-bore. & 5 & 9{,}550 \\
25 & Wedge block & 60 long, 30 wide, 20 high at one end and 5 at the other,
planar sloped top, flat bottom. & 6 & 22{,}500 \\
26 & Lightened wheel & $\varnothing$120$\times$10, $\varnothing$20 bore, four
$60^\circ$ annular-sector cutouts between radii 20 and 50. & 5 & 65{,}973 \\
27 & End cap & $\varnothing$60, 15 high, $\varnothing$50 cavity 12 deep from
below, $\varnothing$8 hole in the 3-mm top plate. & 5 & 18{,}699 \\
28 & Link bar & 6 thick, $\varnothing$24 round ends 80 apart, 14-wide web in
one outline, $\varnothing$10 holes at both centers. & 6 & 9{,}311 \\
29 & Fork end & $\varnothing$20$\times$35 shank, $40\times24\times10$ yoke
block, two 8-mm ears flush with its ends with R12 tips 18 above the block,
$\varnothing$10 cross pin hole, $\varnothing$6.8, 20-deep tap hole in the
shank end. & 9 & 29{,}144 \\
30 & Three-lug housing & 10-mm flange of R32 with three lugs at $120^\circ$
(22 wide, R11 tips at radius 48, $\varnothing$9 holes), $\varnothing$46 hub
38 high, $\varnothing$30 through-bore, $\varnothing$38, 9-deep counterbore. &
9 & 62{,}774 \\
31 & Pillow block & Symmetric; $120\times40\times12$ base, full-width body of
a 60-wide rectangle and an R30 top arc centered 45 above the floor,
$\varnothing$35 bearing bore, two 11-wide slots 45 from the center. & 8 &
149{,}943 \\
32 & Twin-arm rocker & $\varnothing$30$\times$40 hub with $\varnothing$12
bore, two 8-mm arm plates flush with the hub ends, R15/R10 outline with
tangent lines and 70-mm arm, $\varnothing$8 pin hole through both. & 7 &
47{,}876 \\
33 & Clevis bracket & $80\times50\times10$ base with four $\varnothing$9 holes,
two 10-mm ears 30 apart with R15 tops 25 above the base, $\varnothing$12
pin hole through both, symmetric. & 9 & 57{,}262 \\
\bottomrule
\end{tabular}
\end{table*}

\begin{table*}[t]
\ContinuedFloat
\caption{Text-to-part easy-split cases 34--50 (continued).}
\centering
\footnotesize
\renewcommand{\arraystretch}{1.08}
\setlength{\tabcolsep}{3.5pt}
\begin{tabular}{r >{\raggedright\arraybackslash}p{1.1in}
                  >{\raggedright\arraybackslash}p{4.13in} c r}
\toprule
\textbf{\#} & \textbf{Task} & \textbf{Brief (paraphrased)} & \textbf{Items} &
\textbf{Volume (mm$^3$)} \\
\midrule
34 & Diamond flange & 8-mm plate outlined by an R25 center arc and two R12
end arcs 42 away joined by tangents, $\varnothing$10 holes in the ends,
$\varnothing$34 boss to 26, $\varnothing$20 through-bore. & 7 & 36{,}602 \\
35 & Support seat & $100\times50\times10$ base, 20-mm upright of a 50-wide
rectangle and an R25 top arc centered 40 above the floor, $\varnothing$24
bore, two $\varnothing$10 holes 38 from the center. & 7 & 89{,}016 \\
36 & Crank & $\varnothing$36$\times$24 hub with $\varnothing$16 bore, 10-mm
arm flush with one hub end, R18/R11 outline 60 apart, $\varnothing$22,
15-high pin boss on the flush side, $\varnothing$10 hole through boss and
arm. & 8 & 37{,}672 \\
37 & Gusset bracket & $60\times40\times6$ base, 6-mm upright 30 high flush
with one end, 4-mm right-angle rib with 18-mm legs on the mid-plane. & 6 &
20{,}808--22{,}248 \\
38 & Corner-rib stand & $80\times50\times8$ base, 8-mm upright 36 high flush
with one end, 5-mm right-angle rib with 24-mm legs on the mid-plane. & 6 &
44{,}640--47{,}840 \\
39 & Open box & $80\times50\times40$ outside, 4-mm walls and floor, open top,
made by shelling. & 6 & 51{,}136 \\
40 & Lidless housing & $100\times60\times50$ outside, 5-mm walls and floor,
open top, made by shelling. & 6 & 97{,}500 \\
41 & Counterbore plate & $80\times60\times10$ plate, four $\varnothing$6
holes with $\varnothing$11, 3.5-deep counterbores on top, 14 mm from the
edges. & 5 & 45{,}935--46{,}448 \\
42 & Countersunk cover & $90\times70\times12$ plate, four $\varnothing$8 holes
with $\varnothing$14, 4-deep countersinks on top, 15 mm from the edges. & 5 &
71{,}529--72{,}433 \\
43 & Side pad & $50\times40\times30$ block; a $\varnothing$16 circle on a plane
20 mm beyond the right face, extruded up to that face. & 4 & 64{,}021 \\
44 & Through boss & $60\times48\times36$ block; a $\varnothing$20 circle on a
plane 24 mm beyond the right face, extruded up to that face. & 4 &
111{,}220 \\
45 & Sheet plate A & Sheet-metal plate, $70\times40$ outline, 1.5 thick. & 3 &
4{,}200 \\
46 & Sheet plate B & Sheet-metal plate, $90\times50$ outline, 1.2 thick. & 3 &
5{,}400 \\
47 & Flanged elbow & Solid $\varnothing$16 tube on a quarter arc of radius
40, $\varnothing$36$\times$4 flanges at both ends protruding outward along
the tube axis. & 7 & 19{,}168--20{,}776 \\
48 & Bent tube & Solid $\varnothing$20 tube on a quarter arc of radius 50,
$\varnothing$44$\times$5 flanges at both ends protruding outward. & 7 &
39{,}879 \\
49 & Square-to-round & 40 high, $30\times30$ square bottom, $\varnothing$24
circular top, parallel and coaxial, joined by a loft. & 5 & -- \\
50 & Rect-to-circle & 50 high, $40\times24$ rectangular bottom,
$\varnothing$28 circular top, parallel and coaxial, joined by a loft. & 5 &
-- \\
\bottomrule
\end{tabular}
\end{table*}


\begin{table*}[t]
\caption{Text-to-part medium-split cases 1--30.}
\label{tab:suppl-medium-tasks}
\centering
\footnotesize
\setlength{\baselineskip}{8.5pt}
\renewcommand{\arraystretch}{0.95}
\setlength{\tabcolsep}{3pt}
\begin{tabular}{r >{\raggedright\arraybackslash}p{1.55in}
                  >{\raggedright\arraybackslash}p{4.52in} r}
\toprule
\textbf{\#} & \textbf{Case / target} & \textbf{Defining requirements} &
\textbf{Items} \\
\midrule
1 & \texttt{quad\_frame}: quadrotor frame &
X-shaped diagonal arms; 16 motor holes and four flight-controller holes
in converted coordinates; eight R5 inner fillets. & 12 \\
2 & \texttt{connecting\_rod}: engine connecting rod &
Externally tangent big and small ends; trapezoidal pockets on both faces;
centerline oil drilling; chamfers on both ends of the bearing bore. & 12 \\
3 & \texttt{finned\_housing}: finned motor housing &
Twelve correctly phased fins; rounded-square flange; base and flange hole
patterns sharing the required orientation. & 12 \\
4 & \texttt{sprocket}: 18-tooth roller-chain sprocket &
Tooth-space pattern on the pitch circle; keyway; aligned lightening holes,
tooth spaces, and keyway; radial set-screw hole. & 11 \\
5 & \texttt{handwheel}: five-spoke handwheel &
Annular rim and five mid-plane radial spokes; hexagonal bore; offset crank
handle with spherical knob. & 9 \\
6 & \texttt{camshaft}: four-cylinder camshaft &
Four tangent-line cam profiles; clockwise $90^\circ$ phase progression;
locating pin hole and keyway at specified orientations. & 10 \\
7 & \texttt{planet\_carrier}: planetary carrier &
Two plates joined by posts; planet-shaft holes alternating with posts;
tapped output-shaft bore; four circular edge chamfers. & 11 \\
8 & \texttt{impeller}: backward-curved impeller &
Circular-arc blades from prescribed centers and radii in a $45^\circ$
pattern; blade height and cutoff radius; hub chamfer. & 10 \\
9 & \texttt{crankshaft}: two-throw crankshaft &
Throws phased $180^\circ$; tangent obround webs with sector counterweights;
cross-drilled oil holes through crankpins; oriented keyway and pin hole. & 14 \\
10 & \texttt{brake\_disc}: vented brake disc &
Two friction rings separated by 40 radial vanes; 20 cooling holes in
alternate vane gaps; hat and five bolt holes aligned with vanes. & 10 \\
11 & \texttt{wheel\_rim}: five-spoke wheel rim &
Phased annular-sector windows with 20 R8 corners; spokes aligned with bolt
holes; radial valve hole aligned with a window. & 12 \\
12 & \texttt{geneva\_wheel}: six-slot Geneva wheel &
Six radial slots alternating with concave locking arcs; lightening holes on
bisectors; radial screw hole through the hub along a slot direction. & 9 \\
13 & \texttt{spiral\_staircase}: spiral staircase &
Twelve counterclockwise rising sector steps; one post on each step
bisector; hemispherical post caps, base, column, and mounting holes. & 10 \\
14 & \texttt{engine\_piston}: engine piston &
Three ring grooves; internal pin bosses and through pin bore; two valve
reliefs perpendicular to the bore; radial drain holes behind the oil groove. & 11 \\
15 & \texttt{pipe\_manifold}: three-branch manifold &
Main and three branch bores connected internally; end-flange bolts offset
$22.5^\circ$; branch bolts on diagonals; bottom drain boss. & 11 \\
16 & \texttt{gearbox\_housing}: lower gearbox housing &
Thin-wall open box; two unequal bearing bosses and bores on each side;
ribs below bosses; flange, feet, drain, and sight features. & 12 \\
17 & \texttt{heat\_sink}: pin-fin heat sink &
$9\times9$ array omitting four corners (77 pins); corner mounting holes
with counterbores; bottom thermal pad. & 9 \\
18 & \texttt{honeycomb\_panel}: honeycomb panel &
Five staggered rows of 43 blind hexagonal pockets; specified hexagon
orientation and 3-mm webs; corner fillets concentric with mounting holes. & 10 \\
19 & \texttt{gear\_pump\_body}: gear-pump body &
Figure-eight twin chambers; through drive bore and blind driven bore;
different side-port bosses and bores; diagonal dowel holes. & 10 \\
20 & \texttt{cold\_plate}: serpentine cold plate &
Five straight channel segments joined by four half-turns; side inlet and
outlet reaching the channel ends; rounded-rectangle seal groove. & 10 \\
21 & \texttt{truss\_beam}: Warren truss beam &
Nine alternating triangular openings with 27 R4 inner corners; unequal end
boss bores; reinforcement ribs along the long edges. & 8 \\
22 & \texttt{turbine\_disk}: dovetail-slot turbine disc &
Thirty outward-narrowing axial dovetail slots; lightening holes on slot-gap
bisectors; four annular blends joining hub, web, and rim. & 10 \\
23 & \texttt{bt40\_holder}: BT40 tool holder &
7:24 taper; $60^\circ$ V groove; two drive keyways $90^\circ$ from the
orientation hole; center coolant passage joining both end bores. & 10 \\
24 & \texttt{hydraulic\_manifold}: hydraulic valve block &
P/A/T, main, and cross drillings entering from four directions; passages
must intersect without over-drilling; plug counterbores. & 9 \\
25 & \texttt{rocket\_fin\_can}: rocket fin can &
Thin-wall de Laval nozzle; four swept trapezoidal fins with specified
planform and orientation; vents offset $45^\circ$ from fins. & 9 \\
26 & \texttt{ratchet\_wheel}: ratchet wheel &
Twenty-four one-way saw teeth with specified tooth-back handedness;
hexagonal bore oriented with mounting holes. & 8 \\
27 & \texttt{injector\_plate}: injector plate &
Four rear manifold grooves; 80 injector holes on four rings with distinct
starting phases; dowel holes positioned relative to bolt holes. & 9 \\
28 & \texttt{robot\_arm\_link}: robot-arm link &
External-tangent arm body; pockets on both faces; shoulder threaded-hole
pattern; elbow split and transverse clamp-screw hole. & 10 \\
29 & \texttt{menger\_sponge}: level-2 Menger sponge &
Two recursive removal levels forming 27 correctly positioned through
openings while retaining one valid connected solid. & 7 \\
30 & \texttt{u\_joint\_yoke}: universal-joint yoke &
Two fork arms and coaxial bearing holes; semicircular tips; internal
keyway, radial set-screw hole, and outer bearing-cap screw holes. & 9 \\
\bottomrule
\end{tabular}
\end{table*}


\begin{table*}[t]
\caption{Text-to-part hard-split cases 1--20.}
\label{tab:suppl-hard-tasks}
\centering
\footnotesize
\renewcommand{\arraystretch}{1.06}
\setlength{\tabcolsep}{3pt}
\begin{tabular}{r >{\raggedright\arraybackslash}p{1.55in}
                  >{\raggedright\arraybackslash}p{4.52in} r}
\toprule
\textbf{\#} & \textbf{Case / target} & \textbf{Defining requirements} &
\textbf{Items} \\
\midrule
1 & \texttt{worm\_shaft}: double-start worm shaft &
Left-hand ZA thread with prescribed axial module, profile, and phase;
shoulder termination, retaining-ring groove, keyway, and pointed blind
center hole. & 12 \\
2 & \texttt{compound\_drill\_block}: compound-angle drilling block &
One bore from two projected angles; a second bore normal to an inclined
face and intersecting the first; drill points and counterbore. & 11 \\
3 & \texttt{variable\_fillet\_bracket}: L bracket &
Linearly varying R3--R9 fillet followed by a rib; unequal-leg chamfer,
countersunk holes, and an obround slot. & 11 \\
4 & \texttt{roller\_cam}: roller-follower disc cam &
Cycloidal rise and 3-4-5 polynomial return; roller-envelope profile;
keyway-referenced phase and radial set-screw hole. & 9 \\
5 & \texttt{helical\_ring\_gear}: internal helical ring gear &
Normal-to-transverse gear conversion; involute internal teeth with specified
phase and right-hand helix; one-sided chamfer. & 10 \\
6 & \texttt{naca\_wing\_rib}: NACA 4412 wing rib &
Normal-offset airfoil with a closed trailing edge; thickness-dependent
lightening holes and spar notch. & 8 \\
7 & \texttt{scroll\_wrap\_plate}: stationary scroll plate &
Wrap bounded by two circular involutes, trimmed at both generating-line
ends, with a top sealing groove. & 8 \\
8 & \texttt{variable\_pitch\_spring}: compression spring &
Left-hand helix with piecewise-linear pitch and integrated axial position;
ground planar ends. & 9 \\
9 & \texttt{pipe\_route\_3d}: three-dimensional pipe route &
PI-defined spatial centerline with prescribed bend radii; oriented flange
holes and a connected branch bore. & 9 \\
10 & \texttt{sheet\_metal\_kfactor}: bent sheet-metal bracket &
Two bends with specified thickness, radius, and K-factor; holes located from
flat-pattern coordinates after bend-allowance conversion. & 9 \\
11 & \texttt{cast\_manifold\_fillets}: cast tee manifold &
R5 saddle blends along complete cylinder intersections; inner flange
fillets, phased bolt holes, and connected internal passages. & 8 \\
12 & \texttt{die\_cast\_cover}: drafted die-cast cover &
Opposed wall drafts about a neutral parting plane; reverse-drafted bosses,
two-sided drafted ribs, pads, and through holes. & 10 \\
13 & \texttt{balanced\_flywheel}: statically balanced flywheel &
Balance-hole diameter derived after accounting for an eccentric keyway;
window-edge and web blends. & 9 \\
14 & \texttt{inch\_callout\_plate}: inch-callout plate &
Inch conversion, letter and number drill sizes, 10--32 tap drill,
counterbore, 82-degree countersink, and phased bolt-circle holes. & 9 \\
15 & \texttt{clipped\_hex\_perforation}: perforated disc &
Pointy-top hexagonal lattice clipped only when all six vertices remain
inside a truncated circular plate. & 9 \\
16 & \texttt{tangent\_link\_lever}: three-bore link lever &
External common tangents and concave tangent arcs with solved contact
points; keyway and curved slot. & 9 \\
17 & \texttt{filleted\_shell\_enclosure}: thin-wall enclosure &
Fillet-before-shell geometry, locating rabbet, vents crossing the top
fillets, and internal screw bosses. & 9 \\
18 & \texttt{tricylinder\_node}: three-axis structural node &
Steinmetz intersection core, three mutually connected blind bores, and a
spatial brace ending in a sphere. & 7 \\
19 & \texttt{twisted\_flute\_knob}: fluted control knob &
Six constant-rate left-hand twisted grooves with a planar cutoff, D-shaped
bore, and radial set-screw hole. & 8 \\
20 & \texttt{spherical\_dome\_ports}: domed pressure-vessel head &
Radial ports specified by latitude and longitude, an off-center horizontal
port, phased flange holes, and a connected cavity. & 9 \\
\bottomrule
\end{tabular}
\end{table*}

\subsection{Three-view reconstruction tasks}
\label{sec:suppl:threeview-tasks}

\Cref{tab:suppl-threeview-cases,tab:suppl-threeview-cases-cont}
lists all 50 three-view-to-part tasks. Each input is a dimensioned third-angle
drawing with aligned front, top, and right views, accompanied only by a
neutral one-sentence instruction. ``Items'' is the number of explicit
drawing-grounded rubric items.
Each case has an independently constructed reference STEP. All cases
require one solid except \texttt{c3\_gear\_pair} and
\texttt{d17\_pipe\_clamp}, which require exactly two.

\begin{table*}[t]
\caption{Three-view reconstruction cases 1--31. Dimensions are in millimeters.}
\label{tab:suppl-threeview-cases}
\centering
\footnotesize
\renewcommand{\arraystretch}{1.07}
\setlength{\tabcolsep}{3pt}
\begin{tabular}{r >{\raggedright\arraybackslash}p{1.35in}
                  >{\raggedright\arraybackslash}p{4.75in} r}
\toprule
\textbf{\#} & \textbf{Case / target} & \textbf{Defining geometry visible in the drawing} &
\textbf{Items} \\
\midrule
1 & \texttt{b01\_l\_bracket}: L bracket &
$60\times40\times20$ envelope; 30-wide corner notch, 20-high step, and
uniform 20-mm thickness. & 5 \\
2 & \texttt{b02\_plate\_with\_holes}: plate &
$80\times40\times10$ plate; two $\varnothing$10 through-holes, first center
20 from the left and 40 center-to-center. & 5 \\
3 & \texttt{b03\_t\_bracket}: T bracket &
$60\times45\times15$; $60\times15$ top bar and centered $20\times30$ stem,
both prismatic through 15. & 6 \\
4 & \texttt{b04\_keycap}: keycap &
$18\times18\times8$ tapered cap; centered $14\times14$ top joined to the
$18\times18$ base by planar sides. & 6 \\
5 & \texttt{b05\_keycap\_full}: detailed keycap &
The tapered keycap with a $12\times12\times3$ blind bottom pocket and R1.5
top-edge fillets. & 6 \\
6 & \texttt{b06\_lock\_striker}: striker plate &
$30\times80\times3$ plate with four R3 corners, two $\varnothing$5 holes at
64 pitch, and a centered $20\times10$ round-ended slot. & 6 \\
7 & \texttt{b07\_switch\_panel}: switch panel &
$86\times86\times10$ panel with R3 corners, four $\varnothing$3 holes on a
$70\times70$ pattern, and a $50\times50\times3$ front pocket. & 6 \\
8 & \texttt{b11\_gear\_disc}: gear disc &
$\varnothing60\times10$ disc, $\varnothing$15 bore, and 12
$\varnothing$4 through-holes on a $\varnothing$48 pitch circle. & 5 \\
9 & \texttt{b22\_round\_bowl}: round bowl &
$\varnothing80\times40$ body with an open $\varnothing$68 cylindrical
cavity 34 deep. & 4 \\
10 & \texttt{c1\_mech\_keyboard}: mini keyboard &
$285\times95\times20$ overall; 15-high base and a $13\times5$ grid of
tapered keys at 18 pitch. & 7 \\
11 & \texttt{c2\_gearbox}: reducer housing &
$200\times120\times20$ flange under a $160\times100\times100$ open-top
8-mm-wall box; $\varnothing$25 bore and four $\varnothing$8 mounting holes
on a $170\times90$ pattern. & 8 \\
12 & \texttt{c3\_gear\_pair}: gear pair &
Two separate solids: $\varnothing$40/$\varnothing$80 root circles, 70 center
distance, 10/20 rectangular teeth, and 15-mm thickness. & 7 \\
13 & \texttt{c4\_desktop\_pen\_holder}: desk organizer &
$140\times110\times90$ stepped multi-bin body with 8-mm floors, a
$14\times10$ cable slot, a $24\times10$ handle slot, and R3 outer edges. & 10 \\
14 & \texttt{d01\_v\_block}: V-block &
$80\times50\times40$ block; $90^\circ$ V-groove 40 wide, 4-mm relief,
$20\times6$ bottom slot, and two $8\times5$ side grooves. & 7 \\
15 & \texttt{d04\_flange\_cover}: flange cover &
$\varnothing100\times32$ overall; 12-thick flange, $\varnothing$50 boss,
$\varnothing$30 bore, two $\varnothing$70 recesses, and six
$\varnothing$9 holes on $\varnothing$80. & 8 \\
16 & \texttt{d05\_stepped\_shaft}: stepped shaft &
120 long with $\varnothing20\times30$, $\varnothing30\times50$, and
$\varnothing24\times40$ stages; $8\times30$ keyway and $\varnothing$6
cross-hole. & 6 \\
17 & \texttt{d06\_u\_clevis}: U clevis &
$40\times73\times30$; 16-wide slot, R15 ear tops around a $\varnothing$10
pin hole, plus a $\varnothing$24 boss with a $\varnothing$12 blind bore. & 7 \\
18 & \texttt{d07\_eccentric\_cam}: eccentric cam &
$\varnothing80\times25$ cam with $\varnothing$20 bore offset 10,
$\varnothing$36 hub, keyway, and two $\varnothing$12 holes at 28 pitch. & 7 \\
19 & \texttt{d09\_angle\_bracket}: angle bracket &
$60\times40\times100$ L-section with 8-mm legs and R5 inside corner; two
$22\times8$ slots and two $\varnothing$9 side holes. & 7 \\
20 & \texttt{d10\_clamp\_plate}: clamp plate &
$100\times70\times15$ plate; $40\times20$ window, four $\varnothing$9
holes on $76\times40$ with $\varnothing$15 counterbores, and edge rebate. & 7 \\
21 & \texttt{d11\_dovetail\_slide}: dovetail slide &
$80\times40\times60$ block; 24-wide, 15-deep $60^\circ$ dovetail, top slot,
and two $\varnothing$8 blind holes at 40 pitch. & 6 \\
22 & \texttt{d12\_cut\_block}: chamfered support &
$70\times50\times40$ stepped block with a $45^\circ$ top cut,
$16\times25$ slot, and a transverse $\varnothing$10 hole. & 7 \\
23 & \texttt{d18\_gear\_rack}: gear rack &
$120\times25\times15$; 12 teeth at 10 pitch with $20^\circ$ flanks, plus
two $\varnothing$6.6 holes with bottom $\varnothing$11 counterbores. & 7 \\
24 & \texttt{d21\_t\_slot\_block}: T-slot block &
$80\times40\times50$; through T-slot with 12-wide mouth and 24-wide lower
portion, plus two $\varnothing$9 holes at 60 pitch. & 7 \\
25 & \texttt{d22\_flanged\_bushing}: flanged bushing &
$\varnothing60\times38$; 8-thick flange, $\varnothing$40 sleeve,
$\varnothing$25 bore, end counterbore, radial oil hole, and four flange
holes. & 8 \\
26 & \texttt{d23\_cross\_plate}: cross plate &
$120\times120\times10$ cross with 30-wide arms, four R5 inner corners,
$\varnothing$30 center bore, and four $\varnothing$8 arm holes. & 7 \\
27 & \texttt{d24\_u\_channel}: U-channel seat &
$100\times48\times60$ channel with 8-mm base and walls, a $40\times10$
top slot, and two $\varnothing$12 wall holes at 60 pitch. & 7 \\
28 & \texttt{d25\_step\_pin}: headed pin &
66 long; $\varnothing24\times6$ head, $\varnothing$16 shank,
$\varnothing$13 groove, C1 end chamfer, and $\varnothing$4 cross-hole. & 8 \\
29 & \texttt{d26\_square\_flange}: square flange &
$80\times80\times30$ with 10-thick R10 base, $\varnothing$40 boss,
$\varnothing$20 bore, and four $\varnothing$9 holes on a $60\times60$
pattern. & 8 \\
30 & \texttt{d27\_guide\_rail}: guide rail &
$150\times25\times40$ with 40-wide base and 20-wide rail; three
$\varnothing$6.6 holes at 60 pitch with $\varnothing$11 counterbores. & 6 \\
31 & \texttt{d35\_shaft\_collar}: split collar &
$\varnothing40\times14$ ring, $\varnothing$20 bore, 2-mm radial slit, and
a transverse $\varnothing$4.5 clamp hole through the slit. & 6 \\
\bottomrule
\end{tabular}
\end{table*}

\begin{table*}[t]
\ContinuedFloat
\caption{Three-view reconstruction cases 32--50 (continued). Dimensions are in millimeters.}
\label{tab:suppl-threeview-cases-cont}
\centering
\footnotesize
\renewcommand{\arraystretch}{1.03}
\setlength{\tabcolsep}{3pt}
\begin{tabular}{r >{\raggedright\arraybackslash}p{1.35in}
                  >{\raggedright\arraybackslash}p{4.75in} r}
\toprule
\textbf{\#} & \textbf{Case / target} & \textbf{Defining geometry visible in the drawing} &
\textbf{Items} \\
\midrule
32 & \texttt{d02\_bearing\_block}: bearing block &
$120\times80\times50$; 14-thick base, 60-wide R30 bearing body with
$\varnothing$35 bore, vertical oil hole, and two $\varnothing$12 base holes. & 10 \\
33 & \texttt{d03\_rib\_bracket}: ribbed bracket &
$80\times70\times60$; 10-mm base and upright, R10 top corners,
$\varnothing$20 hole, 6-mm gusset, and two $\varnothing$10 base holes. & 9 \\
34 & \texttt{d08\_pipe\_tee}: flanged pipe tee &
$\varnothing40$ run with $\varnothing$30 bore, $\varnothing30$ branch with
$\varnothing$22 bore, and $\varnothing60\times8$ drilled end flanges. & 7 \\
35 & \texttt{d13\_v\_pulley}: double V-pulley &
$\varnothing100$ pulley with $\varnothing40$ hub and $\varnothing20$ bore;
two 12-wide, $38^\circ$ grooves, four holes, and a keyway. & 9 \\
36 & \texttt{d14\_link\_rod}: connecting link &
120 center distance; $\varnothing60/\varnothing36$ and
$\varnothing32/\varnothing16$ ends joined by common-tangent sides, with
two shallow side pockets. & 7 \\
37 & \texttt{d15\_motor\_mount}: motor mount &
$80\times86\times50$ base/upright bracket; $\varnothing$22 center hole,
four motor holes, two base slots, and two 6-mm gussets. & 8 \\
38 & \texttt{d16\_heat\_sink}: heat sink &
$70\times30\times50$ with 6-mm base, seven 2-mm fins at 8 pitch, and four
$\varnothing$4.5 mounting holes. & 6 \\
39 & \texttt{d17\_pipe\_clamp}: pipe clamp &
Two separate solids (strap and base); $100\times55\times20$ overall, R30
outside around a $\varnothing$40 bore, 2-mm split, and two bolt holes. & 7 \\
40 & \texttt{d19\_hinge\_leaf}: hinge leaf &
$60\times50\times3$ plate with three 10-long knuckles around a
$\varnothing$3 pin bore and three $\varnothing$5 countersunk holes. & 6 \\
41 & \texttt{d20\_ribbed\_base}: ribbed base &
$140\times35\times90$; 10-mm base with bottom pocket, four bosses, two rib
families, and a central $\varnothing$30 boss with blind bore. & 7 \\
42 & \texttt{d28\_round\_cup}: flanged cup &
$\varnothing96\times30$; $\varnothing80$ wall, $\varnothing72\times24$
cavity, one-sided $\varnothing$12 side hole, and three bottom holes. & 7 \\
43 & \texttt{d29\_hanger\_bracket}: hanger bracket &
$80\times70\times40$; 10-mm top plate, two R20 legs separated by 44, a
$\varnothing$16 pin hole, and two top mounting holes. & 7 \\
44 & \texttt{d30\_valve\_body}: valve body &
$70\times60\times50$ block with horizontal $\varnothing$20 passage,
vertical $\varnothing$16 port and $\varnothing$30 boss, plus four blind
front holes. & 7 \\
45 & \texttt{d31\_lever\_arm}: lever arm &
80- and 50-long, 12-wide arms with three hubs; keyed $\varnothing$14 center
bore and $\varnothing$8/$\varnothing$10 end holes. & 7 \\
46 & \texttt{d32\_tool\_holder}: tool holder &
$130\times30\times40$ overall; four $\varnothing$12 blind tool holes at 22
pitch and two end ears with $\varnothing$8 mounting holes. & 7 \\
47 & \texttt{d33\_shell\_housing}: housing cover &
$110\times40\times50$ open-bottom shell with 5-mm walls and 6-mm top,
$\varnothing$20 top opening, and two drilled mounting ears. & 8 \\
48 & \texttt{d34\_fixture\_plate}: fixture plate &
$120\times80\times15$ R5 plate with $40\times30\times6$ pocket, six
$\varnothing$8 grid holes, and two $\varnothing$6 blind locating holes. & 7 \\
49 & \texttt{d36\_flanged\_tube}: double-flanged tube &
60 long; $\varnothing40/\varnothing30$ tube between an R-square and a round
$\varnothing70$ flange, with distinct four-hole patterns at each end. & 7 \\
50 & \texttt{d37\_column\_base}: column base &
$\varnothing80\times12$ base and $\varnothing30$ column, R5 root, top
$\varnothing$10 blind hole, transverse $\varnothing$8 hole, and four base
holes. & 9 \\
\bottomrule
\end{tabular}
\end{table*}

\subsection{Motion-aware assembly tasks}
\label{sec:suppl:assembly-bench}

\Cref{tab:suppl-assembly-cases,tab:suppl-assembly-cases-new}
lists all 50 cases. ``Roles'' counts component occurrences, so repeated pins
and links count separately. $M$ is the target instantaneous mobility. Motor
cases are sampled over one or three input revolutions, as specified; pose
cases are evaluated at three prescribed configurations. Every case has a
component-interface rubric and a role-level assembly rubric. 

The mutually exclusive assembly types in the main paper correspond to the
following case numbers. Rotary transmissions, geared mechanisms, and
couplings are cases 1, 5, 7, 14--18, 20, 32, and 37--40. Reciprocating and
motion-conversion machines are cases 2, 6, 10, 11, 13, 19, 28--31, and 41.
Closed-loop linkages and path generators are cases 8, 9, 12, 23, 33--36,
44, and 45. Articulated robots and manipulators are cases 3, 21, 22, 46,
47, and 49. Deployable, steering, lifting, and gripping systems are cases
4, 24--27, 42, 43, 48, and 50.

\begin{table*}[t]
\caption{Assembly cases 1--27. ``Roles'' counts component occurrences, so
repeated pins and links count separately. $M$ is the target instantaneous
mobility. Kinematic quantities in the last column are evaluated independently
of the agent.}
\label{tab:suppl-assembly-cases}
\centering
\footnotesize
\renewcommand{\arraystretch}{1.08}
\setlength{\tabcolsep}{2.3pt}
\begin{tabular}{r >{\raggedright\arraybackslash}p{1.08in}
                  >{\raggedright\arraybackslash}p{3.30in} c c
                  >{\raggedright\arraybackslash}p{1.50in}}
\toprule
\textbf{\#} & \textbf{Case} & \textbf{Required mechanism} &
\textbf{Roles} & \textbf{$M$} & \textbf{Motion validation} \\
\midrule
1 & Two-stage gearbox &
Three shafts in a grounded housing; module-3 external pairs
$12{:}24$ and $12{:}36$ at 54 and 72\,mm centers. &
8 & 1 & Output $+1/6$; zero interference. \\
2 & Single-cylinder engine &
Crank--slider with 25-mm throw; $16{:}32$ external timing pair drives
the camshaft. &
12 & 1 & 50-mm piston stroke; cam $-1/2$. \\
3 & Adjustable desk lamp &
Grounded base, two serial arms, and shade connected by three revolute
joints. &
7 & 3 & Three shoulder--elbow--wrist poses. \\
4 & Folding umbrella &
Shaft, sliding runner, and two rib--stretcher closed loops attached to a
fixed upper hub. &
7 & 1 & Three runner heights and rib-tip positions. \\
5 & Drill planetary gearbox &
Two fixed-ring planetary stages, $12/18/48$ and $15/15/45$, with one
planet per stage. &
11 & 1 & Carriers $+1/5$ and $+1/20$. \\
6 & Sewing-machine drive &
Needle crank--slider, $40{:}20$ hook gears, and a feed
crank--rocker driven from one main shaft. &
15 & 1 & 36-mm needle stroke; hook $-2$. \\
7 & Film-projector drive &
Idler-synchronized shutter and a $16{:}64$ reduction from the same drive
shaft to the film sprocket. &
12 & 1 & Shutter $+1$; sprocket $-1/4$. \\
8 & Crank--rocker linkage &
Four-bar with crank, coupler, rocker, and ground lengths
20/70/50/80\,mm. &
8 & 1 & Rocker swing $47.39^\circ$. \\
9 & Offset slider--crank &
15-mm crank and 60-mm rod driving a slider on an axis offset by 10\,mm. &
7 & 1 & Slider stroke 30.46\,mm. \\
10 & Twin-cylinder pump &
Two slider--cranks on opposite sides of a plate, sharing 180-degree
opposed cranks. &
12 & 1 & Both strokes 30\,mm; antiphase. \\
11 & Pumpjack drive &
$10{:}30$ external reduction driving a crank--rocker walking beam. &
11 & 1 & Crank $-1/3$; beam swing $47.71^\circ$. \\
12 & Windshield wiper &
Crank--rocker followed by a parallelogram that carries a second wiper arm. &
13 & 1 & Both arm swings $44.19^\circ$. \\
13 & Beam engine &
Crank-driven walking beam whose opposite end drives a piston rod and
slider. &
12 & 1 & Beam swing $40.01^\circ$; stroke 40.94\,mm. \\
14 & Clock motion works &
Compound $10{:}30$ and $8{:}32$ external train, with the hour tube
coaxial with the minute arbor. &
10 & 1 & Intermediate $-1/3$; hour hand $+1/12$. \\
15 & Hand winch &
Hand crank and $12{:}36$ external reduction rigidly driving a cable drum. &
8 & 1 & Drum $-1/3$. \\
16 & Ring-drive turntable &
12-tooth pinion internally meshed with a 48-tooth ring fixed to the
turntable. &
6 & 1 & Turntable $+1/4$. \\
17 & Star-gear reducer &
Sun drives a planet on a fixed axis; the planet internally drives the
output ring. &
8 & 1 & Ring output $-1/4$. \\
18 & Lead-screw stage &
Vertical screw with 2-mm lead drives a carriage constrained by a parallel
guide rod. &
6 & 1 & Translation 2\,mm per revolution. \\
19 & Geared slider--crank &
$12{:}36$ reduction drives a centered crank--slider mechanism. &
10 & 1 & Crank $-1/3$; slider stroke 40\,mm. \\
20 & Planetary carrier output &
Fixed 72-tooth ring, 18-tooth sun, 27-tooth planet, and carrier output. &
8 & 1 & Carrier $+1/5$. \\
21 & Excavator arm &
Boom, stick, and bucket with a cylinder--rod closure between frame and
boom. &
11 & 3 & Three commanded arm configurations. \\
22 & SCARA arm &
Two serial revolute links followed by a coaxial vertical cylindrical
joint. &
6 & 4 & Three planar-reach and height poses. \\
23 & Pantograph &
Four-bar parallelogram linkage arranged as a two-times geometric
pantograph. &
10 & 2 & Three poses with $2{\times}$ point mapping. \\
24 & Ackermann steering &
Two steering knuckles joined by a tie rod in a steering trapezoid. &
8 & 1 & Three steering inputs; both knuckle pivots. \\
25 & Toggle clamp &
Ground link, handle, coupler, and clamping arm in a toggle four-bar. &
8 & 1 & Three handle-angle poses. \\
26 & Parallel gripper &
Two parallelogram-guided jaws with independent symmetric or asymmetric
actuation. &
15 & 2 & Three jaw poses, including asymmetric. \\
27 & Double-wishbone suspension &
Planar upper and lower control arms locate one upright relative to the
chassis. &
10 & 1 & Three wheel-travel positions. \\
\bottomrule
\end{tabular}
\end{table*}

\begin{table*}[t]
\caption{Assembly cases 28--50. ``Roles'' counts component occurrences, so
repeated pins and links count separately. $M$ is the target instantaneous
mobility.
}
\label{tab:suppl-assembly-cases-new}
\centering
\footnotesize
\renewcommand{\arraystretch}{1.08}
\setlength{\tabcolsep}{2.3pt}
\begin{tabular}{r >{\raggedright\arraybackslash}p{1.08in}
                  >{\raggedright\arraybackslash}p{3.30in} c c
                  >{\raggedright\arraybackslash}p{1.50in}}
\toprule
\textbf{\#} & \textbf{Case} & \textbf{Required mechanism} &
\textbf{Roles} & \textbf{$M$} & \textbf{Motion validation} \\
\midrule
28 & L-twin engine &
Perpendicular 15-mm-throw cylinders share one crank pin; a $16{:}32$
timing pair drives the camshaft. &
15 & 1 & Both strokes 30\,mm; camshaft $-1/2$. \\
29 & Four-cylinder radial engine &
One master rod and three articulating rods connect four radial pistons to
a common crank and propeller shaft. &
21 & 1 & Strokes 30/30.25/30.25/30\,mm. \\
30 & Quick-return shaper &
Crank-driven slotted lever and connecting rod drive a guided ram; a
$30{:}15$ pair connects the handwheel. &
15 & 1 & Lever swing $49.25^\circ$; ram stroke 116.67\,mm. \\
31 & Knuckle-joint press &
A $14{:}42$ reduction drives a knuckle linkage that translates the press
ram near toggle. &
15 & 1 & Knuckle swing $36.16^\circ$; ram stroke 12.49\,mm. \\
32 & Oldham coupling &
Two parallel shafts offset by 12\,mm connect through a cross-shaped middle
disc with orthogonal sliding pairs. &
12 & 1 & Output $+1$; middle-disc X/Y strokes 12\,mm. \\
33 & Elliptical trainer &
Opposed cranks drive left and right pedal links and coupled swinging
handles in antiphase. &
18 & 1 & Handle swing $29.06^\circ$; pedal travel and antiphase. \\
34 & Jansen walking leg &
Strandbeest-proportion eight-bar leg driven by a geared crank. &
22 & 1 & Foot path $135.8{\times}44.9$\,mm and link swing. \\
35 & Chebyshev lambda conveyor &
A geared Chebyshev lambda linkage produces an approximately straight
transfer segment for a stepping conveyor. &
19 & 1 & Gear ratio and foot path; 65\% near-linear stroke. \\
36 & Geared five-bar plotter &
Equal input gears synchronize two grounded cranks of a closed five-bar
drawing linkage. &
13 & 1 & Gear ratio and pen-tip path. \\
37 & Planetary mixer &
A geared carrier orbits two planets around a fixed sun, with beaters rigidly
attached to the planets. &
14 & 1 & Over three input turns: carrier $-1/3$, planets $-2/3$; 96-mm path. \\
38 & Spirograph &
Two 20-tooth planets carrying pens roll inside a fixed 60-tooth ring on a
geared carrier. &
16 & 1 & Carrier $-1/3$; planets $+2/3$. \\
39 & Wankel rotary mechanism &
A 12-mm eccentric drives a triangular rotor constrained by a stationary
gear and an internal rotor gear. &
11 & 1 & Rotor $+1/3$; 24-mm eccentric travel. \\
40 & Earth--Moon orrery &
A fixed sun, idler, and Earth gear form a three-mesh train carried by a
geared orbital arm. &
16 & 1 & Over three input turns: arm $-1/3$, idler $-1$, Earth $+1/6$;
128-mm orbit. \\
41 & Oscillating fan &
Two $12{:}36$ reductions drive a crank--rocker that oscillates the fan head. &
15 & 1 & Intermediate shaft $-1/3$; head swing $60.14^\circ$. \\
42 & Gear--screw lift &
A gear pair synchronizes two 6-mm-lead screws that raise a guided platform. &
12 & 1 & Gear ratio; platform rises 3\,mm per input turn. \\
43 & Three-stage scissor lift &
Three stacked scissor stages connect a guided base slider to a
horizontally constrained top platform. &
21 & 1 & Three slider offsets; platform and stage-joint positions. \\
44 & Five-blade mechanical iris &
A rotating drive ring actuates five overlapping four-bar blade linkages
arranged in three depth layers. &
28 & 1 & Three ring angles and all five blade-tip positions. \\
45 & Peaucellier linkage &
A rhombus and equal-length links form an exact straight-line Peaucellier
cell driven by a crank. &
14 & 1 & Three crank angles; output remains on $x=93.75$\,mm. \\
46 & Five-bar parallel robot &
Two independently actuated grounded links drive a closed five-bar
end-effector chain. &
11 & 2 & Three input-angle pairs; both elbows and end effector. \\
47 & Palletizing robot &
A driven parallelogram and two leveling parallelograms maintain the wrist
flange orientation. &
18 & 2 & Three input pairs; elbow, wrist, and flange pose. \\
48 & Gear-synchronized gripper &
Equal gears synchronize two parallelogram-guided jaws for symmetric
single-DOF closure. &
17 & 1 & Three openings: 79.4/47.9/8.1\,mm. \\
49 & Coupled robot finger &
Two anti-parallelogram linkages couple the proximal, middle, and distal
phalanges. &
13 & 1 & Three proximal angles; middle endpoint and fingertip. \\
50 & Level-luffing crane &
A four-bar boom, rocker, and horse-head linkage is actuated by a closing
hydraulic-cylinder pair. &
14 & 1 & Three boom angles; hook, boom head, and cylinder joint. \\
\bottomrule
\end{tabular}
\end{table*}

\clearpage
\clearpage

\section{Evaluation methodology details}
\label{sec:suppl:evaluation}

\subsection{Text-to-part evaluation and validation}
\label{sec:suppl:rubric}
\label{sec:suppl:rubric-validation}

Text-to-part evaluation combines a deterministic geometric checker with an
independent validation suite. The evaluator receives each submitted model in
STEP format and evaluates it using OpenCASCADE, an open-source CAD geometry
kernel that provides B-rep parsing, topology and geometry queries, and shape
analysis. The checker operates only on the resulting B-rep, without access
to the feature tree, sketches, construction history, or agent report.
\Cref{tab:suppl-checks} summarizes its deterministic checks for topology,
geometric relations, material distribution, symmetry, and volume. We then
validate every check against independently built references, admissible
alternative constructions, targeted defects, and failures observed in agent
runs.

\paragraph{Geometric check example.}

Listing~\ref{lst:end-cap-rubric} shows the checks for the end cap task (task 27),
which is graded by five explicit items. Entities select
the $\varnothing$60 outer cylinder, the $\varnothing$50 cavity, and the
$\varnothing$8 hole; \texttt{cframe} is the axis frame of the outer
cylinder with $t{=}0$ on the end with the larger planar area, \ie the top
plate.
The \texttt{line\_material} check casts a line along the axis at radius 15
and requires material exactly on $[0,3]$ in the frame, which verifies the
plate thickness and that the cavity is open at the bottom. The
\texttt{radial\_material} checks cast a ray from the axis at a given height
and verify the wall ($[25,30]$) and the hole ($[4,30]$).

\begin{nolinenumbers}
\begin{lstlisting}[
  float,
  floatplacement=t,
  caption={Example: geometric checks for the end-cap task.},
  label={lst:end-cap-rubric}
]
entities:
  outer:  {cylinder: {d: 60, side: outer}, count: 1}
  cavity: {cylinder: {d: 50, side: inner}, count: 1}
  hole:   {cylinder: {d: 8,  side: inner}, count: 1}
  cframe: {axis_frame: {axis: outer,
                        low: larger_area}}
rubrics:
# implied by "a part"
- id: solid
  hard: true
  check: [single_valid_solid: {}]
# "OD 60, total height 15"
- id: outer
  hard: true
  check:
  - exists: outer
  - extent_along: {axis: outer, of: outer,
                   value: 15}
  - bbox_sorted: [15, 60, 60]
# "cavity D50, 12 deep from below,
#  leaving a 3-mm top plate"
- id: cavity
  check:
  - exists: cavity
  - coaxial: [cavity, outer]
  - pocket_depth: {pocket: cavity, value: 12,
                   blind: true, probe_r: 15,
                   open_at_extreme: true}
  - line_material: {frame: cframe, along: 0,
                    polar: [15, 0],
                    expect: [[0, 3]]}
  - radial_material: {frame: cframe, t: 9,
                      expect: [[25, 30]]}
# "D8 through-hole in the top plate"
- id: hole
  check:
  - exists: hole
  - coaxial: [hole, outer]
  - through: {holes: hole}
  - radial_material: {frame: cframe, t: 1.5,
                      expect: [[4, 30]]}
# pi(30^2*15 - 25^2*12 - 4^2*3)
- id: volume
  check: [volume: 18698.8]
\end{lstlisting}
\end{nolinenumbers}

\begin{table*}[t]
\caption{Check families of the geometric checker. Every check is a
deterministic predicate on the exported B-rep; coordinates are expressed in
part frames recovered from the solid.}
\label{tab:suppl-checks}
\centering
\footnotesize
\renewcommand{\arraystretch}{1.12}
\setlength{\tabcolsep}{4pt}
\begin{tabular}{l >{\raggedright\arraybackslash}p{5.35in}}
\toprule
\textbf{Family} & \textbf{Checks (examples) and what they verify} \\
\midrule
Selectors and frames
  & Cylinders by diameter and side, cones by semi-angle, tori, spheres,
    prismatic pockets by sides and across-flats size, bounding-box axes.
    Principal, world, axis, and base frames; an axis frame can be oriented
    toward a second entity or along its direction. Surfaces of the same
    analytic geometry are merged across faces, so the result does not
    depend on how the modeler split them. \\
Topology and envelope
  & \texttt{single\_valid\_solid}, \texttt{genus}, \texttt{bbox\_sorted}
    (with alternatives), \texttt{frame\_box}, \texttt{footprint},
    \texttt{total\_height}, \texttt{extreme\_plane},
    \texttt{bottom\_planar}. \\
Features and relations
  & \texttt{exists} with counts, \texttt{coaxial}, \texttt{parallel},
    \texttt{parallel\_to\_frame}, \texttt{axis\_positions},
    \texttt{axis\_distance}, \texttt{equal\_angular}, \texttt{through},
    \texttt{pocket\_depth} (blind, open at an end face), \texttt{recess}
    (counterbore or countersink, all on one side), \texttt{cone\_size}
    (chamfers), \texttt{blend\_faces} (fillets near an edge),
    \texttt{parallel\_faces}, \texttt{stepped\_length},
    \texttt{sweep\_angle}, \texttt{flanges\_at\_arc\_ends},
    \texttt{end\_face\_shape}, \texttt{pits} (die faces and opposite
    sums). \\
Material probes
  & \texttt{line\_material}, \texttt{radial\_material},
    \texttt{ray\_thickness}, \texttt{first\_hit}, \texttt{material\_at},
    \texttt{no\_material\_in}: exact material intervals along a line or
    ray. \texttt{angular\_openings}: circumferential scans that count
    grooves and slots and measure their arc, chord, and radial extent.
    \texttt{opening\_size}, \texttt{edge\_notches}, \texttt{section\_areas},
    \texttt{revolved\_profile}. Phone-stand checks locate the base, the
    inclined plate, and its slots and notch from the solid. \\
Symmetry and volume
  & \texttt{mirror\_symmetric} and \texttt{rotation\_invariant} by the
    volume of the symmetric difference; \texttt{volume} (exact, 1\%) and
    \texttt{volume\_range}. \\
\bottomrule
\end{tabular}
\end{table*}

\paragraph{Checks validation.}

Some check targets are derived rather than copied directly from a brief,
including total height, chord width, bend allowance, and volume. The rubric
records each derivation next to the check so that the target can be audited.
All 100 reference oracle models score 100 by the checker, and each
volume agrees with its rubric target within 0.05\%.

We also test admissible alternatives and targeted failures. Twelve easy-split
variants and one medium-split impeller variant exercise reasonable readings
that the brief permits, such as an extra socket fillet or a key-shaft notch
truncated by the part end; all must score 100. The 62 easy, 73 medium, and 71
hard mutants each violate one or more known requirements, such as moving a
hole by 5\,mm, changing a feature count, reversing a flange or helix,
converting a blind hole to a through hole, or using a wrong bend allowance.
Each test asserts the exact set of rubric items that should fail, detecting
both checks that fire too often and checks that miss a defect.
Agent runs provide new model variants, and manual inspection of these runs
yields additional checks, thereby making our check suite increasingly
comprehensive.


\subsection{Three-view evaluation and validation}
\label{sec:suppl:threeview-metric}

\paragraph{Image score: three-view projection metric.}
Below, we detail how the Image Score measures the discrepancy between the
submitted model's projections and the reference three-view projections.
Let \(L\) be the largest side of the reference solid's axis-aligned bounding
box. We triangulate the reference and submitted solids and rasterize each
along the three coordinate axes at a pixel pitch of \(L/400\). For every
viewing ray, the rasterizer records occupancy, first and last intersection
depths, and total material thickness. Projected bounding boxes are centered
on a common canvas, and depths are measured relative to each solid's
bounding-box face, making the comparison invariant to translation.

For a candidate rotation \(R\) and view \(v\), let \(U_v(R)\) be the
effective union of pixels occupied by either projection. To avoid penalizing
small silhouette shifts caused by rasterization, we remove from this union a
pixel occupied by only one solid when the other projection contains an
occupied pixel in its \(3\times3\) neighborhood. Let
\(M_v(R)\subseteq U_v(R)\) contain the remaining pixels that are occupied by
both projections and agree in all three channels: first depth, last depth,
and total material thickness.

We set the channel tolerance to \(\epsilon=0.005L\). A channel agrees when
its two values differ by at most \(\epsilon\) at the same pixel, or when one
value lies within the occupied-value range of the other projection in the
\(3\times3\) neighborhood, expanded by \(\epsilon\). The per-view score and
the final score are
\begin{equation}
\begin{aligned}
S_v(R) &= \frac{|M_v(R)|}{|U_v(R)|},\\
\operatorname{ImageScore}
&=100\max_{R\in\mathcal{R}_{24}}
\left(\frac{1}{3}\sum_{v\in\{x,y,z\}}S_v(R)\right),
\end{aligned}
\label{eq:threeview-image-score}
\end{equation}
Here, \(\mathcal{R}_{24}\) denotes the 24 proper axis-aligned rotations;
reflections are not included. Before computing pixel-level scores, we compare
the bounding-box extents under all candidate rotations and discard any
rotation whose extent error is more than \(0.05L\) above the minimum. Among
the retained rotations, we select the one with the highest mean score
across the three views, as defined in
\cref{eq:threeview-image-score}. Within each
\(S_v(R)\), the denominator measures all image regions that the submission
should explain, while the numerator counts the regions whose occupancy and
depth structure match the reference.

We validate the image score with all 50 independently constructed reference
models and 133 targeted mutants containing moved holes, incorrect depths or
angles, missing features, wrong pattern counts, and other
single-requirement defects. Every reference scores 100, while every mutant
scores below 99.5. Properly rotated references retain a score of at least
99.9, whereas a mirrored asymmetric control remains distinguishable. These
tests verify that the metric tolerates orientation and rasterization effects
without hiding meaningful geometric errors.

\paragraph{Checks validation.}
Each drawing is generated from an independently constructed OpenCASCADE
reference and annotated from a case-specific dimension list. We validate the
drawing-grounded checks separately from the image score: all 50 references
pass every check and receive a score of 100. For the same 133 targeted
mutants, each test asserts the exact set of rubric items that should fail,
detecting both checks that fire on valid geometry and checks that miss the
intended defect. In total, 368 automated tests cover references, mutants,
rotations, rendered images, drawing-grounded checks, and released assets.

\subsection{Assembly evaluation and validation}
\label{sec:suppl:assembly-validation}

\subsubsection{Scoring protocol}
As described in the main text, assembly evaluation combines a structure score with one of two motion
scores, depending on whether the task specifies a continuous rotary input or
three target configurations. We now detail the scoring protocol in the following paragraphs.

\noindent\textbf{Assembly structure.}
The evaluator reopens the delivered assembly and reads component transforms,
mate definitions, and the geometry referenced by every mate. A
screw-theory-based constraint matrix scores six requirements: rigid-body
grouping (20 points), semantic joint types (25 points), total mobility (15 points),
grounding and absence of floating components (10 points), absence of over-defined
or failed mates (10 points), and gear coupling (including ratio, direction, axis
alignment, and center distance, 20 points). Different native mate decompositions
are accepted when they impose the same semantic joint.

\noindent\textbf{Motor-driven motion.}
The 33 cases with one rotary input are driven independently by the evaluator.
Kinematic trajectories normally use 121 samples over one input revolution;
the planetary mixer and Earth--Moon orrery use 361 samples over three
revolutions because their orbital motion completes over multiple turns.
Native interference is checked at 13 evenly spaced frames. The evaluator
measures required ratios, leads, strokes, swings, and antiphase relations.
The motor-driven score is 100 times a weighted pass rate: task-specific
kinematics and collision-free motion each have weight 2, while movement of
components that should move and stability of fixed components each have
weight 1. Interference and fixed-body credit are multiplied by the fraction
of expected moving components that actually move, preventing a jammed
mechanism from scoring well merely because it remains collision-free.

\noindent\textbf{Prescribed-pose motion.}
The remaining 17 cases are evaluated at three prescribed configurations.
For each pose, key-point accuracy, interference, and fixed-body stability
have weights 2, 2, and 1. Interference and stability receive credit only
after the key-point check passes; a missing required component or sampling
failure yields zero. The final score is the mean over the three poses.

\subsubsection{Assembly Evaluator Validation}
We validate the evaluator at three levels: testing the evaluator logic, 
checking reference assemblies offline, and validating reference assemblies in SolidWorks.

\noindent\textbf{Testing the evaluator logic.}
We first test the structure evaluator on synthetic assemblies whose component
transforms and mate records are known exactly. For example, we construct such
fixtures for external and internal gear trains, a planetary stage, a lead
screw, and linkage mechanisms. From these fixtures, we inject one fault at a time:
for example, unlocking a nominally rigid shaft--gear connection, deleting a coupling,
disconnecting an output group, marking a mate as failed, reversing a gear
direction or tooth ratio, shifting a gear center, marking every component as fixed so that the mechanism cannot move, or
removing components and mates. For each injected fault, the evaluator should
both lower the total score and identify the corresponding problem. Examples
include excessive freedom of motion, a disconnected component group, an
incorrect rigid connection, and an invalid gear ratio, rotation direction,
or center distance. Empty, fully locked, and half-built assemblies should
score near zero, while partially correct assemblies should receive credit
only for the requirements they satisfy.

We test the motion evaluator separately using synthetic trajectories whose
expected rotations, translations, and key-point positions can be calculated
exactly from the mechanism definition. These tests check whether it correctly
measures gear ratios and rotation directions (including internal gears),
screw lead, linear travel, swing angle, opposite-phase motion, and key-point
positions defined in a transformed ground frame. They also cover difficult
numerical cases, such as very small ratios that require relative tolerance,
mirrored coordinates, fast drivers that require denser adaptive sampling,
and mechanisms whose output is stuck. A stuck mechanism should not receive
motion credit simply because it has no interference and its fixed components
remain stationary. These tests validate the reusable motion-scoring logic,
rather than creating a separate set of faulty variants for each of the 50
benchmark mechanisms.

\noindent\textbf{Checking reference assemblies offline.}
For all assembly reference oracle models, structured definitions of the intended assembly geometry and motion undergo four
offline consistency checks before any live SolidWorks validation. Across all sampled
configurations, (i) joint constraints must hold, including coincident hole
centers, fixed relative angles, slider offsets, gear center distances and
pure rolling, and screw lead; (ii) planar part profiles, holes, and depth
layers must remain free of unintended intersections; (iii) components and
mates 
must score 100 under
the structure evaluator; and
(iv) synthesized full-cycle trajectories must score 100 under the motion
evaluator, or, for pose-driven cases, satisfy every prescribed pose. These
offline checks cannot expose closed-loop branch changes or solver timeouts,
so they are followed by live SolidWorks validation.

\noindent\textbf{Validating reference assemblies in SolidWorks.}
For all 50 assembly cases, we build an oracle reference assembly independently
of the evaluated agent runs. We run the
structure and motion evaluators for every reference assembly. 
All 50 reference assemblies score 100
on both evaluators and have zero sampled interference. Across the 17
prescribed-pose cases, the largest key-point discrepancy is
\(0.012\,\mathrm{mm}\).

\section{Comparison with existing CAD benchmarks}
\label{sec:suppl:bench-compare}

This section explains the entries of \cref{tab:bench-compare}. Each entry
reflects the evaluation protocol described in the cited paper or official
repository.

\paragraph{Column criteria.}
\textbf{Multimodal} is supported if the evaluated tasks give the model inputs
beyond text, such as images, engineering drawings, or meshes.
\textbf{Assembly} is supported if the benchmark evaluates complete
multi-component assemblies, and partially supported if it evaluates only a
single connection between two bodies.
\textbf{Ref.\ geometry} is supported if the score compares the output with a
reference shape at the shape level, \eg, by IoU, Chamfer distance, or
projection maps, and partially supported if the comparison uses only coarse
quantities such as bounding boxes or volumes.
\textbf{Motion checks} is supported if the benchmark drives a complete
mechanism or poses it at prescribed configurations and checks the resulting
kinematics, and partially supported if it tests the motion of a single joint
only.
\textbf{Fine-grained} is supported if each task is scored by checks tied to
its individual requirements.

\paragraph{Benchmark-specific notes.}
\begin{itemize}
    \item \textbf{Text2CAD.} It provides text prompts at four expertise levels
    for single DeepCAD parts and compares the output with the ground-truth
    model using global metrics such as Chamfer
    distance~\citep{khan2024text2cad}.

    \item \textbf{CADTests.} It is text-only and reference-free by design.
    Each prompt is paired with executable tests grouped by
    requirement~\citep{DBLP:journals/corr/abs-2605-07807}.

    \item \textbf{CADGenBench.} It combines a validity gate with shape
    similarity, interface keep-in/keep-out volumes, and topology
    matching~\citep{cadgenbench2026}. These aggregate metrics do not check
    each task requirement individually.

    \item \textbf{BenchCAD.} It reports IoU, Chamfer and Hausdorff distances,
    feature F1, and operation recall across four capability
    axes~\citep{DBLP:journals/corr/abs-2605-10865}, rather than checking each
    stated requirement individually.

    \item \textbf{CADBench.} It reports volumetric and surface IoU, Chamfer
    distance, valid-shape rate, token count, and operation
    count~\citep{DBLP:journals/corr/abs-2605-10873}. Its tasks use rendered
    images and meshes as inputs but contain only single-body designs, and its
    metrics do not check each stated requirement individually.

    \item \textbf{Op-CAD.} It reports Pass@$k$, Chamfer and Hausdorff
    distances, and chamfer/fillet edge IoU at the step and sequence
    levels~\citep{bai2026op}, rather than checking each requirement
    individually. Its five modalities describe the dataset, but both
    evaluated tasks use textual instructions; the sequence-level task also
    receives preceding CadQuery scripts. Images are used only in the
    annotation pipeline.

    \item \textbf{CADWorld.} Its part tasks check numerical properties of the
    saved model, such as bounding boxes, volumes, and surface areas, while its
    CAM tasks compare removed material with a
    target~\citep{dong2026cadworld}. Shape-level measures such as bounding-box
    IoU are reported only as diagnostics and do not enter the success score.
    Its 25 FreeCAD assembly tasks inspect joint metadata, such as grounded,
    slider, and rack-and-pinion joints, without driving the joints; its
    simulation tasks concern finite-element analysis.

    \item \textbf{CADEngBench.} Its assembly track ``evaluates recorded joints
    between body pairs rather than complete
    assemblies''~\citep{DBLP:journals/corr/abs-2608-09296}. Given two bodies,
    the model predicts the joint entities and joint frame; the resulting URDF
    joint is then tested in PyBullet for allowed and blocked motion. This
    verifies the mobility of one joint, but not coordinated motion in a
    complete mechanism, such as closed kinematic loops, gear or screw
    couplings, or interference between moving components. CADEngBench does
    not compare shape similarity; instead, it checks measured B-rep
    properties, parameter--geometry relations, and finite-element responses.

    \item \textbf{\bench (ours).} Inputs are text briefs and three-view
    engineering drawings. Its assembly track contains 50 complete mechanisms
    with 629 component occurrences, each driven or posed and checked for
    task-specific kinematics, interference, and drift
    (\cref{sec:bench:assembly}). Every track is scored with weighted
    per-requirement checks. The three-view track compares silhouette, depth,
    and thickness projections with a reference model
    (\cref{sec:bench:threeview}). The text-to-part and assembly tracks are
    reference-free because a brief can admit several valid designs.
\end{itemize}

Benchmarks based mainly on aggregate shape metrics, 
such as Text2CAD, CADGenBench, BenchCAD, CADBench, and Op-CAD, 
may assign a high score even when an output violates a specific stated dimension or feature.
As summarized in
\cref{tab:bench-compare}, \bench is the only benchmark in this comparison that
combines multimodal inputs, complete assemblies, reference-geometry
comparison, mechanism-level motion checks, and fine-grained
requirement-specific evaluation.

\section{Experimental setup and implementation details}
\label{sec:suppl:setup}
\label{sec:suppl:protocol}

\begin{table*}[t]
\caption{Final \method results by functional category on all three \bench
tracks. All runs use Qwen3.8-Flash-Next (180B). Tokens are category means per
task and include input, reasoning, and output.}
\label{tab:bykind}
\centering
\footnotesize
\renewcommand{\arraystretch}{1.06}
\setlength{\tabcolsep}{2pt}
\begin{tabular*}{\linewidth}{@{\extracolsep{\fill}}>{\raggedright\arraybackslash}p{2.45in}
                >{\centering\arraybackslash}p{0.42in}
                >{\centering\arraybackslash}p{0.52in}
                >{\centering\arraybackslash}p{0.46in}
                >{\centering\arraybackslash}p{0.58in}
                >{\centering\arraybackslash}p{0.65in}@{}}
\toprule
\multicolumn{6}{l}{\textbf{Text-to-part} (auxiliary metric: core-form rate)} \\
\textbf{Part type} & \textbf{Tasks} & \textbf{pass@1} & \textbf{Score} &
\textbf{Core form} & \textbf{Tokens} \\
\midrule
Daily objects & 9 & 0.89 & 98.9 & 1.00 & 122.1k \\
Structural / mounting / linkage & 36 & 0.92 & 97.3 & 0.97 & 157.6k \\
Axisymmetric / rotating / transmission & 35 & 0.86 & 95.1 & 0.97 & 191.4k \\
Housings / fluid / thermal & 20 & 0.85 & 96.3 & 1.00 & 177.0k \\
\textbf{All} & \textbf{100} & \textbf{0.88} & \textbf{96.5} &
\textbf{0.98} & \textbf{170.1k} \\
\midrule
\multicolumn{6}{l}{\textbf{Three-view-to-part} (auxiliary metric: image score)} \\
\textbf{Part type} & \textbf{Tasks} & \textbf{pass@1} & \textbf{Score} &
\textbf{Image} & \textbf{Tokens} \\
\midrule
Daily objects & 6 & 1.00 & 100.0 & 100.0 & 209.5k \\
Structural / mounting / linkage & 27 & 0.78 & 94.4 & 96.6 & 104.9k \\
Axisymmetric / rotating / transmission & 10 & 1.00 & 100.0 & 100.0 & 140.7k \\
Housings / fluid / thermal & 7 & 0.71 & 95.9 & 95.8 & 191.2k \\
\textbf{All} & \textbf{50} & \textbf{0.84} & \textbf{96.4} &
\textbf{97.6} & \textbf{136.7k} \\
\midrule
\multicolumn{6}{l}{\textbf{Assembly} (auxiliary metric: motion score)} \\
\textbf{Assembly type} & \textbf{Tasks} & \textbf{pass@1} & \textbf{Score} &
\textbf{Motion} & \textbf{Tokens} \\
\midrule
Rotary transmissions / geared mechanisms / couplings
& 14 & 0.43 & 94.4 & 73.6 & 2271.7k \\
Reciprocating / motion-conversion machines
& 11 & 0.55 & 93.0 & 86.4 & 2926.2k \\
Closed-loop linkages / path generators
& 10 & 0.30 & 83.4 & 43.3 & 3393.0k \\
Articulated robots / manipulators
& 6 & 0.83 & 97.6 & 83.3 & 2263.7k \\
Deployable / steering / lifting / gripping
& 9 & 0.56 & 88.1 & 55.6 & 2560.8k \\
\textbf{All} & \textbf{50} & \textbf{0.50} & \textbf{91.1} &
\textbf{68.3} & \textbf{2691.0k} \\
\bottomrule
\end{tabular*}
\end{table*}

\paragraph{System architecture and workflow interfaces.}
\method consists of a Python agent layer and a C\# service that exposes the
SolidWorks 2025 automation API through local HTTP endpoints. The service
handles document management, sketch and feature creation, face, body, and
model-state queries, and part and assembly operations. Each workflow composes
these atomic service operations and is exposed to the agent as a typed tool.
The agent can call only these workflow tools; it cannot invoke the underlying
service endpoints directly.

\paragraph{Workflow execution, validation, and recovery.}
\label{sec:suppl:errors}
Every geometry-changing workflow follows the same execution sequence. The
runtime resolves semantic references, validates typed arguments, executes
the underlying SolidWorks operations, and compares the model state before
and after execution. If execution fails or has no intended geometric effect,
the runtime removes partial changes and returns a machine-readable error code
with a correction hint.

The error code identifies the stage and cause of failure. Parameter errors
cover invalid syntax, values, expressions, and dependency cycles. Reference
errors distinguish malformed, missing, ambiguous, non-planar, and
non-axis-aligned entities. Profile errors report open, degenerate,
self-intersecting, overlapping, or misplaced loops, as well as revolve
profiles that cross their axis. Execution errors distinguish a failed native
feature, a feature with no geometric effect, and a disconnected feature that
was rolled back. Capture and session errors report unavailable views,
documents, bodies, or SolidWorks. At the agent-loop level, duplicate calls
are rejected, an identical call is stopped after two failures, and
\texttt{finish} is refused while multiple solid bodies remain. Together,
these checks turn low-level failures into specific recovery actions instead
of a generic ``feature failed'' message.

\paragraph{Agent and model configuration.}
The part and assembly agents operate in a closed loop. After each workflow
call, the LLM receives a compact structured state and may request labeled
native screenshots before selecting the next action. Only the latest image
is retained in the multimodal context; older images are replaced by short
placeholders. Checkpoints, traces, and final result records omit image
payloads. Recovery replays only state-changing calls, allowing an interrupted
run to resume without recapturing historical screenshots.

For the main results in the~\cref{tab:main}, we use the LLM
\texttt{Qwen3.8-Flash-Next-FP8}. We use temperature 0, medium reasoning effort, and
disable LLM caching. Text-to-part and three-view-to-part runs allow at most 40
LLM calls. In the assembly track, each generated part allows at most 40 calls
and the assembly stage allows at most 100 calls. 
The backbone comparison changes only the model while keeping
the Workflow interface and track-specific protocol fixed.

\paragraph{Atomic baseline.}
To isolate the contribution of the Workflow abstraction, the Atomic API
baseline uses the same SolidWorks service, Qwen backbone, and evaluator, but
does not use the Workflow agent loop. For text-to-part, it generates one
complete sequence of low-level SolidWorks operations and then executes that
sequence without observing intermediate geometry, correcting arguments, or
rolling back ineffective calls. Three-view-to-part first parses the drawing
and then generates the same type of one-shot atomic plan. Assembly likewise
uses a one-shot plan composed of component insertion, pose, grounding, and
individual mate operations; aggregated joint, loop-closure, coupling, and
motion-repair workflows are unavailable. 

\paragraph{Execution, evaluation, and reporting protocol.}
Evaluation is separated from agent execution. Part geometry is scored offline
from exported STEP files, while assembly evaluation reads the delivered
components and constraints and independently executes the prescribed
structural and motion checks. In every track, the evaluator receives only the
final artifact and task rubric, never the agent trajectory, intermediate
observations, or self-reported success.


\section{Detailed results}
\label{sec:suppl:detailed-results}

\subsection{Results by category level}
\label{sec:suppl:bykind}

\Cref{tab:bykind} breaks down the \method results with Qwen3.8-Flash-Next using the mutually
exclusive functional categories defined in
\cref{sec:suppl:benchmark-taxonomy}. Text-to-part
and three-view-to-part share the four part categories, while assembly tasks
use five kinematic categories. The auxiliary column reports core-form rate,
image score, and motion score for the three tracks, respectively.

On text-to-part, all four categories score above 95. Structural, mounting,
and linkage parts obtain the highest pass@1 (0.92), while daily objects
obtain the highest score (98.9). Daily objects and housings have perfect
core-form rates; most remaining failures therefore concern task-specific
details rather than the main shape. On three-view-to-part,
daily objects and axisymmetric, rotating, and transmission parts both reach
1.00 pass@1, whereas structural parts and housings are harder at 0.78 and
0.71, respectively.

The largest variation appears in assembly. Articulated robots and
manipulators achieve 0.83 pass@1, while closed-loop linkages and path
generators reach 0.30 pass@1 and a motion score of 43.3. The latter also use
the most tokens (3393.0k per task), consistent with the longer construction
and correction trajectories required to close kinematic loops.

\subsection{Results by difficulty level}
\label{sec:suppl:bydifficulty}

\Cref{tab:bydifficulty} reports the final \method results on the three
text-to-part difficulty splits defined in
\cref{sec:suppl:benchmark-tasks}.
Performance decreases as the tasks become harder: pass@1 falls from 0.96 on
easy tasks to 0.90 on medium tasks and 0.65 on hard tasks, while mean score
falls from 98.6 to 95.8 and 92.5. Core-form rates remain at least 0.97,
showing that most failures concern secondary requirements rather than the
main shape. Token use also increases consistently with difficulty, from
85.3k to 233.9k and 286.5k, reflecting longer modeling and correction
trajectories.

\begin{center}
\begin{minipage}{\columnwidth}
\captionof{table}{Final \method text-to-part results by difficulty. All runs use
Qwen3.8-Flash-Next (180B). Tokens are means per task and include input,
reasoning, and output.}
\label{tab:bydifficulty}
\centering
\footnotesize
\renewcommand{\arraystretch}{1.08}
\setlength{\tabcolsep}{1.5pt}
\begin{tabular*}{\linewidth}{@{\extracolsep{\fill}}>{\raggedright\arraybackslash}p{0.59in}
                >{\centering\arraybackslash}p{0.36in}
                >{\centering\arraybackslash}p{0.43in}
                >{\centering\arraybackslash}p{0.38in}
                >{\centering\arraybackslash}p{0.48in}
                >{\centering\arraybackslash}p{0.48in}@{}}
\toprule
\textbf{Difficulty} & \textbf{Tasks} & \textbf{pass@1} & \textbf{Score} &
\textbf{Core form} & \textbf{Tokens} \\
\midrule
Easy & 50 & 0.96 & 98.6 & 0.98 & 85.3k \\
Medium & 30 & 0.90 & 95.8 & 0.97 & 233.9k \\
Hard & 20 & 0.65 & 92.5 & 1.00 & 286.5k \\
\midrule
\textbf{All} & \textbf{100} & \textbf{0.88} & \textbf{96.5} &
\textbf{0.98} & \textbf{170.1k} \\
\bottomrule
\end{tabular*}
\end{minipage}
\end{center}

\section{Analysis}
\label{sec:suppl:analysis}

\subsection{Failure Cases}
\label{sec:suppl:failure-analysis}

\subsubsection{\method with Qwen3.8-Flash-Next}
\label{sec:suppl:qwen-failures}

\begin{figure}[!h]
  \centering
  \includegraphics[width=\columnwidth]{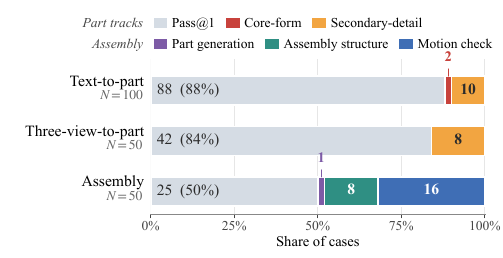}
  \caption{Case-level outcomes for Workflow-CAD with Qwen3.8-Flash-Next.
  Numbers inside the bars are case counts; percentages use the size of each
  track. Part-track failures are divided by whether the core-form checks pass.
  Assembly failures are assigned to the earliest failed stage in the order:
  generated parts, assembly structure, and motion verification.}
  \label{fig:qwen-failures}
\end{figure}

We analyze every case used for the Qwen3.8-Flash-Next row of
\cref{tab:main}. \Cref{fig:qwen-failures} assigns each failed case to one
primary failure stage. For the two part tracks, the geometric rubric
distinguishes a wrong core form from an error in a secondary requirement. For
assembly, we use a hierarchical assignment: part-generation failure first,
then invalid assembly structure, and finally failed motion verification after
the parts and structure pass their preceding checks. The categories are
therefore mutually exclusive.

\paragraph{Part tracks.}
Text-to-part rejects 12 of 100 cases. Only two, \texttt{crank} and
\texttt{handwheel}, fail the core-form checks; both omit or substantially
misconstruct several major features. The other ten preserve the main body but
miss a local requirement. Examples include missing flange bolts on
\texttt{pipe\_manifold}, misplaced corner holes on
\texttt{inch\_callout\_plate}, a missing branch on
\texttt{pipe\_route\_3d}, and incorrect flank, keyway, or center-hole geometry
on \texttt{worm\_shaft}. This agrees with the aggregate core-form rate of
0.98: once the main construction is correct, precise secondary features
remain the dominant source of rejection.

Three-view-to-part rejects 8 of 50 cases, but all eight retain the correct
core form. Their failures are localized drawing details, such as slot
placement on \texttt{b06\_lock\_striker}, the slot and boss geometry of
\texttt{d06\_u\_clevis}, flange-hole placement on
\texttt{d08\_pipe\_tee}, and the hole pattern of
\texttt{d19\_hinge\_leaf}. Thus, the image-conditioned agent usually recovers
the global shape, while small features whose dimensions and locations must be
matched across views remain harder.

\paragraph{Assembly track.}
Assembly rejects 25 of 50 cases. Only one failure begins in part generation:
\texttt{spirograph} has one rejected component among its 13 generated parts.
Eight further cases generate all required parts but fail the structural
assembly checks, typically because a required joint or coupling is missing or
has the wrong type. For example, \texttt{single\_cylinder\_engine} models the
rod--crank connection as cylindrical rather than revolute and the
cam--frame connection as planar rather than revolute; the more complex
\texttt{geared\_five\_bar} leaves several links merged, free, or incorrectly
constrained.

The largest group contains 16 cases whose parts and assembly structure pass,
but whose verified motion does not. These failures are invisible to a
structure-only check. \texttt{star\_gear\_reducer}, for example, receives a
structure score of 100 but the evaluator cannot create the requested rotary
drive. \texttt{orrery\_earth\_moon} also receives a structure score of 100
and a motion score of 91.67, yet fails one required kinematic relation.
\texttt{folding\_umbrella} has the expected component graph but cannot reach
the prescribed open and closed poses. Moreover, 21 of the 25 failed assembly
runs exhaust the 100-call budget. Together, these results show that the
remaining bottleneck is not producing individual solids, but completing and
validating long-horizon kinematic systems.

\subsubsection{Atomic API baseline with Qwen3.8-Flash-Next}
\label{sec:suppl:atomic-failures}

\begin{figure}[!h]
  \centering
  \includegraphics[width=\columnwidth]{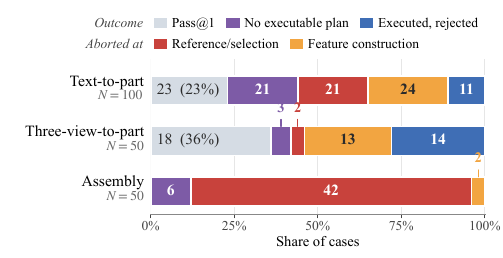}
  \caption{Case-level outcomes for the Atomic API baseline with
  Qwen3.8-Flash-Next. Numbers are case counts. Each failure is assigned to
  its earliest stage: no executable plan, the first failed atomic call
  (reference/selection or feature construction), or a fully executed plan
  rejected by the evaluator. No assembly case passes.}
  \label{fig:atomic-failures}
\end{figure}

We apply the same case-level analysis to the Atomic API row of
\cref{tab:main}. Because an atomic plan is generated once and executed
without feedback, we assign each failed case to the earliest stage at which
it breaks: no executable plan, the first failed atomic call, or a completed
plan whose result is rejected.
The analysis results are shown in \cref{fig:atomic-failures}.

Of the 159 failed cases, 134 never complete their plan, so most failures
occur before geometric quality can be judged. In 30 cases, the model returns
no executable plan: reasoning exhausts the output budget, or the long
assembly plan is truncated. These cases concentrate on long-horizon tasks,
including 13 of the 20 hard text-to-part tasks.
The other 104 cases stop at their first failed atomic call, and the rest of
the plan is never executed. In the aborted text-to-part runs, only 462 of
1{,}363 planned calls are reached.

The most common error is an unresolved reference (65 cases). The model must
name every plane, face, and edge in the low-level selector syntax; 16
text-to-part runs instead write names such as \texttt{Right Plane} or
\texttt{top}, and fillets or chamfers miss their target edge. In assembly,
42 mates fail because the referenced cylindrical face does not exist. In 22
of them, the component's own plan had aborted earlier, typically at the hole
cut, so the mate inherits an upstream failure. Seven gear mates reference a
cylinder at the pitch diameter, which a toothed gear does not contain.
Feature construction accounts for 39 further cases. Most are cut extrusions
whose hand-specified sketch plane, coordinates, or direction do not meet the
existing solid. The same failure ends 54 of the 65 aborted part plans inside
the assembly track.

Only 25 cases execute every atomic call and are still rejected. On
three-view tasks, 32 runs complete without an execution error, but only 18
pass. Error-free execution alone does not ensure a correct drawing reading.

\paragraph{Summary.}
Compared with \method, the Atomic API is too fine-grained. Each engineering
step becomes a long sequence of trivial calls, such as plane selection,
sketch entities, feature options, and mates, so even a moderate part requires
a long-horizon plan. Over such a horizon, small errors in coordinates or
references accumulate. A single wrong call halts the plan, and an upstream
part failure can break later assembly steps. Even basic operations, such as
selecting a plane, face, or edge, frequently fail. As a result, most plans
stop before the task is completed. The missing capability
is not the set of modeling operations, which the Atomic API exposes in full,
but an appropriate level of abstraction. \method packages these low-level
sequences into engineer-level workflows with semantic references, effect
checks, and local recovery, so the agent can plan over a short horizon and
repair one step before continuing.

\subsubsection{CADDesigner with Qwen3.8-Flash-Next}
\label{sec:suppl:caddesigner-failures}

\begin{figure}[!h]
  \centering
  \includegraphics[width=\columnwidth]{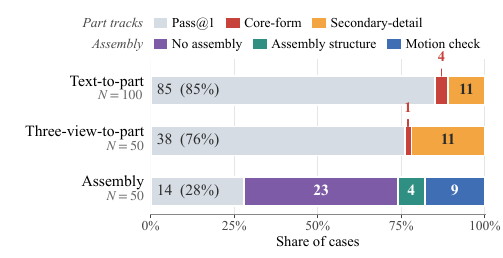}
  \caption{Case-level outcomes for CADDesigner with Qwen3.8-Flash-Next,
  using the categories of \cref{fig:qwen-failures}. ``No assembly'' marks
  runs that end without delivering an assembly. No case fails at part
  generation.}
  \label{fig:caddesigner-failures}
\end{figure}

We analyze the CADDesigner row of \cref{tab:main}, and
the analysis results are shown in \cref{fig:caddesigner-failures}.

\paragraph{Part tracks.}
CADDesigner rejects 15 of 100 text-to-part cases and 12 of 50 three-view
cases, compared with 12 and 8 for \method. It also fails the core form more
often (4 and 1 cases, compared with 2 and 0). Most rejected parts miss or
misplace a secondary feature. For example, \texttt{u\_joint\_yoke} has only
two of its four tapped holes, \texttt{b06\_lock\_striker} lacks both slots,
and \texttt{d36\_flanged\_tube} is modeled without its bore. The
code-generation loop is also costly. Across the two part tracks, 307 of 992
program executions (31\%) crash on API or programming errors, such as calls
to nonexistent methods like \texttt{Solid.get\_bbox} or low-level OCP calls
with mismatched signatures. Another 107 fail in the geometry kernel, for
example when a union leaves separated solids or a fillet cannot be built.
Because a script keeps no model state, the agent inspects a part by
rebuilding it; 149 further text-to-part executions return no part and
mostly serve such inspection. This overhead raises the token
cost above that of \method on both tracks (228.6k vs.\ 170.1k and 170.6k
vs.\ 136.7k).

\paragraph{Assembly track.}
The largest gap is on assembly, where CADDesigner fails 36 of 50 cases,
compared with 25 for \method. Most failures occur before any assembly can be
evaluated: 23 runs deliver no assembly. Each run executes 47.2 Python scripts
and reads 19.7 API documents on average, and the conversation retains all of
them. Consequently, 13 of these runs exceed the 73{,}728-token context
window. The other 10 exhaust the 100-call budget, with 37 to 65 of their 100
replies containing no tool call. Overall, only 18 of the 50 runs call
\texttt{finish}. By contrast, \method edits a persistent native assembly, so
every run, including those stopped by the call budget, leaves a scorable
model (lowest structure score: 31.9). Among the 27 delivered assemblies,
four fail the structural checks. For example, \texttt{lead\_screw\_stage}
and \texttt{gear\_screw\_lift} leave the mechanism without the required
mobility. Nine more pass the structural checks but fail motion
verification; \texttt{star\_gear\_reducer} receives a structure score of 100
but a motion score of only 33.3.

\paragraph{Summary.}
On single parts, CADDesigner is competitive with \method. Its whole-program
interface, however, has three weaknesses relative to \method. API usage is
unchecked, so many executions crash before producing geometry. The model has
no persistent state, so every inspection or repair regenerates and re-runs
the full program. The resulting context and budget pressure grows with the
task horizon. These weaknesses are tolerable on single parts but dominate on
long-horizon assembly, where nearly half of the runs deliver no assembly.
\method avoids them by invoking typed workflows on a persistent native model
and checking each step as it is executed. Its intermediate state therefore
remains valid and scorable throughout the run.

\subsection{Workflow Coverage}
\label{sec:suppl:workflow-coverage}
\label{sec:suppl:operation-coverage}

\begin{table*}[t]
\caption{Structural coverage of the workflow interface and its expansion into
native operations. Each row is a parameterized capability rather than a
task-specific macro; this is an audit of the implemented action space, not a
coverage statistic derived from the workflow-mining corpus or \bench.}
\label{tab:coverage}
\label{tab:suppl-operation-coverage}
\centering
\footnotesize
\renewcommand{\arraystretch}{1.12}
\setlength{\tabcolsep}{4pt}
\begin{tabular}{>{\raggedright\arraybackslash}p{1.15in}
                >{\raggedright\arraybackslash}p{2.55in}
                >{\raggedright\arraybackslash}p{2.75in}}
\toprule
\textbf{Dimension} & \textbf{Generic workflow mechanisms} &
\textbf{Native expansion and scope} \\
\midrule
Reference, profile, and path &
Semantic datum planes and observed faces; eight profiles; open 2D paths,
splines, helices, and 3D paths. &
Plane/face selection, reference-plane creation, batched circles, polylines,
and general line--arc--spline profiles. Placement is separated from
geometry. \\
Primary solid &
\texttt{extrude}, \texttt{cut}, \texttt{revolve}, \texttt{sweep},
\texttt{loft}, and \texttt{axis\_cylinder}, with independent placement,
direction, and end conditions. &
Native extrusion, cut, revolve, sweep, loft, and up-to-geometry features;
profiles, paths, placement, and termination combine independently. \\
Derived features &
Intent-specific holes, fillet, chamfer, rib, shell, draft, sheet-metal base,
patterns, and mirror. &
Native local-feature and transform operations. Current sheet-metal support
is limited to a base flange; full bends and edge-flange families are
excluded. \\
Revision and control &
Named parameters, state observation, feature editing and deletion, effect
checks, rollback, and optional native views. &
Dimension modification, feature-tree and geometry queries, persistence, and
recovery. Document and control operations remain separate from shape
operators. \\
Assembly structure &
Insertion and grounding; generic mate; seven joint types; loop closure; gear,
rack--pinion, screw, and linear couplings. &
Seven atomic modeling operations expand semantic joints and couplings into native mates
while preserving the ability to revise or remove them. \\
Assembly validation &
State, degree-of-freedom, interference, motion, prescribed-pose checks, and
optional native views. &
Native assembly state and motion measurements close the construction loop;
these observers complement rather than replace structural mate operations. \\
\bottomrule
\end{tabular}
\end{table*}

Why can a compact catalog support a broad range of native CAD operations?
The key is factorization, rather than one workflow per SolidWorks command.
As summarized in \cref{tab:coverage}, we factor each part-modeling action
into several largely independent components, including a semantic reference
and local frame, a profile or path, a feature operator, its direction and end
condition, and execution checks. These choices are largely orthogonal.

For example, the same profile language can define the section of an
extrusion, cut, revolve, sweep, or loft; changing the reference places that
section on a datum plane or an existing face; and changing the end condition
gives a blind, symmetric, through-all, or up-to-geometry result. The eight
profile primitives also include general line--arc--spline paths, so a finite
schema represents an open set of cross-sections and trajectories rather than
a fixed shape library.

The remaining workflows cover operations that do not fit this
profile--operator factorization: holes with intent-specific checks, derived
features such as fillets, shells, ribs, drafts, patterns, and mirrors, and
stateful operations for parameter editing, deletion, inspection, and
rollback. Spatial constructions use semantic faces, axes, and reference
planes instead of adding a separate workflow for each orientation. Complex
motifs therefore remain compositions: a bent tube is a sweep over a 3D path
with a hollow profile, while a stepped shaft is a sequence of revolved
sections. 

Assembly modeling follows the same principle. 
Seven atomic modeling operations insert components, set their poses, ground them, 
create or delete mates, and save the assembly.
Engineer-level workflows
compose them into fixed, revolute, cylindrical, slider, planar, ball, and
pin--slot joints, then add loop closure and gear, rack--pinion, screw, or
linear couplings. State, mobility, interference, motion, and prescribed-pose
checks close the loop. Thus, the interface spans the construction lifecycle
from creating geometry to validating mechanism behavior, rather than merely
covering a list of isolated feature commands.

To summarize, our extracted workflows establish structural coverage
of parametric solid and mechanism modeling, but not of the entire SolidWorks
product: drawings, simulation, CAM, weldments, routing, and the full
surface-design and sheet-metal modules remain outside our current scope.

\subsubsection{Coverage statistics on Text2CAD dataset}

We measure whether the part workflows of \method can express the operations
required by the Text2CAD test set~\citep{khan2024text2cad}, whose
ground-truth models come from DeepCAD~\citep{wu2021deepcad}.
We parse the original DeepCAD construction sequences of the de-duplicated
Text2CAD test pool and consider the 6{,}895 models whose ground truth is a
single solid. For every extrusion we record its Boolean operation, end
condition, draft angle, sketch-plane orientation, and curve types, and we
rebuild each model step by step to track the number of solids after each
feature. Each requirement is then mapped to the workflow interface: a model
is \emph{directly covered} if every feature corresponds to one workflow call,
and \emph{covered by rewriting} if it becomes expressible after an
equivalent reformulation that preserves the final geometry.

The required operation set is narrow. All features are sketch--extrude pairs
whose profiles use only lines (78.7\%), arcs (10.2\%), and circles
(11.1\%), all of which the profile language supports. Among the 10{,}939
extrusions, 99.9\% are new-body, join, or cut operations, 98.5\% use
one-sided or symmetric end conditions, none uses a draft angle, and 99.8\%
lie on axis-aligned sketch planes. These map directly to \texttt{extrude}
and \texttt{cut} on datum planes, offset planes, or existing planar faces.

As a result, 92.4\% of the single-solid models are directly covered, and a
further 7.3\% are covered by rewriting. The rewriting cases are an
asymmetric two-sided extrusion, expressed as a one-sided extrusion from an
offset plane (1.8\%); an intermediate state with disconnected pieces,
resolved by reordering features (5.5\%); and a circular
extrusion on an inclined plane, expressed with \texttt{axis\_cylinder}
(0.1\%). In total, 99.7\% of the single-solid models (6{,}877 of 6{,}895)
are covered by the workflow catalog. The remaining 0.3\% need an
intersection feature or a non-circular sketch on an inclined plane. The
200-case evaluation subset shows the same picture: 189 of its 191
single-solid models (99.0\%) are covered. The remaining gap to the ground
truth in \cref{tab:text2cad-results} therefore mainly reflects the
under-specified abstract prompts rather than missing operations.

\subsubsection{Coverage statistics on \bench}

Every one of the 200 benchmark cases is paired with an executable oracle
modeling trajectory. Starting from an empty SolidWorks document, each
trajectory uses only the same workflows exposed to the evaluated agent to
construct the required part or assembly. The model produced by every oracle
trajectory receives a score of 100 under the corresponding track evaluator.
This covers all 100 text-to-part cases, 50 three-view-to-part cases, and 50
assembly cases, demonstrating that every task has at least one valid
construction within the Workflow action space.
Thus, \bench has 100\% constructive coverage by our workflow
interface: benchmark failures cannot be attributed simply to a required
operation being absent from that interface. 

\section{Ablation Studies Details}
\label{sec:suppl:abla}

\subsection{Workflow Components Explanations}

\Cref{tab:ablation} evaluates leave-one-component-out variants on all 50
three-view-to-part tasks. All variants use the same self-hosted
Qwen3.8-Flash-Next backbone, medium reasoning effort, and 40-call budget. They receive the same
input engineering drawing and expose the same modeling operations except for
the component named by each row. We choose this track because it jointly
tests multimodal understanding of engineering drawings and textual reasoning
over dimensions, tool observations, and multi-step CAD operations.
We next describe each ablation in turn, specifying what is removed and what
functionality remains available.

\paragraph{Without canonical \texorpdfstring{$(u,v)$}{(u,v)} frames.}
The full workflow gives every sketch plane a shared, world-aligned 2D frame:
the two coordinates correspond to fixed model axes, and the runtime maps them
to the selected SolidWorks plane or face. This variant removes that canonical
mapping and its coordinate hints from tool schemas and state observations.
It does not remove the ability to specify 2D geometry. Instead, before each
sketch operation, the agent must call \texttt{inspect\_sketch\_frame} to
obtain the plane's native SolidWorks origin and local \((s,t)\) basis, then
express the profile in that frame. The runtime reopens the exact inspected
sketch for the subsequent feature. Thus, the ablation isolates the shared
coordinate convention while retaining an exact, executable native-coordinate
alternative.

\paragraph{Without semantic face references.}
The full workflow accepts stable geometric descriptions such as the
outermost \(+Z\) face, top or bottom faces, and named faces, and resolves them
against the current feature tree. This variant removes those directional and
named face references. Datum planes, coordinate-offset planes, explicitly
created reference planes, and raw \texttt{face\#<index>} identifiers from the
latest model-state observation remain available. The agent can therefore
perform the same face-based operations, but it must repeatedly inspect the
state and select transient native indices rather than referring to a face by
its geometric meaning.

\paragraph{Without profile pre-checks.}
The full workflow validates a 2D profile before sending any mutating request
to SolidWorks. These checks detect malformed or self-intersecting regions,
partially overlapping loops, invalid inner/outer-loop relationships,
out-of-bound profiles, and inconsistent arc or path geometry, and return a
structured repair hint without changing the model. This variant disables
those geometric pre-execution checks and sends the proposed profile directly
to SolidWorks. Typed tool schemas, numeric and unit conversion, required
arguments, SolidWorks API errors, post-execution checks, and rollback remain
enabled. It therefore removes early geometric protection rather than general
argument validation.

\paragraph{Without post-check and rollback.}
After a mutating call, the full workflow checks whether the intended native
feature was created and whether its geometric effect is plausible, including
the expected volume direction, body count, connectivity, and nonzero effect.
If a semantic post-condition fails, it deletes the features created by that
transaction and returns a structured error so that the agent can retry from
the previous clean state. This variant removes both the geometric
post-condition test and that automatic rollback. Schema validation, profile
pre-checks, and API/COM error reporting are retained, but suspicious or
partially created features can remain in the feature tree and influence later
calls.

\paragraph{Without native visual inspection.}
The full multimodal workflow can inspect the evolving model through
\texttt{view\_model}, which returns labeled native SolidWorks views, and
\texttt{render\_views}, which returns standardized three-view projections for
comparison with the task drawing. This variant removes both feedback tools
from the callable catalog. It still receives the original engineering drawing
as multimodal input and retains all structured model-state observations and
geometry tools. The variant therefore measures the contribution of visual
feedback during construction, not the contribution of the input image itself.

\subsection{Expert-designed workflow vs. data-mined workflow}
\label{sec:suppl:abla_expert}

To isolate the value of large-scale workflow mining from the value of using
engineer-level actions in general, we compare two Workflow-CAD variants. The
\emph{expert-designed} variant uses a workflow catalog constructed manually
from the domain knowledge of one mechanical engineer, without mining feature
trees or assembly relations.
The \emph{data-mined} variant is the full \method system, whose reusable
workflows are extracted from large collections of part construction histories
and assembly constraints. We keep the agent loop, SolidWorks backend, LLM
backbone, prompts, call budgets, and evaluators fixed, so the workflow source
is the only intended difference. A performance gap between the two variants
therefore directly measures the additional value contributed by data mining
beyond expert design alone.


The results are shown in \cref{tab:workflow-source-ablation}. Data-mined
workflows improve performance on all three tracks. Overall pass@1 rises from
0.60 to 0.78 and the mean score from 78.4 to 95.1. The expert-designed
catalog does not cover the full variety of part-feature operations or
assembly mate and coupling operations. The agent must therefore compose
longer sequences from less suitable tools or repeatedly work around missing
operations, which lowers modeling accuracy. This incomplete coverage also
makes the expert-designed variant less token-efficient: its mean consumption
is 1014.5k tokens per task, compared with 792.0k for the data-mined variant,
a reduction of 21.9\%. These consistent quality and efficiency gains show
that large-scale mining contributes value beyond manually designing a small
set of workflows from expert knowledge alone.

\begin{table*}[t]
\caption{Expert-designed versus data-mined workflows under the same agent,
backbone, budgets, SolidWorks backend, and evaluation protocols. Tokens are
mean counts per task.}
\label{tab:workflow-source-ablation}
\centering
\scriptsize
\setlength{\tabcolsep}{1pt}
\resizebox{\textwidth}{!}{%
\begin{tabular}{l cccc cccc cccc ccc}
\toprule
& \multicolumn{4}{c}{\textbf{Text-to-part}} &
\multicolumn{4}{c}{\textbf{Three-view-to-part}} &
\multicolumn{4}{c}{\textbf{Assembly}} &
\multicolumn{3}{c}{\textbf{Overall}} \\
\cmidrule(lr){2-5}\cmidrule(lr){6-9}\cmidrule(lr){10-13}
\cmidrule(lr){14-16}
\textbf{Workflow source} &
Pass@1 & Score & Form & Tokens &
Pass@1 & Score & Image & Tokens &
Pass@1 & Score & Motion & Tokens &
Pass@1 & Score & Tokens \\
\midrule
Expert-designed Workflow
  & 0.64 & 76.2 & 0.88 & 218.3k
  & 0.68 & 84.5 & 87.8 & 166.5k
  & 0.42 & 76.5 & 53.2 & 3454.9k
  & 0.60 & 78.4 & 1014.5k \\
Data-mined Workflow
  & 0.88 & 96.5 & 0.98 & 170.1k
  & 0.84 & 96.4 & 97.6 & 136.7k
  & 0.50 & 91.1 & 68.3 & 2691.0k
  & 0.78 & 95.1 & 792.0k \\
\bottomrule
\end{tabular}
}
\end{table*}
\section{Visualization}
\label{sec:suppl:visualization}

\Cref{fig:suppl-workflow-cad-text,fig:suppl-workflow-cad-drawing,%
fig:suppl-workflow-cad-gears,fig:suppl-workflow-cad-linkage} provide
qualitative examples of the \method agentic system across the three benchmark
tracks. The examples show the task input and agent execution trace alongside
the resulting native CAD model. For additional visualizations of \bench,
please refer to \cref{fig:teaser}.

\begin{figure*}[t]
\centering
\includegraphics[width=\textwidth]{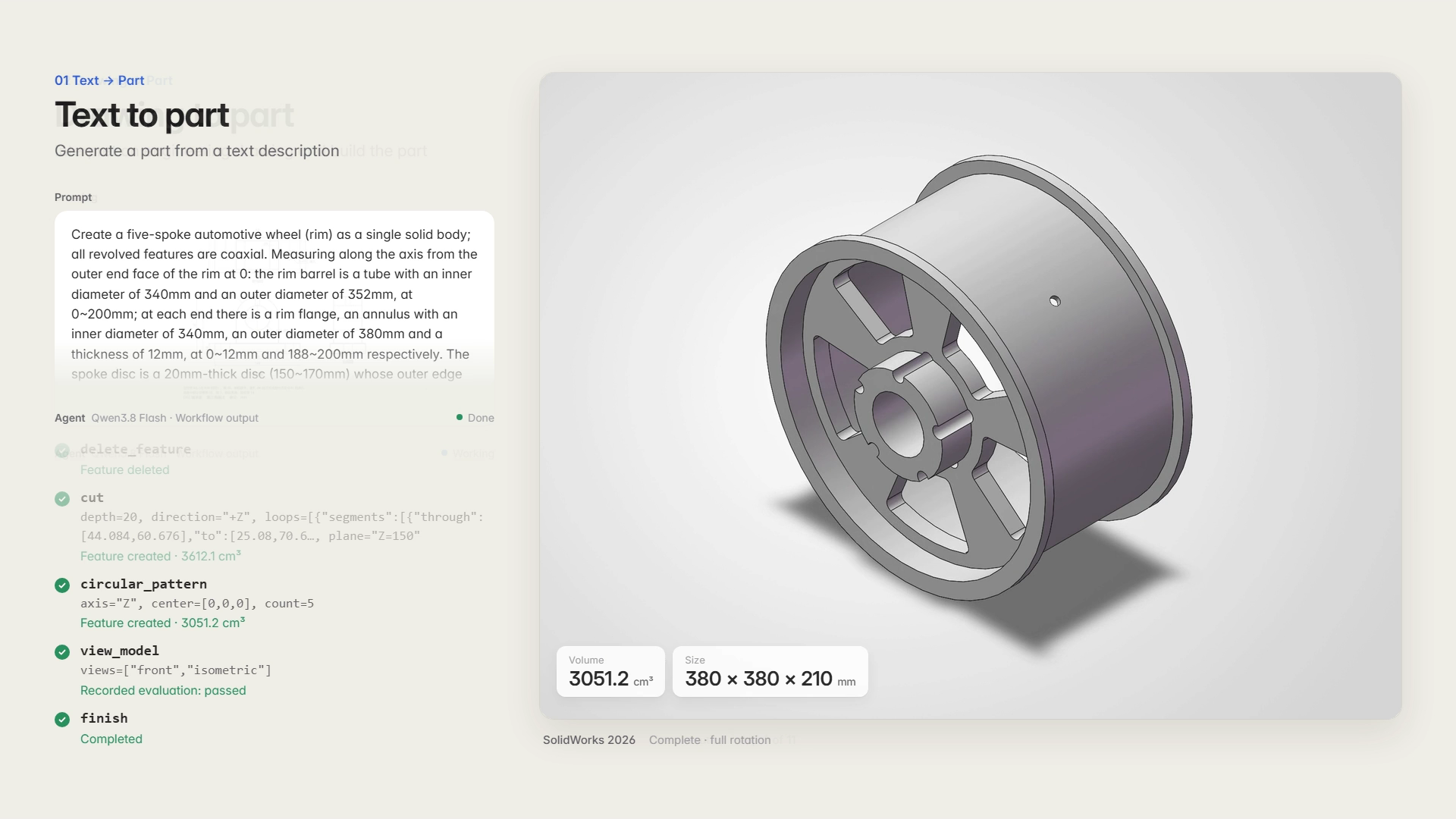}
\caption{Text-to-part visualization of the \method agentic system. The agent
constructs a five-spoke wheel rim from a natural-language specification.}
\label{fig:suppl-workflow-cad-text}
\end{figure*}

\begin{figure*}[t]
\centering
\includegraphics[width=\textwidth]{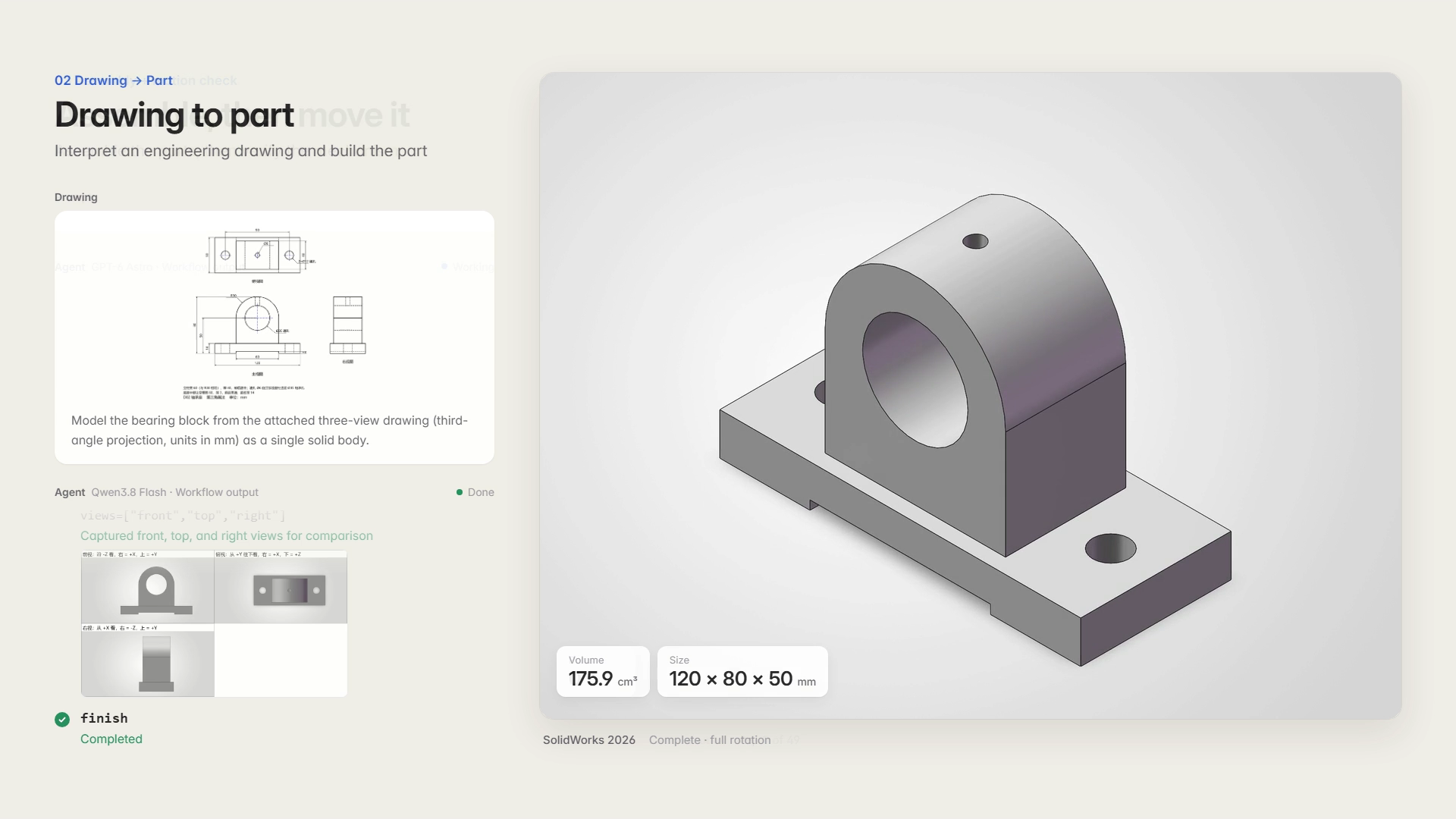}
\caption{Drawing-to-part visualization of the \method agentic system. The
agent interprets a dimensioned three-view drawing and reconstructs the
corresponding bearing block.}
\label{fig:suppl-workflow-cad-drawing}
\end{figure*}

\begin{figure*}[t]
\centering
\includegraphics[width=\textwidth]{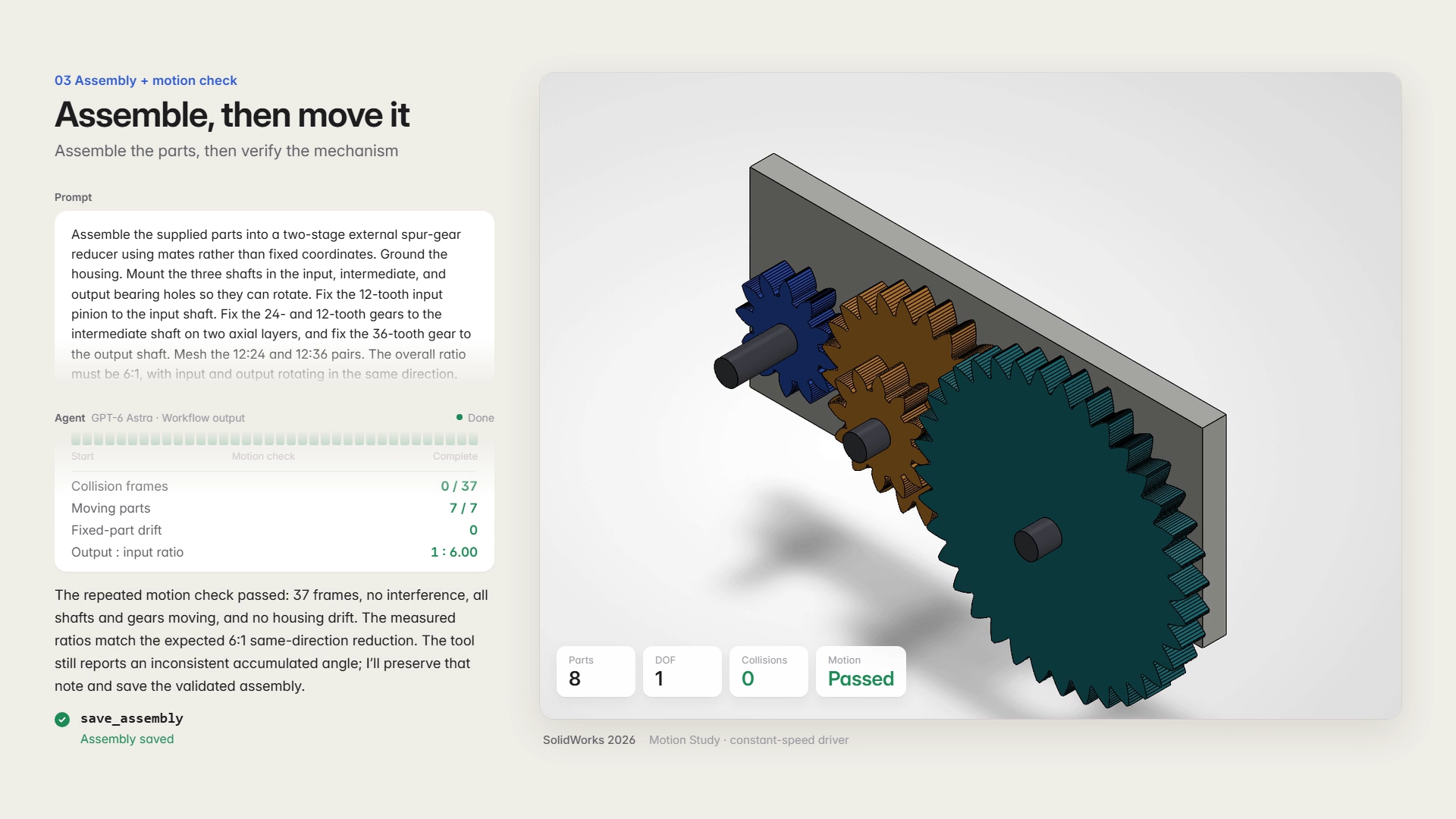}
\caption{Assembly visualization of the \method agentic system. The agent
builds a two-stage gear train and verifies its motion, gear ratio, and
collision-free operation.}
\label{fig:suppl-workflow-cad-gears}
\end{figure*}

\begin{figure*}[t]
\centering
\includegraphics[width=\textwidth]{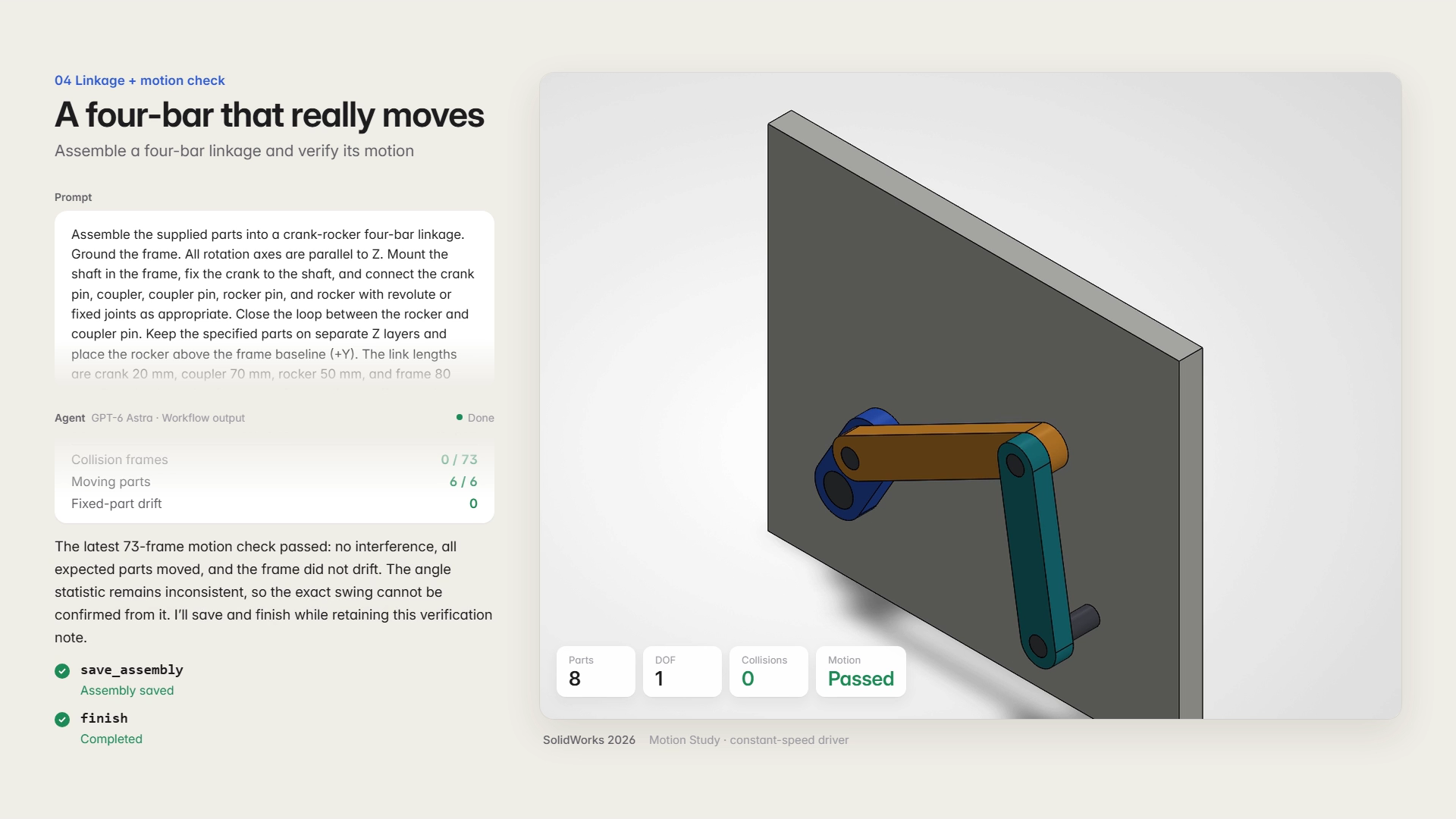}
\caption{Assembly visualization of the \method agentic system. The agent
assembles a crank--rocker four-bar linkage and validates the resulting motion
over sampled configurations.}
\label{fig:suppl-workflow-cad-linkage}
\end{figure*}

The supplementary material also includes two videos. The first demonstrates
the end-to-end operation of the \method agentic system, including native CAD
construction and iterative verification. The second presents
\bench visualizations, covering representative tasks and outputs across its
three tracks.

\section{Limitation}
\label{sec:suppl:limitations}

Our workflow coverage is deliberately limited to parametric solid and
mechanism modeling in SolidWorks. The factorized profile, feature, reference,
joint, and coupling interfaces cover a broad operation space, but they do not
expose engineering drawings, simulation, CAM, weldments, routing, or the full
surface-design and sheet-metal modules.

\bench evaluates multi-turn CAD modeling from scratch. It does not assess
human--agent collaboration or an agent's ability to inspect and modify an
existing CAD model created by a human user.

\section{Human participants and ethics statement}
\label{sec:suppl:ethics}

This work involves two groups of human participants: (i) the participants who
produced the human baseline in \cref{tab:main,tab:backbones}, and (ii) the
mechanical-domain experts who authored, guided, and reviewed the benchmark
cases (\cref{sec:bench:anno} and \cref{sec:suppl:abla_expert}). No other data are collected from people. The
corpora used for workflow mining (CAD-1K~\citep{markovai2026cad1k}, Fusion 360
Gallery~\citep{DBLP:journals/tog/WillisPLCDLSM21}, and
AutoMate~\citep{jones2021automate}) are publicly released CAD datasets, 
and their license/terms permit this use.

\paragraph{Ethics review.}
The study was reviewed by the Institutional Review Board of
our institution and determined to be exempt from full review
because it is a minimal-risk task performance study.

\paragraph{Human baseline.}
The human baseline was produced by 3 graduate students with
professional training in mechanical engineering. 
Each participant solved all 200
tasks in SolidWorks, receiving the same inputs as the agents
(the text brief or the three-view drawing) under a time limit of two hours per
task. Participants did not see the evaluation rubrics or the reference models,
and did not use AI assistants. Each task was attempted once, and
the delivered model was scored by the same evaluators as the agent outputs.

\paragraph{Mechanical-domain experts.}
The benchmark cases were produced with one mechanical-domain experts,
who authored about 50 cases
directly in SolidWorks, guided and corrected the LLM-assisted construction of
the remaining cases, reviewed every finished case, and reviewed the generated
text prompts.

\paragraph{Consent and compensation.}
All participants were informed of the purpose of the study and of how their
outputs would be used, including the future release of benchmark data and the
aggregate reporting of baseline results, and gave electronic
informed consent before participating. 
Both groups were compensated.

\paragraph{Data protection and risks.}
We record only the delivered CAD files and task completion times; no names,
demographics, screen recordings, or other personally identifiable information
are stored with the results. Human-baseline results are reported
only in aggregate, and individual outputs are not released. 
The released benchmark contains CAD models, text
briefs, drawings, and evaluation scripts, and no personal information. The
tasks are standard CAD modeling exercises; they pose minimal risk, and no
minors or vulnerable individuals were involved.

\end{document}